\documentclass[10pt,twocolumn,letterpaper]{article}

\usepackage[pagenumbers]{cvpr} 

\usepackage{graphicx}
\usepackage{amsmath}
\usepackage{amssymb}
\usepackage{booktabs}
\usepackage{algorithm}
\usepackage{lipsum}
\usepackage{url}

\usepackage{mathtools}

\DeclarePairedDelimiter\floor{\lfloor}{\rfloor}

\usepackage{tabularx}
\usepackage{multirow}

\usepackage{trimclip}
\newcommand{\shorttextrightarrow}{\fontfamily{cmr}\clipbox*{{.3\width} 0pt {\width} {1ex}} \textrightarrow}

\usepackage{listings}
\usepackage{upquote}  
\usepackage{xcolor,colortbl}
\lstdefinestyle{overleaf}{
    backgroundcolor=\color[rgb]{0.95,0.95,0.92},   
    commentstyle=\color[rgb]{0,0.6,0},
    keywordstyle=\color{magenta},
    numberstyle=\tiny\color[rgb]{0.5,0.5,0.5},
    stringstyle=\color[rgb]{0.58,0,0.82},
    basicstyle=\ttfamily\footnotesize,
    breakatwhitespace=false,         
    breaklines=true,                 
    captionpos=b,                    
    keepspaces=true,                 
    numbers=left,                    
    numbersep=5pt,                  
    showspaces=false,                
    showstringspaces=false,
    showtabs=false,                  
    tabsize=2
}
\lstdefinestyle{mocov3}{
  backgroundcolor=\color{white},
  basicstyle=\fontsize{7.5pt}{7.5pt}\ttfamily\selectfont,
  columns=fullflexible,
  breaklines=true,
  captionpos=b,
  commentstyle=\fontsize{7.5pt}{7.5pt}\color[rgb]{0.25,0.5,0.5},
  keywordstyle=\fontsize{7.5pt}{7.5pt}\color[rgb]{0.85,0.18,0.50},
}
\usepackage{etoolbox}
\makeatletter
\AfterEndEnvironment{algorithm}{\let\@algcomment\relax}
\AtEndEnvironment{algorithm}{\kern2pt\hrule\relax\vskip3pt\@algcomment}
\let\@algcomment\relax
\newcommand\algcomment[1]{\def\@algcomment{\footnotesize#1}}
\renewcommand\fs@ruled{\def\@fs@cfont{\bfseries}\let\@fs@capt\floatc@ruled
  \def\@fs@pre{\hrule height.8pt depth0pt \kern2pt}%
  \def\@fs@post{}%
  \def\@fs@mid{\kern2pt\hrule\kern2pt}%
  \let\@fs@iftopcapt\iftrue}
\makeatother

\usepackage[pagebackref,breaklinks,colorlinks]{hyperref}

\usepackage[capitalize]{cleveref}
\crefname{section}{Sec.}{Secs.}
\Crefname{section}{Section}{Sections}
\Crefname{table}{Table}{Tables}
\crefname{table}{Tab.}{Tabs.}

\newcommand{\PAR}[1]{\vskip4pt \noindent {\bf #1~}}

\def\cvprPaperID{0000} 
\def\confName{CVPR}
\def\confYear{2023}

\newcommand{\authsep}{\;\;}

\begin{document}

\title{FlexiViT: One Model for All Patch Sizes}

\author{Lucas Beyer$^{\star}_1$ \authsep Pavel Izmailov$^{\star}_{1,3}$ \authsep Alexander Kolesnikov$^{\star}_1$ \authsep Mathilde Caron$^{\star}_2$ \authsep Simon Kornblith$^{\star}_1$ \\
Xiaohua Zhai$^{\star}_1$ \authsep Matthias Minderer$^{\star}_1$ \authsep Michael Tschannen$^{\star}_1$ \authsep Ibrahim Alabdulmohsin$^{\star}_1$ \authsep Filip Pavetic$^{\star}_1$ \\
\\
Google Research}
\maketitle
{\let\thefootnote\relax\footnote{
{$^{\star}$All authors made significant technical contributions. \\
$^{\quad\;\;\ \phantom{0}}$ Lucas started and led the project.\\ 
$^{\quad\;\;\ 1}$ Google Research, Brain Team.
$^{\quad\;\;\ 2}$ Google Research.\\
$^{\quad\;\;\ 3}$ work done at Google Brain, while being a PhD student at NYU.}}}

\begin{figure}[t]
  \centering
  \includegraphics[width=0.8\linewidth]{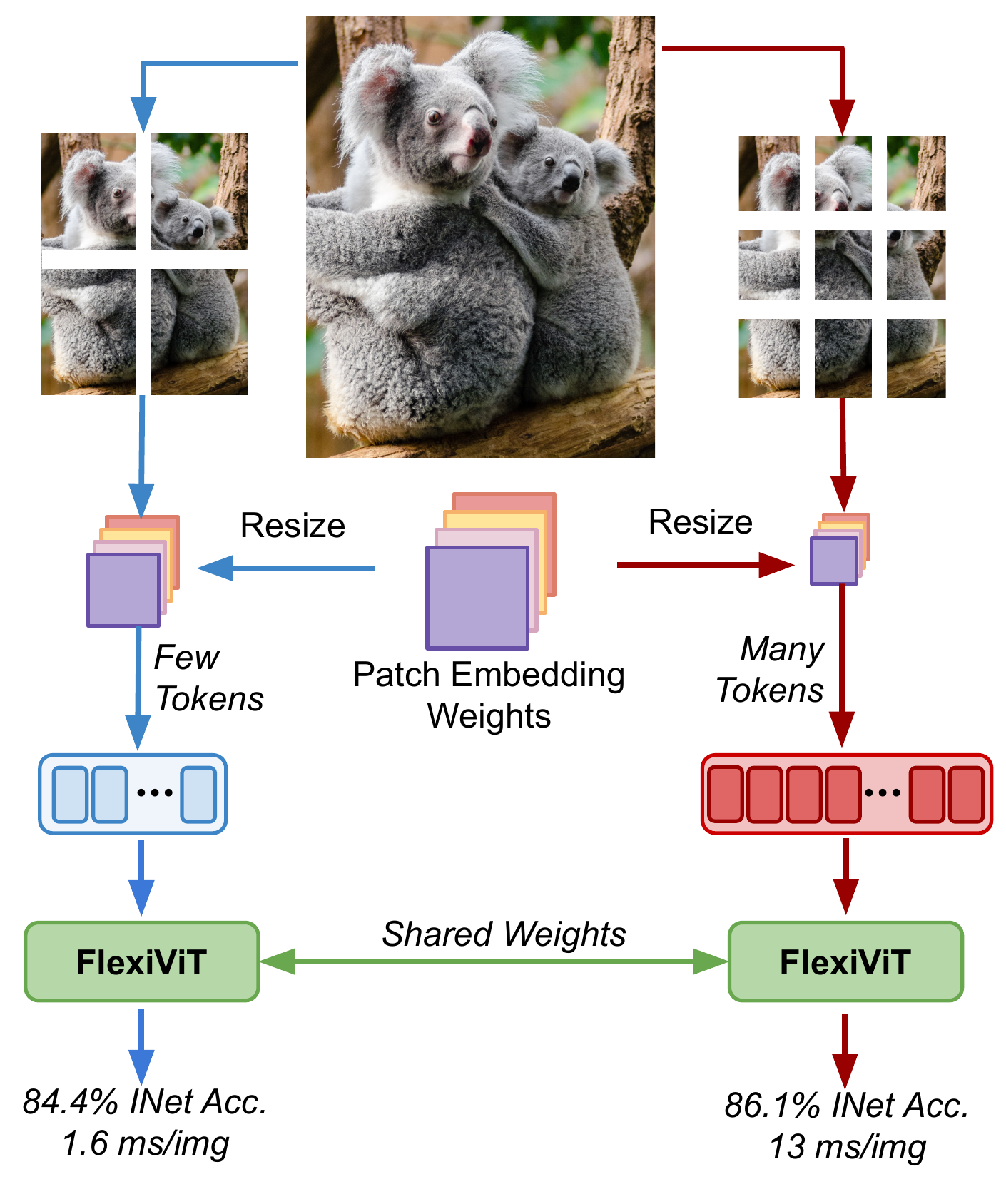}
  \caption{\textbf{FlexiViT} is a standard ViT model that sees randomized patch sizes, hence sequence lengths, during training.
  The patch embedding weights are resized adaptively for each patch size and the model weights are shared as-is across all patch sizes.}
  \label{fig:teaser}
\end{figure}

\vspace{-1em}
\begin{abstract}
Vision Transformers convert images to sequences by slicing them into patches.
The size of these patches controls a speed/accuracy tradeoff, with smaller patches leading to higher accuracy at greater computational cost, but changing the patch size typically requires retraining the model.
In this paper, we demonstrate that simply randomizing the patch size at training time leads to a single set of weights that performs well across a wide range of patch sizes, making it possible to tailor the model to different compute budgets at deployment time.
We extensively evaluate the resulting model, which we call FlexiViT, on a wide range of tasks, including classification, image-text retrieval, open-world detection, panoptic segmentation, and semantic segmentation, concluding that it usually matches, and sometimes outperforms, standard ViT models trained at a single patch size in an otherwise identical setup.
Hence, FlexiViT training is a simple drop-in improvement for ViT that makes it easy to add compute-adaptive capabilities to most models relying on a ViT backbone architecture.
Code and pre-trained models are available at \url{github.com/google-research/big_vision}.
\end{abstract}

\begin{figure}[t]
  \centering
  \includegraphics[width=1.0\linewidth]{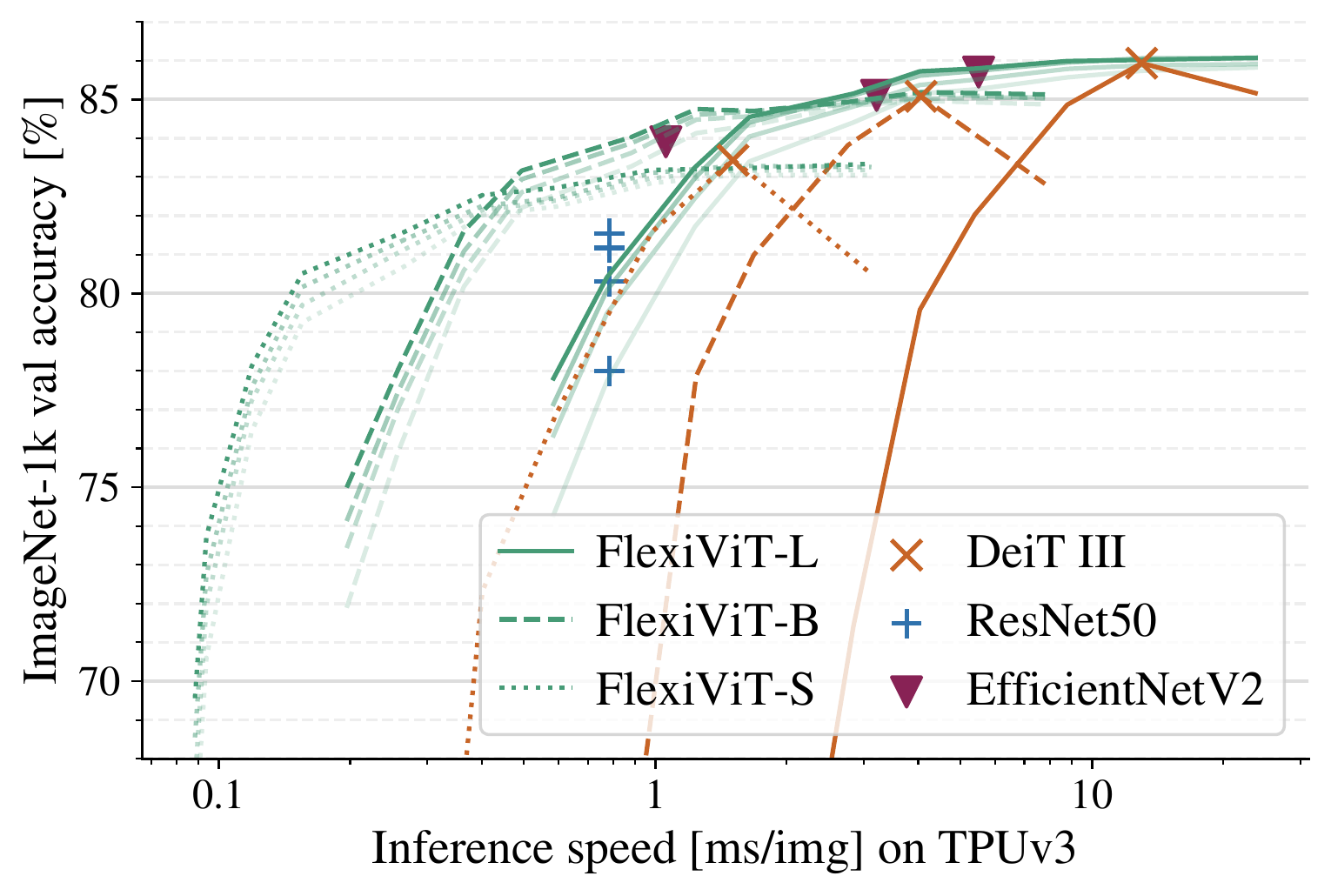}\vspace{-10pt}
  \caption{\textbf{FlexiViT results on ImageNet-1k.}
  We train three FlexiViTs based on DeiT\,III on ImageNet-1k and show their speed-accuracy tradeoff when evaluated at various patch sizes.
  Each curve corresponds to a single model with a single set of weights run with different patch sizes.
  We also evaluate DeiT\,III at various patch sizes to show ViT's natural inflexibility.
  EfficientNet-v2 numbers from~\cite{effnetv2} and ResNet50 numbers from~\cite{beyer2022knowledge}, the latter distills from an ImageNet-21k pretrained teacher.
  FlexiViT was trained for 1000 epochs, but runs for 600, 300, and 90 epochs shown as shaded curves indicate that long training mostly benefits the short-sequence setting and is not strictly necessary.}\label{fig:i1k}\vspace{-10pt}
\end{figure}

\section{Introduction}\label{sec:intro}

Vision Transformers (ViTs) cut images into non-overlapping patches and perform all computations on tokens created from these patches. This ``patchification'' procedure represents a significant shift away from the previously dominant convolutional neural network (CNN) approach~\cite{lecun1998gradient}, where an image is processed with small local and typically overlapping filters.
Patchification has unlocked new capabilities, such as (random) dropping of image patch tokens~\cite{mae,autoscalingvit,rao2021dynamicvit,tang2022patch,avit}, adding specialized tokens for new tasks~\cite{deit, cait} or mixing image tokens with tokens from other modalities~\cite{mustafa2022multimodal,merlotreserve,vatt}.

Despite the importance of patchification for ViT models, the role of the patch size has received little attention.
While the original ViT paper~\cite{dosovitskiy2021vit} works with three patch sizes (32$\times$32, 16$\times$16, and 14$\times$14 pixels), many follow-up works fix the patch size at 16$\times$16 pixels~\cite{zhai2022scaling,deit,deit3}.
In this work, we show that \emph{the patch size provides a simple and effective lever to change the compute and predictive performance of a model, without changing model parametrization}.
For example, a ViT-B/8 model achieves $85.6\%$ top-1 accuracy on ImageNet1k with 156\,GFLOPs and 85\,M parameters, while a ViT-B/32 model achieves only $79.1\%$ accuracy with 8.6\,GFLOPs and 87\,M parameters.
Despite the major difference in performance and compute, these models have essentially the same parametrization.
However, standard ViT models perform well only at the patch size that they have been trained at. Tuning the patch size therefore requires complete re-training of the model.

To overcome this limitation, we propose \textit{FlexiViT}, a flexible ViT which matches or outperforms standard fixed-patch ViTs across a wide range of patch sizes with no added cost.
To train FlexiViT, we randomize the patch size during training, and resize the positional and patch embedding parameters adaptively for each patch size, as shown in Figure~\ref{fig:teaser}.
These simple modifications are already sufficient for strong performance, but we also propose a optimized resizing operation and a training procedure based on knowledge distillation which achieves even better results.

We demonstrate the efficiency of FlexiViT models in many downstream tasks, such as image classification, transfer learning, panoptic and semantic segmentation, image-text retrieval and open-world recognition,
and provide a general recipe for \textit{flexifying} existing ViT-based training setups.
Furthermore, we show that \textit{flexibility} of the backbone, i.e. strong performance across patch sizes, is often preserved even after fine-tuning with a fixed patch size.
We leverage this observation to perform resource-efficient transfer learning: we finetune the model cheaply with a large patch size, but then deploy it with a small patch size for strong downstream performance.
We further show that flexible patch size can be used to accelerate pre-training.

To explain the effectiveness of FlexiViT, we analyze the model's representations. We find that the representations are often similar across different patch sizes, especially in the deeper layers.
Finally, we show that FlexiViT outperforms alternative architectural ways of controlling the performance-compute trade-off in ViT models.


\section{Related work}\label{sec:relwork}

Several recent works explore improving ViT's efficiency by exploiting patchification.
Some suggest removing tokens, either in randomized~\cite{mae} or structured~\cite{autoscalingvit} fashion throughout training.
Others aim to quantify a token's importance and remove the least important ones, during~\cite{rao2021dynamicvit,avit} or after~\cite{tang2022patch} training.
~\cite{dynamic_vit} trained a cascade of Transformers using increasing number of tokens to allow early exiting during inference.
Conversely, we always keep all tokens and do not discard any information.
It may be possible to combine such approaches with FlexiViT in future work.

Another related line of work looks at changing input resolution during training, typically in order to speed-up pre-training~\cite{multigrid,effnetv2,swinv2,deit3} or as data augmentation for self-supervised learning of ViTs~\cite{effssvit}.
We do not explore the data augmentation inpact of Flexi training to avoid an explosion in scope.
The aforementioned models all work only at their \emph{single}, final resolution, while our ai is to have \emph{one model} that works well at \emph{all trained patch-sizes}.

More similar to our approach, the Neural Architecture Search (NAS) field is converging towards training one ``supernet'' from which individual, differently-shaped ``subnets'' can be extracted~\cite{ofanet,bignas,nasvit}.
Since these works aim for changes in most or all model dimensions, they usually involve multiple specialized architectural additions.
SuperViT~\cite{supervit} is most related to FlexiViT as it patchifies an image at multiple scales, passes all these patches to ViT, while dropping random tokens~\cite{mae} to reduce the sequence length.
In contrast to the aforementioned works, our sharpened focus on ViT's patch size only, allows benefiting from existing pretrained models, future ViT improvements, and is an easy drop-in to any existing training procedure.

Matryoshka representation learning~\cite{matryoshka} proposes training models whose output vector contains meaningful sub-vectors. This can be seen as the complement of FlexiViT.

\section{Making ViT flexible}\label{sec:method}

In this section we show that standard ViT models are not flexible, and introduce the FlexiViT model and training procedure in the supervised image classification setting.
We perform all experiments in this section on the public ImageNet-21k dataset \cite{russakovsky2015imagenet}.
We use the base (ViT-B) scale model and \emph{unregularized light2} setting from \cite{vit_augreg}, and train the models for 90 epochs following \cite{liu2022convnet}.

\subsection{Background and notation}\label{sec2:background}

FlexiViT is based on the Vision Transformer (ViT) architecture \cite{dosovitskiy2021vit}.
Here, we briefly describe the ViT architecture and introduce the necessary notation.

Consider an image $x \in \mathbb R^{h\times w \times c}$, where $(h, w, c)$ are the width, height and number of channels respectively.
ViT first tokenizes the input image into a sequence of $s$ patches $x_i \in \mathbb R^{p \times p \times c}$,
where $i \in \{1, \ldots, s\}$.
We refer to this procedure as \textit{patchification} and illustrate it in Figure \ref{fig:teaser}.
The sequence length $s = \floor{h/p}\cdot\floor{w/p}$ is the number of patches (or tokens) after patchification and controls the amount of compute used by the ViT: self-attention scales as $\mathcal{O}(s^2) = \mathcal{O}(h^4) = \mathcal{O}(w^4)$, i.e.\ quartically in terms of image height (or width).

Next, we compute \textit{patch embeddings} $e_i = (e_i^k)_{k=1}^d \in \mathbb R^{d}$ for each patch $x_i$:
$e_i^k = \langle x_i, \omega_{k} \rangle = \operatorname{vec}(x_i)^T \operatorname{vec}(\omega_k)$,
where $\omega_k \in \mathbb{R}^{p \times p \times c}$ are the patch embedding weights,
$\langle \cdot, \cdot\rangle$ denotes the dot product, and $\operatorname{vec}$ is the operation flattening a multi-dimensional array to a vector.
Finally, we add learned position embeddings $\pi_i \in \mathbb{R}^{d}$ to the patch embeddings $t_i = e_i + \pi_i$.
We then pass the sequence of $s$ tokens $t_i$ as input to the Transformer encoder, as illustrated in Figure~\ref{fig:teaser}.

In summary, for a given image size $h \times w$, the patch size $p$ determines the length $s$ of the input sequence to the Transformer model: smaller patch sizes correspond to longer input sequences and slower, more expressive models.
Following \cite{dosovitskiy2021vit}, we denote ViT models as ViT-$\mathcal{S}$/$p$, where $\mathcal{S} \in \{\text{S, M, B, L, \ldots}\}$ is the model scale (small, medium, base, large, \ldots) and $p$ is the patch size.
Note that there are only two parts of the model where the parameter vectors depend on the patch size: the patch embedding weights $\omega_k$ and the position embedding~$\pi$.
In the following sections, we will develop a \textit{flexible} ViT model which works simultaneously for any patch size.

\begin{figure}[t]
  \centering
  \includegraphics[width=1.0\linewidth]{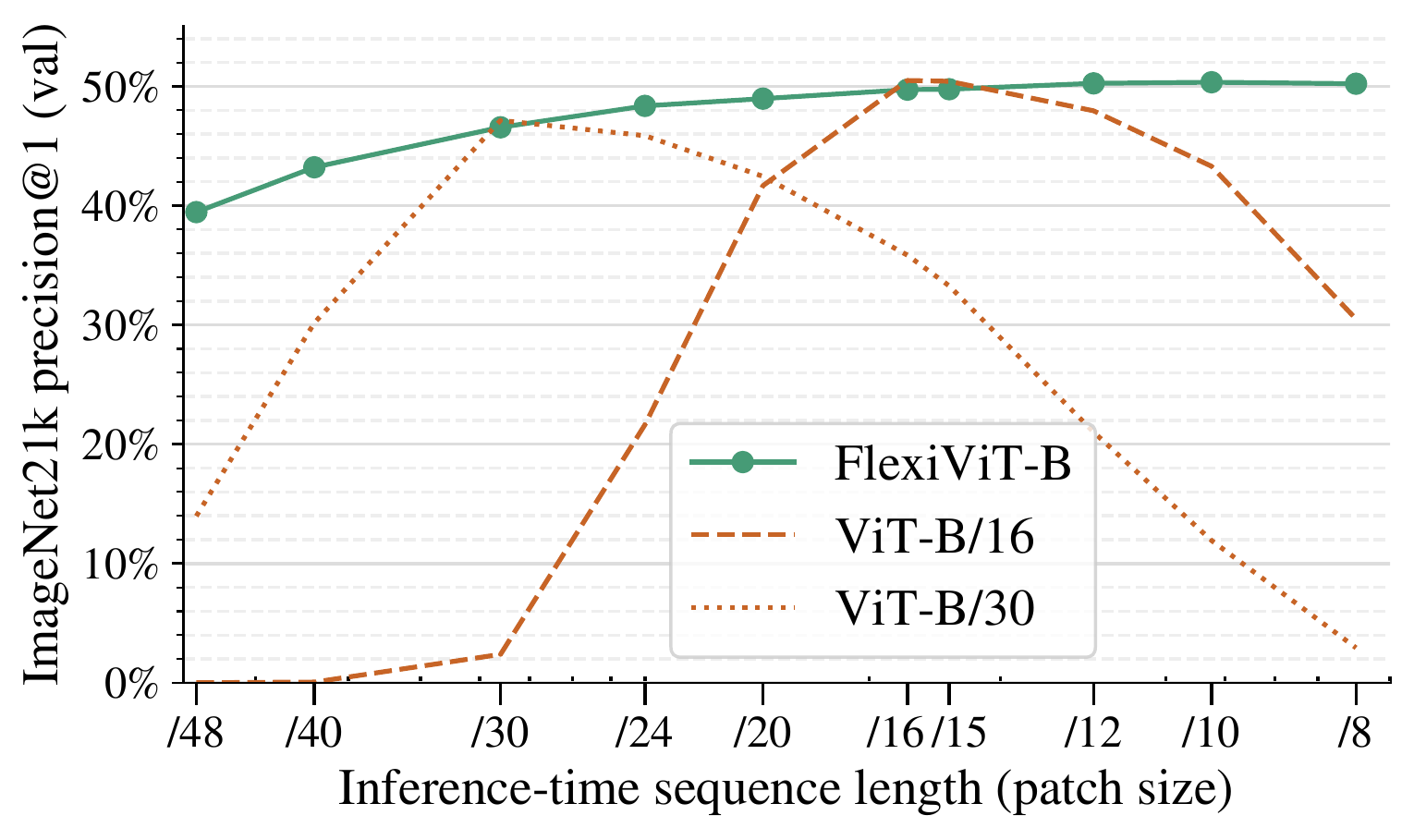}
  \caption{
  \textbf{Standard ViTs are not flexible} in patch size.
  However, FlexiViT can train them to be flexible without loss of performance.
  }\label{fig:inflexible}\vspace{-7pt}
\end{figure}

\begin{algorithm}[t]
\caption{Minimal FlexiViT pseudo-implementation.}\label{alg:code}
\algcomment{
    \textbf{Notes}: Changes to existing code highlighted via violet background.\vspace{-10pt}%
}
\newcommand{\hlbox}[1]{%
  \fboxsep=1.2pt\hspace*{-\fboxsep}\colorbox{blue!10}{\detokenize{#1}}%
}
\lstset{style=mocov3}
\vspace{-3pt}
\begin{lstlisting}[
    language=python,
    escapechar=@,
    label=code:flexivit]
model = ViT(...)
for batch in data:
  @\hlbox{ps = np.random.choice([8, 10, ..., 40, 48])}@
  logits = model(batch["images"]@\hlbox{, (ps, ps)}@)
  # [...] backprop and optimize as usual

class ViT(nn.Module):
  def __call__(self, image@\hlbox{, patchhw}@):
    # Patchify, flexibly:
    w = self.param("w_emb", (@\hlbox{32, 32}@, 3, d))
    b = self.param("b_emb", d)
    @\hlbox{w = resize(w, (*patchhw, 3, d))}@
    x = conv(image, w, strides=patchhw) + b
    # Add flexible position embeddings:
    pe = self.param("posemb", (@\hlbox{7, 7}@, d))
    @\hlbox{pe = resize(pe, (*x.shape[1:3], d))}@
    return TransformerEncoder(...)(x + pe)
\end{lstlisting}\vspace{-5pt}
\end{algorithm}

\subsection{Standard ViTs are not flexible}\label{sec2:inflexible}

We first show that evaluating a standard pre-trained ViT model at different patch sizes yields poor performance.
In order to change the patch size, we simply resize the patch embedding weights $\omega$ and the position embeddings $\pi$ with bilinear interpolation.
\footnote{\href{https://www.tensorflow.org/api_docs/python/tf/image/resize}{\lstinline{tf.image.resize(input, res, method='bilinear')}}}.
For the position embeddings, this resize approach was already proposed in the original ViT paper~\cite{dosovitskiy2021vit} to fine-tune at higher resolution.

The result is shown in Figure~\ref{fig:inflexible}, where we see that the performance of standard ViT models (dashed and dotted lines) rapidly degrades as the inference-time patch size departs from the one used during training.

\subsection{Training flexible ViTs}\label{sec2:flexible}

In Figure~\ref{fig:inflexible} we also show the performance of our FlexiViT-B model (solid line), which matches both ViT-B/16 and ViT-B/30 when evaluated at their training patch sizes, and significantly outperforms them for all other patch sizes.
This model was trained in the same setting as the ViT-B/16 and ViT-B/30 models, except that at each step of training, the patch size was chosen uniformly at random from a set of pre-defined patch sizes.\footnote{We sample patch sizes uniformly in most experiments.
Some early runs used a distribution which slightly favors intermediate patch sizes.
Later experiments showed that the distribution makes little difference (Appendix~\ref{sec:app:distr}). We therefore did not re-run the early experiments.}
In order to do so, two small changes to the model and training code are necessary.

First, the model needs to define an \emph{underlying parameter shape} for $\omega$ and $\pi$.
The learnable parameters are of that shape, and resized on-the-fly as part of the model's forward pass.
We show in Appendix~\ref{sec:app:underlying} that the exact shape of these underlying learnable parameters does not matter much, and we use an underlying size of $32\times32$ for patches and $7\times7$ for position embeddings in all experiments.

Second, to have a large variety of patch sizes that perfectly tile the image, we use an image resolution of 240²\,px, which allows for patch sizes $p \in \{$240, 120, 60, 48, 40, 30, 24, 20, 16, 15, 12, 10, 8, 6, 5, 4, 2, 1$\}$, of which we use all between 48 and 8, inclusive.\footnote{Perfect tiling may not be strictly necessary, and it may be fine to use arbitrary patch sizes and ignore a small border of the image. For simplicity, we focus on the perfect tiling setting.}
At each iteration we sample $p$ from the uniform distribution $\mathcal P$ over these patch sizes.

These are all the changes necessary to \emph{flexify} an existing ViT training procedure. Algorithm~\ref{alg:code} summarizes them.

Note that changing the patch size is related to, but not identical to, changing the image size.
The patch size is purely a change to the model while changing the image size may drastically reduce the available information.
This distinction is further explored in Section~\ref{sec:pi-resize}.

We explore two alternative ways to flexify ViTs in Section~\ref{sec:discuss-alts}: flexible depth and flexible patch stride.
Both of them have merits, but patch size works best.

\subsection{How to resize patch embeddings}
\label{sec:pi-resize}

Consider a patch $x \in \mathbb R^{p\times p}$ of the input image, and the patch embedding weights $\omega \in \mathbb R^{p\times p}$ and let's assume a simple scenario when we are dealing with non-negative values.
If we resize both the patch and the embedding weights with bilinear interpolation, the magnitude of the resulting tokens will differ greatly; for example
$\langle x, \omega \rangle \approx \frac{1}{4} \langle \operatorname{resize}_p^{2p}(x), \operatorname{resize}_p^{2p}(\omega) \rangle$.
We hypothesize that this dramatic change in token norm is part of the reason of ViT's inflexibility, and an inductive bias that hinders learning of a single FlexiViT.
Ideally, as long as there is no loss of information during resizing, the patch embeddings $e_i = \langle x, \omega \rangle$ after resizing both the input $x$ and the embedding $\omega$ should remain the same.

One way to achieve this equality is to normalize the tokens right after their embedding, either explicitly or by using a LayerNorm~\cite{layernorm} module.
However, this approach requires changing the model architecture and is not compatible with existing pre-trained ViTs. Further, it does not exactly preserve the patch embeddings.
As we will show, there is a more principled way of achieving this goal, which is compatible with existing pre-trained models and does not require any architectural change.

\begin{figure}[t]
  \centering
  \includegraphics[width=1.0\linewidth]{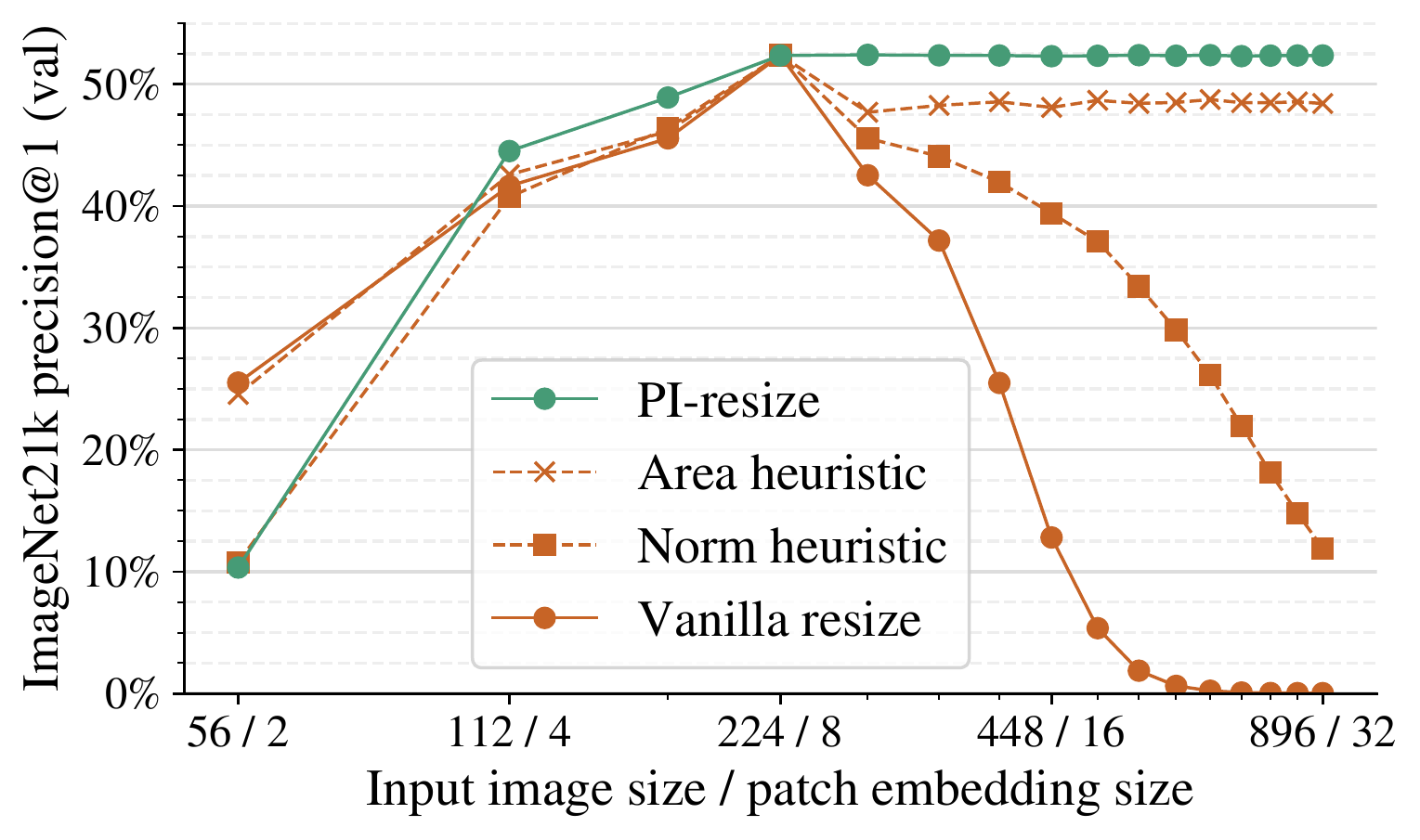}
  \caption{\textbf{Various ways of ``resizing'' ViTs.} We load a ViT-B/8 from~\cite{vit_augreg} trained on $224^2\,\text{px}$, resize patch-embeddings and input images by the same factor, and compute validation accuracy.
  PI-resize is the only method that stays accurate when upscaling.}\label{fig:flexiload}
\end{figure}

First, we note that the linear resize operation introduced in Section~\ref{sec2:inflexible} can be represented by a linear transformation:
\begin{equation}
    \operatorname{resize}_p^{p_*} (o) = B_p^{p_*} \operatorname{vec}(o),
\end{equation}
where $o \in \mathbb{R}^{p\times p}$ is any input, and $B_p^{p_*} \in \mathbb{R}^{{p_*}^2 \times p^2}$.
We resize channels of multi-channel inputs $o$ independently.

Intuitively, we would like to find a new set of patch-embedding weights $\hat \omega$ such that the tokens of the resized patch match the tokens of the original patch.
Formally, we want to solve the optimization problem:
\begin{equation}
\label{eq:pi_resize_optim}
    \hat \omega \in \arg\min_{\hat \omega} \mathbb{E}_{x \sim \mathcal{X}} \left[(\langle x, \omega\rangle - \langle Bx, \hat\omega\rangle)^2 \right],
\end{equation}
where $B = B_{p}^{p_{*}}$ and $\mathcal{X}$ is some distribution over the patches.
In case when we are increasing the patch size, i.e. $p_{*} \ge p$, we can use $\hat\omega = P \omega$ where $P = B (B^T B)^{-1} = (B^T)^+$ is the pseudoinverse of $B^T$:
\begin{equation}
    \langle Bx, \hat\omega\rangle = x^T B^T B (B^T B)^{-1} w = x^T w = \langle x, w \rangle.
\end{equation}
This way we match the patch embeddings \textit{exactly} for all $x$.

In the case of downsampling, i.e. when $p_* < p$, the solution to the optimization problem in Eq.~\eqref{eq:pi_resize_optim} will in general depend on the patch distribution $\mathcal X$.
In Appendix~\ref{sec:app:flexi:pi_resize}, we show that for $\mathcal X = \mathcal N(0, I)$, we recover the pseudoinverse $\hat\omega = P \omega = (B^T)^+ \omega$ as the optimal solution\footnote{
We can also target the patch distribution in the data in place of $\mathcal X$, producing a resize operation which depends on the data.
In our preliminary experiments, we did not observe significant benefits from this approach.
}.
To sum up, we define PI-resize (pseudoinverse resize) as:
\begin{equation}
    \!\!\!\operatorname{PI-resize}_p^{p_*} (w) = \left((B_p^{p_*})^T\right)^{+} \operatorname{vec}(\omega) = P_p^{p_*} \operatorname{vec}(\omega),
\end{equation}
where $P_p^{p_*} \in \mathbb{R}^{{p_*}^2\times p^2}$ is the matrix corresponding to the PI-resize transformation.
The PI-resize operation resizes the patch embedding weights, serving as an \textit{inverse} of the bilinear resize operation.

To experimentally validate the effectiveness of PI-resize and compare it to several alternative heuristics, including standard linear resize, we load a pre-trained ViT-B/8 model from~\cite{vit_augreg} and evaluate it after resizing \emph{both} the image and the model, thus preserving its sequence length $s = (224 / 8)^2 = 784$.
The results, shown in Figure~\ref{fig:flexiload}, demonstrate that PI-resize maintains nearly constant performance when upsampled, and degrades gracefully when downsampling.
None of the heuristics works as well as thoughtful PI-resize across the board.

For completeness, in Appendix~\ref{sec:app:flexi:sup} we experimentally compare the remaining ways of dealing with variable patch sizes when one does not care about maintaining model compatibility.
These methods include fixed normalization, LayerNorm, and learning separate parameters $\omega$ for each patch size.
Adding a LayerNorm works best, but otherwise, PI-resize and bilinear resize are among the best techniques.

\subsection{Connection to knowledge distillation}\label{sec:distill}


\begin{figure}[t]
  \centering
  \includegraphics[width=1.0\linewidth]{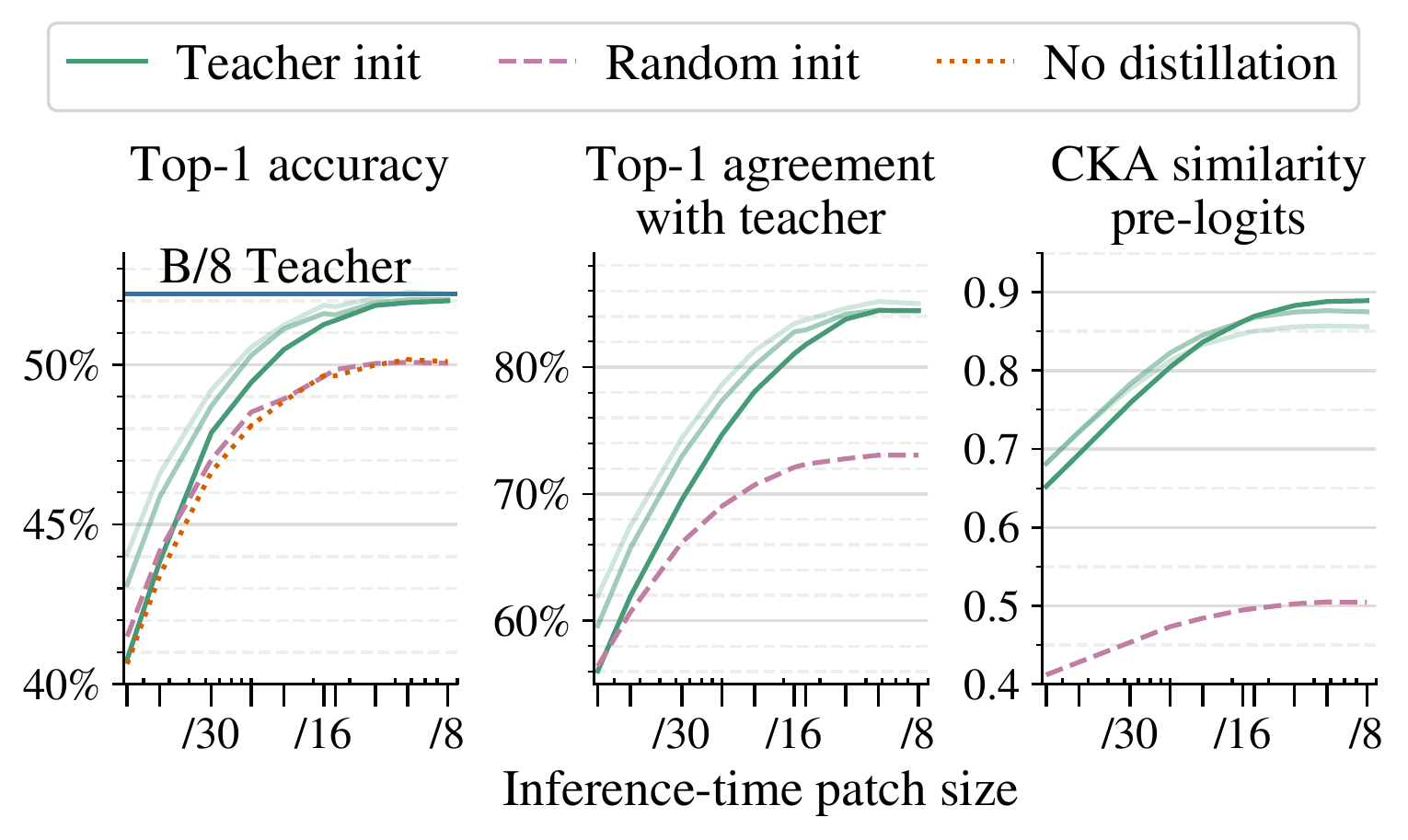}
  \caption{\textbf{The effect of initialization} when distilling to FlexiViT.}
  \label{fig:distill}\vspace{-10pt}
\end{figure}

Knowledge distillation~\cite{hinton2015distilling} is a popular technique, where a typically smaller \emph{student} model is trained to mimic the predictions of a typically larger \emph{teacher} model.
This can significantly improve the performance of the student model compared to standard label-supervised training~\cite{cho2019efficacy,xie2020self,beyer2022knowledge}.

It was recently shown that knowledge distillation corresponds to a much more challenging optimization problem than standard supervised training~\cite{stanton2021does,beyer2022knowledge}, and that initializing the student close to the teacher simplifies alleviates this~\cite{stanton2021does}.
Unfortunately, this solution is impractical since the teacher usually has a different (larger) architecture than the student~\cite{beyer2022knowledge}.
However, with FlexiViT, we \emph{can} initialize a student FlexiViT with the weights of a powerful ViT teacher and significantly improve distillation performance.

Unless otherwise stated, the model we use for the remaining experiments in this paper is a FlexiViT-B initialized and distilled from the powerful ViT-B/8 model of~\cite{vit_augreg}.
At initialization, we PI-resize the teacher's patch embedding weights to $32\times32$, and bilinearly resample its position embeddings to $7\times7$.
We then train the student model following the FunMatch~\cite{beyer2022knowledge} approach, minimizing the KL-divergence between the predictions of the teacher and the student FlexiViT with a randomized patch size:
\begin{equation}
    \label{eq:distillation}
    \mathbb{E}_{x\in\mathcal D} \mathbb{E}_{p \sim \mathcal P} 
        \operatorname{KL}\left(
            f_{\text{FlexiViT}}(x, p) \vert\vert 
            f_{\text{ViT-B/8}}(x)
        \right),
\end{equation}
where $f_{\text{FlexiViT}}(x, p)$ is the distribution over classes for the FlexiViT model on an input $x$ with patch size $p$,
$f_{\text{ViT-B/8}}(x)$ is the predictive distribution of the teacher on the \emph{exact same} input,
$\mathcal{D}$ is the training data distribution with random flips, crops, and mixup, and $\mathcal P$ is the distribution over patch sizes used for training the FlexiViT model.

Figure~\ref{fig:distill} compares the effect of distilling using teacher initialization to random initialization and to supervised training from labels.
The comparison was performed for 90 epochs and shows considerable benefits of this unique initialization capability of FlexiViT.
Since distillation needs patience~\cite{beyer2022knowledge, deit}, we additionally run for 300 and 1000 epochs, shown as pale green curves in the figure.
FlexiViT matches the teacher's performance at small patch sizes, and teacher initialization provide a large improvement in accuracy at the largest patch sizes.
In the following sections, we use the FlexiViT that was trained for 300 epochs and train two fixed ViT-B/30 and ViT-B/16 models in the same setting (including the initialization) as baselines.

\subsection{FlexiViT's internal representation}

Does FlexiViT process inputs with different patch sizes in similar ways?
We investigate this by analyzing the model's internal representations.
We apply minibatch centered kernel alignment (CKA)~\cite{nguyen2021do,kornblith2019similarity,cortes2012algorithms}, a widely-used approach for comparing representations within and across neural networks.
For visualization purposes, we apply an arccosine transform to transform CKA/cosine similarity to proper metrics~\cite{williams2021generalized} and then perform t-SNE.

\begin{figure}[t]
  \centering
  \includegraphics[width=1.0\linewidth]{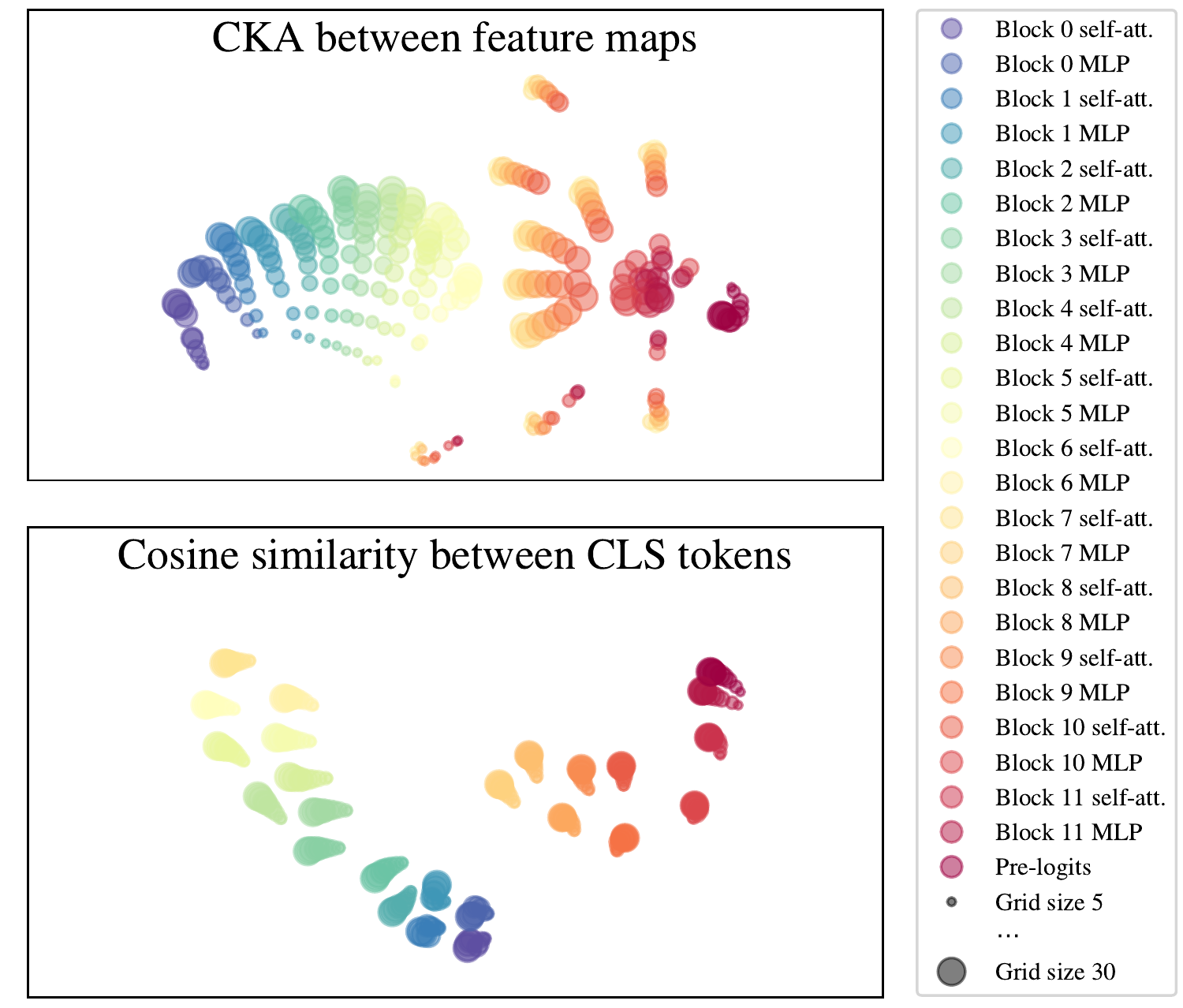}\vspace{-10pt}
  \caption{\textbf{t-SNE visualizations} of intermediate representations of network layers across different grid sizes. Colors reflect different layers; dot sizes reflect different grid sizes.}
  \label{fig:similarity}\vspace{-10pt}
\end{figure}

Results are shown in Figure~\ref{fig:similarity}.
Feature map representations are similar across grid sizes from the first layer until the MLP sublayer of block 6.
At the MLP sublayer of block 6, layer representations diverge, before converging again at the final block.
By contrast, CLS token representations remain aligned across grid sizes.
Thus, although internal representations of a substantial portion of FlexiViT differ by grid size, output representations are generally aligned.

\begin{figure*}[t]
  \centering
  \includegraphics[width=1.0\linewidth]{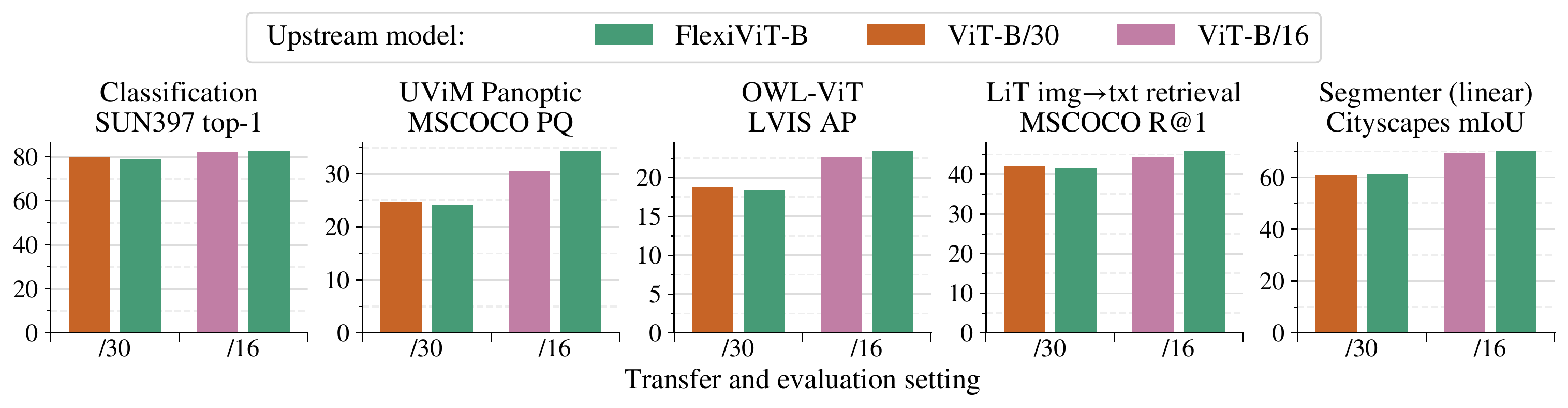}
  \caption{\textbf{Using a pre-trained FlexiViT.}
  We use the flexibly pre-trained FlexiViT-B model in a diverse set of downstream computer vision tasks at two patch sizes, and verify that it performs the same or better than a plain (inflexible) ViT model pre-trained at that patch size.
  These results indicate that flexibly pre-training a single ViT may be preferrable than pre-training several fixed ViTs.}
  \label{fig:using_flexivit}
\end{figure*}

\section{Using pre-trained FlexiViTs}\label{sec:using}

We have shown that ViTs can be trained flexibly without significant loss of \emph{upstream} performance. Next, we verify that pre-trained FlexiViTs are still comparable to individual fixed patch-size ViTs when transferred to other tasks.
We check this by transferring the single pre-trained FlexiViT with its patch size fixed to either $16^2$ or to $30^2$ during transfer. We compare FlexiViT to ViT-B/16 and a ViT-B/30 models that were pre-trained using the same distillation setup as FlexiViT (Section~\ref{sec:distill}), but with a fixed patch size.
We perform this transfer on the following set of diverse tasks.

For each task, we provide more details along with many more results, all with the same take-away, in Appendix~\ref{sec:app:transfer}.

\PAR{Classification} We fine-tune on small- (Pet~\cite{parkhi2012cats}, Flowers~\cite{nilsback2008automated}) and medium-scale (CIFAR10, CIFAR100~\cite{krizhevsky2009learning}, Food101~\cite{bossard2014food}, SUN397~\cite{xiao2010sun}) classification datasets following the setup of~\cite{dosovitskiy2021vit} at $240^2$\,px resolution.

\PAR{Locked-image Tuning (LiT)}
We follow~\cite{lit} to train a text model contrastively~\cite{clip,align} for the frozen FlexiViT, which we evaluate in terms of 0-shot classification and retrieval.

\PAR{Open-vocabulary detection}
We test the transferability of FlexiViT to object detection using OWL-ViT~\cite{minderer2022simple}, an open-vocabulary object detector based on image-text models such as LiT or CLIP~\cite{clip}.
We evaluate its zero-shot open-vocabulary detection performance on LVIS~\cite{gupta2019lvis}.

\PAR{Panoptic segmentation} The Universal Vision Model (UViM) is a general-purpose modeling approach for vision~\cite{kolesnikov2022uvim}.
We train UViM on the COCO panoptic segmentation dataset~\cite{lin2014microsoft,kirillov2019panoptic} and use FlexiViT as initialization for the image encoder in UViM.

\PAR{Semantic segmentation} We transfer to semantic segmentation following Segmenter's linear decoder setup~\cite{strudel2021segmenter}.
We report mean IoU for single scale evaluation and evaluate on Cityscapes~\cite{cityscapes} and ADE-20k~\cite{ade20k}.

\subsection{Results}\label{sec:using:results}

\begin{figure}
    \centering
    \includegraphics[width=1.0\linewidth]{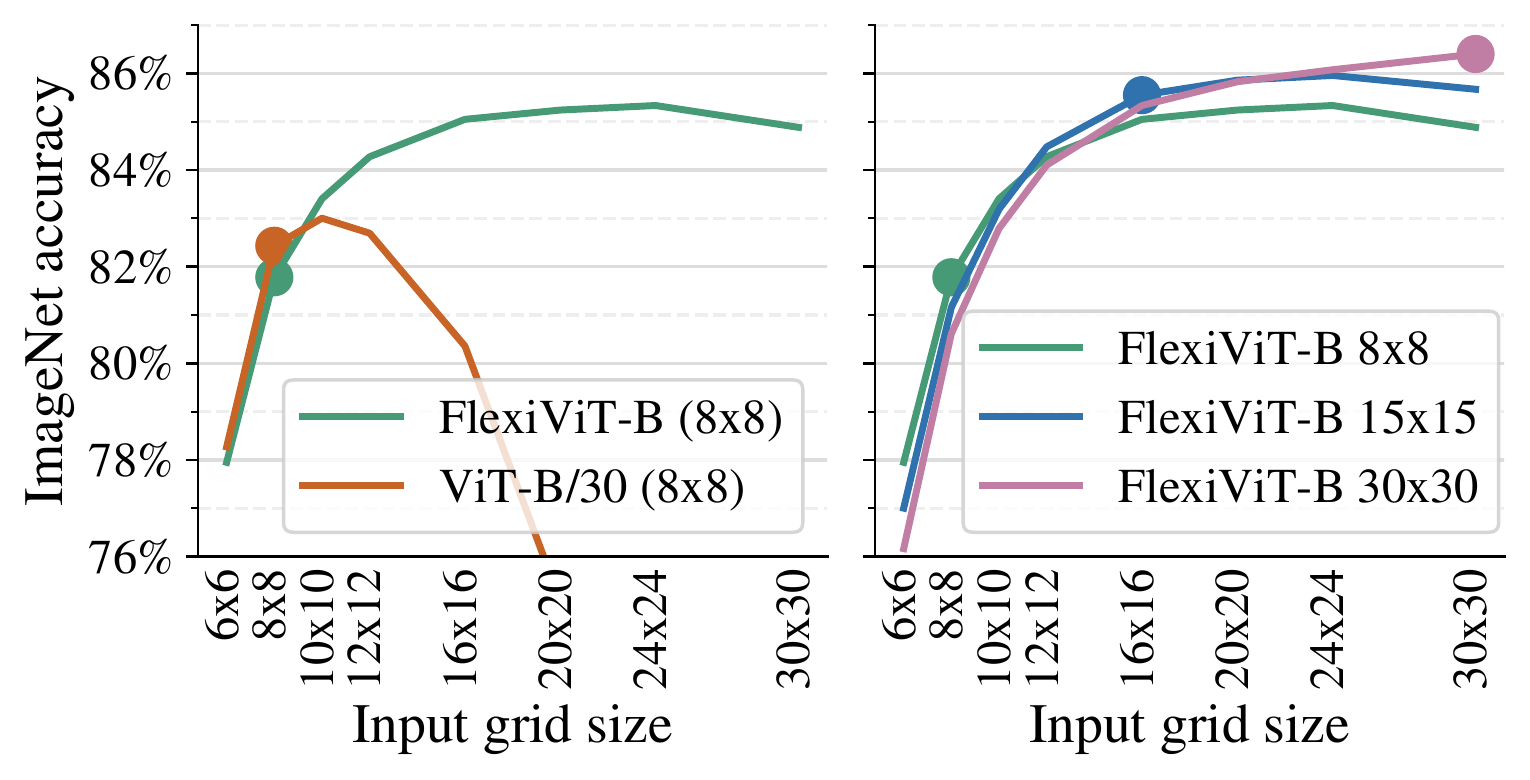}
    \caption{\textbf{Fast transfer.} FlexiViT can be cheaply finetuned at small sequence length and used at test time with much longer sequence to achieve higher performance.
    \textit{(left)} FlexiViT-B and ViT-B/30 models finetuned at grid size 8x8 (indicated by dots) and evaluated at other grid sizes. The standard ViT model's accuracy quickly deteriorates, while FlexiViT demonstrates large performance boost with increased grid size.
    \textit{(right)} A single FlexiViT-B model finetuned at three different grid sizes (indicated by dots) and evaluated at various grid sizes.}
    \label{fig:fast_transfer}
\end{figure}

The results of these transfer experiments are summarized in Figure~\ref{fig:using_flexivit}.
Across the diverse set of tasks, a single FlexiViT model roughly matches the two fixed ViT models, barely lagging behind at large patch size and leading to a small or significant improvement at smaller patch size.

These results confirm that there is no significant downside in using a pre-trained FlexiViT, as opposed to pre-training multiple ViTs for different patch sizes.

\begin{figure*}[t]
\centering
\begin{minipage}[t]{.24\textwidth}
  \includegraphics[width=1.0\columnwidth]{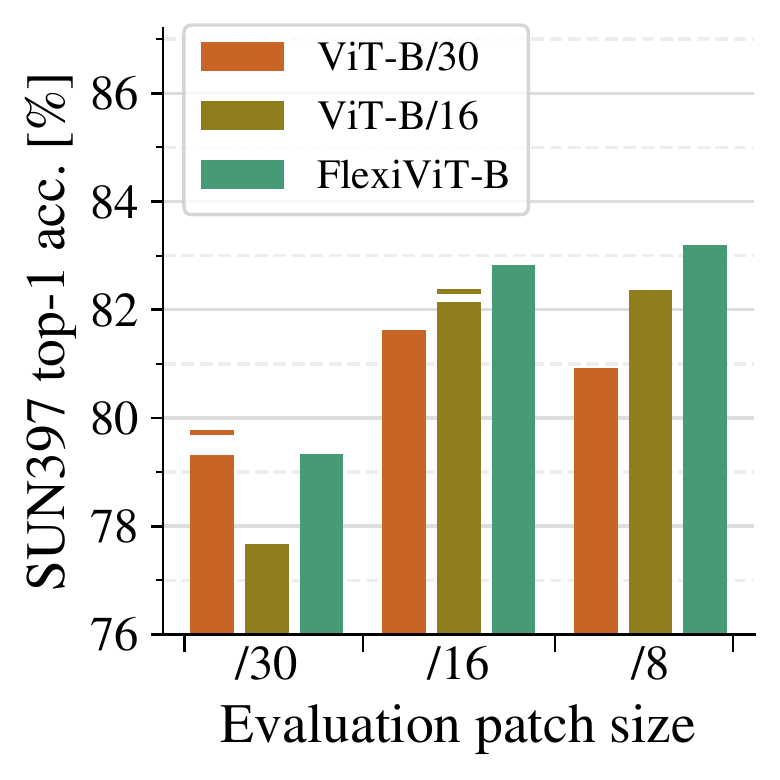}
  \caption{Flexified transfer of a flexible model works best, but even inflexible models can be flexified during transfer.
  The lines represent fixed transfer of fixed models, for reference.}\label{fig:flexify:transfer}
\end{minipage}\hfill
\begin{minipage}[t]{.24\textwidth}
  \includegraphics[width=1.0\columnwidth]{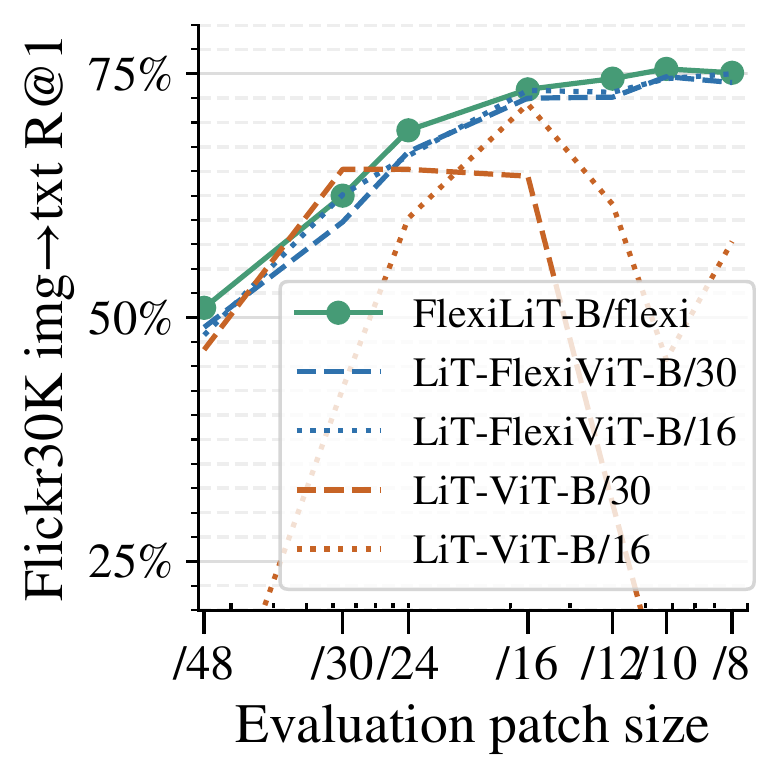}
  \caption{Flexified LiT transfer works the best, while fixed LiT transfer of FlexiViT with a single patch size (30 or 16) performs surprisingly well, similarly to Section~\ref{sec:using:fasttrans}.}\label{fig:flexilit}
\end{minipage}\hfill
\begin{minipage}[t]{.24\textwidth}
  \includegraphics[width=1.0\columnwidth]{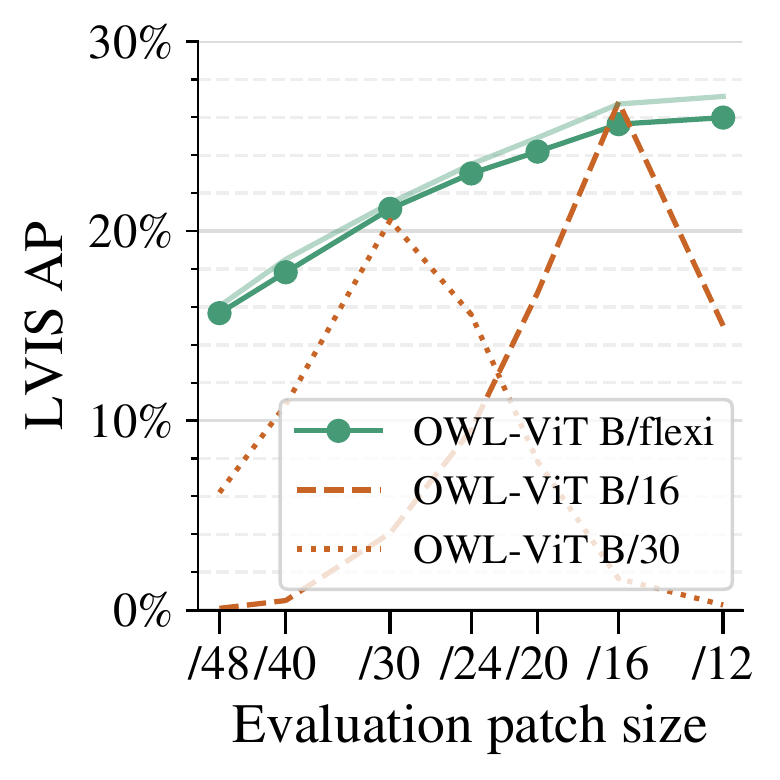}
  \caption{Flexified open-voca-bulary detection. Image-text mo-dels are transferred to detection~\cite{minderer2022simple}.
  Dark green shows transfer of a fixed, pale green of a flexible model.}\label{fig:flexiowl}
\end{minipage}\hfill
\begin{minipage}[t]{.24\textwidth}
  \includegraphics[width=1.0\columnwidth]{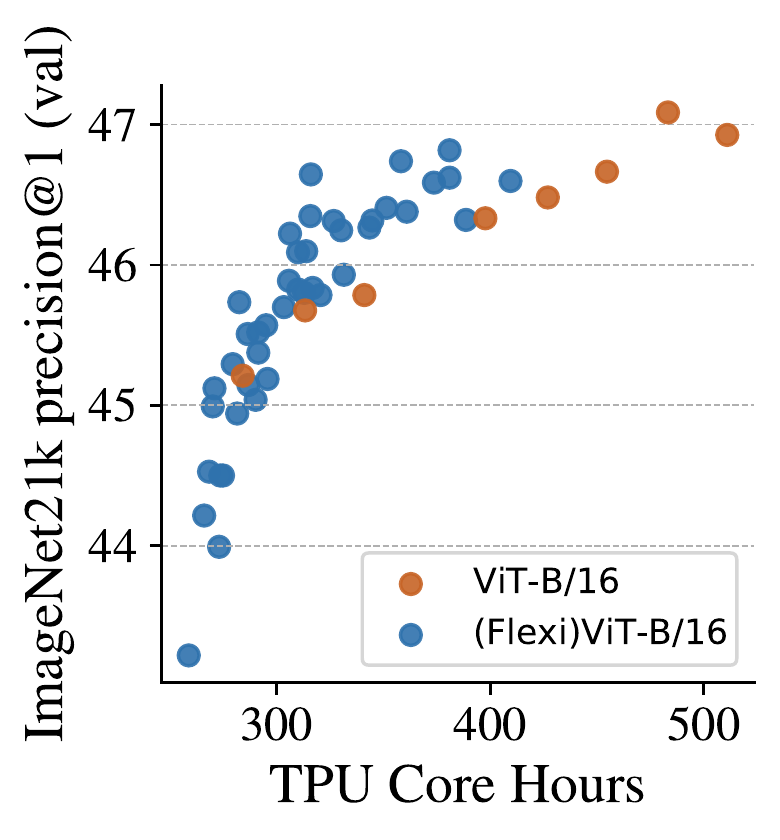}
  \caption{Flexible patch sizes to accelerate pre-training.
  The ViT-B/16 models are trained for different epochs, (Flexi)ViT-B/16 are different curricula.
  Evaluation at patch size 16.
  }\label{fig:accelerate_pretrain}
\end{minipage}
\end{figure*}

\subsection{Resource-efficient transfer via flexibility}\label{sec:using:fasttrans}

FlexiViT enables a new way of making transfer learning more resource efficient, saving accelerator memory and compute. This is possible because, surprisingly, \textit{flexibility is largely retained even after transfer at a fixed patch size}.
We can therefore perform transfer training cheaply with large input patches (small input grid), but later deploy the resulting model using small patch sizes (large input grid).
We preform experiments by transferring a FlexiViT-B model (pretrained on ImageNet-21k with distillation) to the ImageNet-1k dataset, and use a similarly pretrained fixed ViT-B/30 model as the baseline.  The pretrained FlexiViT works well at larger grid sizes even after fixed-size transfer.
For example, we can perform relatively cheap finetuning at $8\times8$ grid size. When evaluated at  $8\times8$ grid size, the model achieves 81.8\% accuracy, but when evaluated at the $24\times24$ grid size, it achieves 85.3\% top-1 accuracy gaining 3.5\% accuracy at no additional training cost (Figure~\ref{fig:fast_transfer}).
More details on the finetuning setup can be found in the Appendix~\ref{sec:app:fast-i1k-transfer}.

\section{Flexifying existing training setups}\label{sec:flexifying}

So far, we have focused on flexifying models during pre-training. We now show that existing pre-trained models can also be flexified during transfer to downstream tasks. Below, we flexify a diverse set of existing training setups.

\subsection{Transfer learning}

We use the same set of 6 transfer datasets from Section~\ref{sec:using}, with the same settings.
We again show the results for SUN397 in Figure~\ref{fig:flexify:transfer} and all other datasets in Appendix~\ref{sec:app:transfer}.
The difference is that we now also randomize the patch size during transfer, and evaluate the single resulting model at different patch sizes (x-axis, three groups of bars).

Flexible transfer of FlexiViT works best, but flexifying a fixed model during transfer also works surprisingly well, considering the very short training and low learning rate used for transfer.
The baseline of a fixed-size model transferred at a fixed patch size and evaluated at that same size is indicated by a small horizontal line.

\subsection{Multimodal image-text training}

Next, we discuss two ways to flexify multimodal image-text training: FlexiLiT and FlexiCLIP. In FlexiLiT, we train a text tower to produce text embeddings that align well with visual embeddings from \textit{various} patch sizes (B/flexi). LiT baselines with direct use of either FlexiViT models at \textit{fixed} resolutions, or ViT models are provided.
Figure~\ref{fig:flexilit} shows zero-shot image to text retrieval results on the Flickr30k~\cite{flickr30k} dataset.
FlexiLiT-B/flexi performs the best on average, while LiT with FlexiViT-B/30 and FlexiViT-B/16 both get very close results. 
Flexification additionally provides the possibility of fast transfer as discussed in Section~\ref{sec:using:fasttrans}.
The LiT-ViT baselines shown in Figure~\ref{fig:flexilit} match FlexiLiT on the sequence length it has been trained for, but performance drops quickly when using a different sequence length during inference.
We observe similar conclusions with a from-scratch image-text training setup, i.e. FlexiCLIP (see Appendix~\ref{sec:app:flexify_lit} for more results).

\subsection{Open-vocabulary detection}\label{sec:flexifying_owl}
Beyond image-level tasks, we find that flexification also works for object detection training. We modify the training of OWL-ViT to introduce flexible patch sizes as described in Algorithm~\ref{alg:code}. Similar to classification, flexible OWL-ViT detection models perform close to or better than fixed-size models at any patch size during inference (Figure~\ref{fig:flexiowl}). In addition, we find that for detection, the optimal patch size is not necessarily the smallest. When evaluated on a set of 35 detection datasets~\cite{li2022elevater}, inference-time tuning of the patch size leads to improved results over evaluation at the smallest patch size (Appendix~\ref{sec:app:transfer}). This makes flexification especially valuable for detection.
\subsection{Training times and flexification}\label{sec:accelerate_pretrain}
Besides having a flexible model, one can use FlexiViT's machinery to pre-train fixed ViTs faster.
In this case, we specify a \emph{curriculum}: a sequence $(p_k)_{k=1}^K$ of probability distributions over the patch sizes along with a mapping $c:\mathbb{N}\to[K]$ that identifies which distribution $p_k$ to use at training step $t$.
For example, if the desired patch size is $16\times 16$, the last probability distribution in the sequence $p_K$ would place its entire mass on said patch size.
A multitude of curricula can be designed, see Appendix~\ref{sec:app:accelerate_pretraining}.
In Figure~\ref{fig:accelerate_pretrain} we show that in general training with a patch size curriculum leads to better performance per compute budget than standard training as we vary the training length.

\section{Analyzing FlexiViTs}

\begin{figure}[t]
  \centering
  \includegraphics[width=1.0\linewidth]{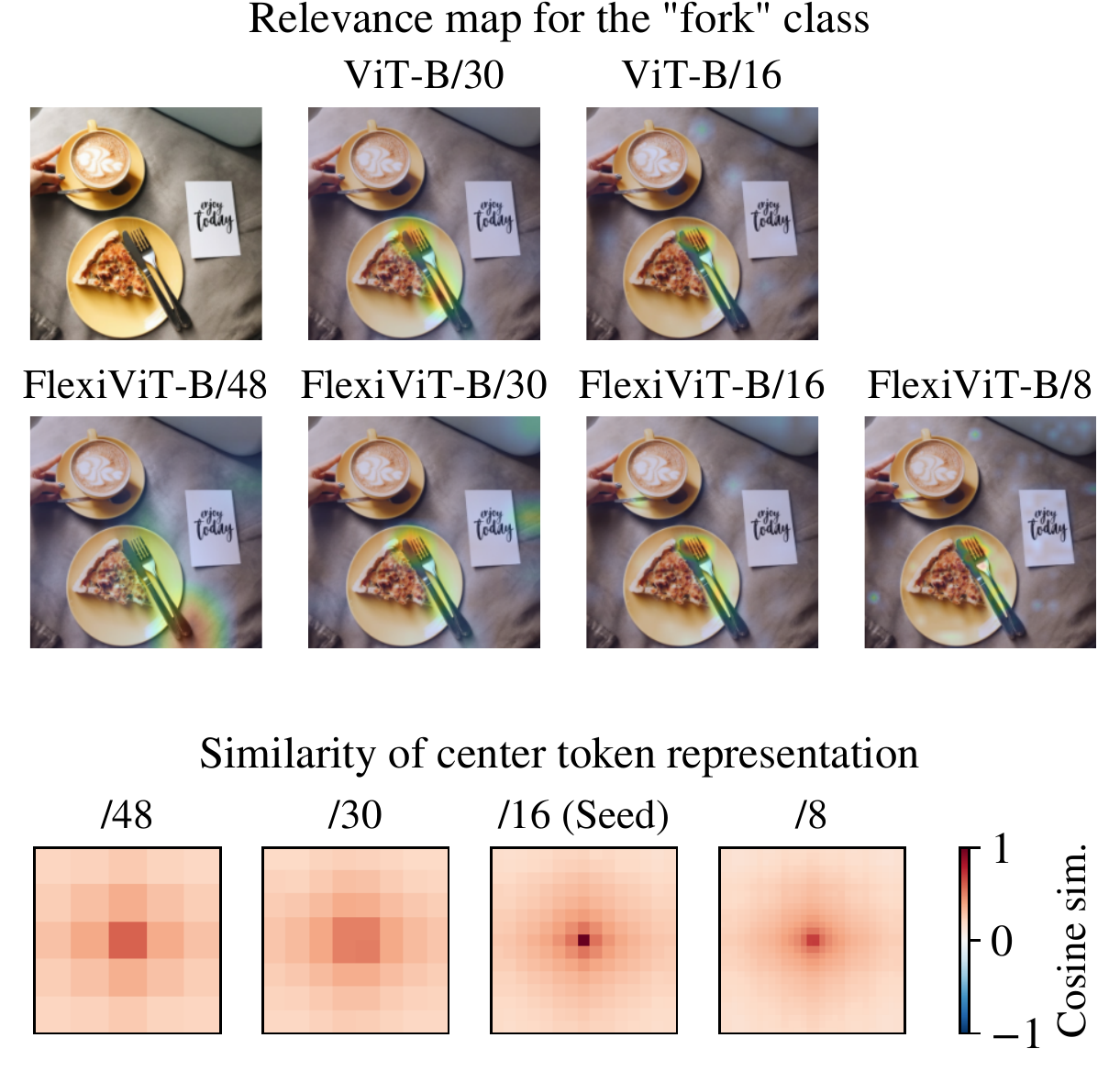}\vspace{-10pt}
  \caption{\textbf{Analysis of FlexiViT attention and token representations across scales.} \textit{Top:} Attention relevance (as in \cite{Chefer_2021_CVPR}) can significantly change at different patch sizes. For example, FlexiViT-B/48 and FlexiViT-B/8 consider different areas of the input most relevant for class 'fork'. See Appendix~\ref{sec:app:attention_relevances} for more examples. \textit{Bottom:} Cosine similarity between a seed token representation at the center of the feature map of FlexiViT-B at patch size 16 and representations of tokens at other patch sizes. Representations are taken from block 6 and averaged across our I21K validation set. See Appendix~\ref{sec:app:cosine_scales} for similar plots for other blocks and patch sizes.}
  \label{fig:att_token}\vspace{-10pt}
\end{figure}

\PAR{Attention relevance patterns across scales}
We find that decreasing the patch size results in attention relevance \cite{Chefer_2021_CVPR} to concentrate into a larger number of smaller areas throughout the image. In Figure~\ref{fig:att_token} (top) we observe that attention can significantly change at different scales.

\PAR{Relation of token representations across scales}
\label{sec:cosine_scales}
As we decrease FlexiViT's patch size, each token ``splits'' into multiple tokens. A natural question is how token representations at larger patch sizes relate to token representations at smaller patch size. To answer this question, we measure cosine similarity between the representation of a ``seed'' token at the center of a feature map at one patch size and representations of other tokens at the same and different patch sizes. As shown in Figure~\ref{fig:att_token} (bottom), we are indeed able to find correspondences between tokens across scales.

\PAR{Ensembling}
We explored whether it is possible to improve prediction accuracy by ensembling the predictions of the same FlexiViT at multiple scales.
We find that, in terms of total compute spent, it is nearly always better to run a single FlexiViT at that compute budget than to ensemble multiple smaller ones. Full results are provided in Appendix~\ref{sec:app:ensemble}.

\PAR{Shape or texture bias} ViT's bias towards using shape or texture features~\cite{geirhos2018imagenet} has been shown to largely depend on its patch size~\cite{bhojanapalli2021understanding}.
In Appendix~\ref{sec:app:shapebias}, we show that FlexiViT evaluated at each patch size has a similar texture bias to a ViT trained and evaluated at that same patch size.


\PAR{Model and dataset size}
Throughout the paper, we focus on FlexiViT models of the \emph{base} size (-B) trained on 12\,M images.
In order to validate that neither of these two settings are required, we train FlexiViT-{S,B,L} models on ImageNet-1k (1.2\,M images) using the ImageNet-1k DeiT\,III model~\cite{deit3} as teacher.
We can see in Fig~\ref{fig:i1k} that a single FlexiViT-L model matches or outperforms all three DeiT\,III models and EfficientNetV2.
However, there is still a point at which it becomes more effective to change model width than patch size.
Numerical results and evaluation on ImageNet-ReaL/v2/A/R~\cite{imagenet_real,imagenet_v2,imagenet_a,imagenet_r} are in Appendix~\ref{sec:app:i1k}, a version of Fig.~\ref{fig:i1k} using GFLOPs~\cite{misnomer} is in Appendix~\ref{sec:app:i1k_gflops}.

\section{Discussion of alternatives}\label{sec:discuss-alts}

Changing the input patch size is not the only way to trade off sequence length and compute in ViTs.
We explore two alternatives in our core setup: distillation on ImageNet-21k.

\begin{figure}[t]
  \centering
  \vspace{-14pt}
  \includegraphics[width=1.0\linewidth]{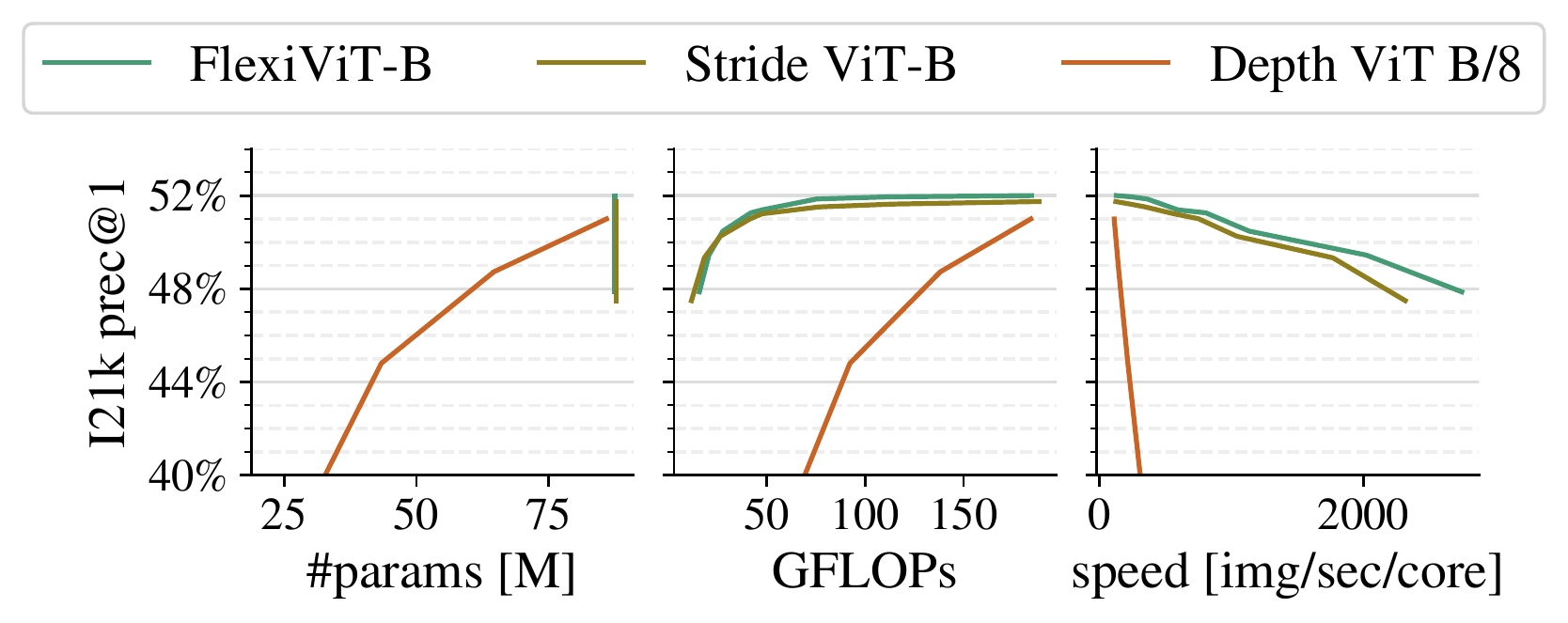}
  \vspace{-10pt}
  \caption{\textbf{Varying stride and depth as alternatives to FlexiViT}}
  \label{fig:flexi_stride_depth}
\end{figure}

\PAR{Varying patch embedding stride} 
One alternative is to fix the patch size and change its sampling stride, i.e.\ extract overlapping patches to increase sequence length.
Intuitively, the advantage of this approach is that the intrinsic patch size is fixed and we avoid any special care when computing patch embeddings.
Results in Figure~\ref{fig:flexi_stride_depth} suggest varying the stride works almost as well, only slightly lagging behind our baseline.

\PAR{Varying model depth}
Another alternative is adding flexibility in terms of depth, i.e.\ number of layers.
Depth pruning has been explored in the context of NLP~\cite{fan2020reducing, schuster2021consistent} and more recently also for ViTs~\cite{fang2022width}.
Depth pruning differs fundamentally from FlexiViT: it scales linearly in depth, uses a subset of parameters, and allows progresively refining a prediction.
We randomize the depth by attaching the shared head to various intermediate layers. We also tried separate heads, which worked worse.
The results in Figure~\ref{fig:flexi_stride_depth} show that FlexiViT provides a significantly better compute-accuracy tradeoff than depth pruning.


\section{Conclusion}
FlexiViT is a simple and efficient way of trading off compute and predictive performance with a single model, enabled by the unique patch embedding strategy of ViTs.
FlexiViT can be used to significantly reduce pre-training costs by only training a single model for all scales at once, and performs well at a variety of downstream tasks.
There are many exciting directions for future work, and we hope that our results inspire the community to explore additional creative applications of patchification.
  
\section{Acknowledgments}

We thank Daniel Keysers for good feedback on a draft of the paper, Geoffrey Hinton for a nudge to pursue this project early on, and our respective teams at Google for encouraging creative and independent research.

We would also like to thank the following artists for making their photographs (used for visualizations) freely available through \url{unsplash.com}:
Tanya Patrikeyeva, Markus Spiske, Matheus Bardemaker, Alexandru Sofronie, Chris Smith, Kajetan Sumila, Julee Juu, Mike Erskine, Piermanuele Sberni, Feyza Yıldırım and Sixteen Miles Out.

We thank the anonymous reviewers for good feedback that further improves our paper.

Finally, the raccoon picture used as background in Figure~\ref{fig:app:owl_elevater} is from \url{publicdomainvectors.org}.

{\small
\bibliographystyle{ieee_fullname}
\bibliography{egbib}
}

\clearpage{}
\appendix
  
Besides providing more details and results on various sections as mentioned in the main paper, we also provide full numerical (tabular) results of all figures in Appendix~\ref{sec:app:tables} in order to facilitate reproduction/comparison in future work.

Finally, more details about the alternative ways of flexibility discussed in Section~\ref{sec:discuss-alts} are provided in Appendix~\ref{sec:app:alternatives}.

\section{More details on flexible patch-sizes}\label{sec:app:flexisup}

In this section, we further elaborate on many details of flexible patch-sizes.
We provide results for alternative ways of dealing with flexible patch-sizes when one does not care about preserving model architecture in Appendix~\ref{sec:app:flexi:sup}.
We provide a detailed derivation of PI-resize in Appendix~\ref{sec:app:flexi:pi_resize}, and show some PI-resize matrices in Appendix~\ref{sec:app:flexi:pi_matrix}.
We further show some visualizations of patch-embedding weights, both raw and resized, in Appendix~\ref{sec:app:flexi:patchembs}

\subsection{Alternatives for dealing with flexible patch-size}\label{sec:app:flexi:sup}

\begin{figure}[t]
  \centering
  \includegraphics[width=1.0\linewidth]{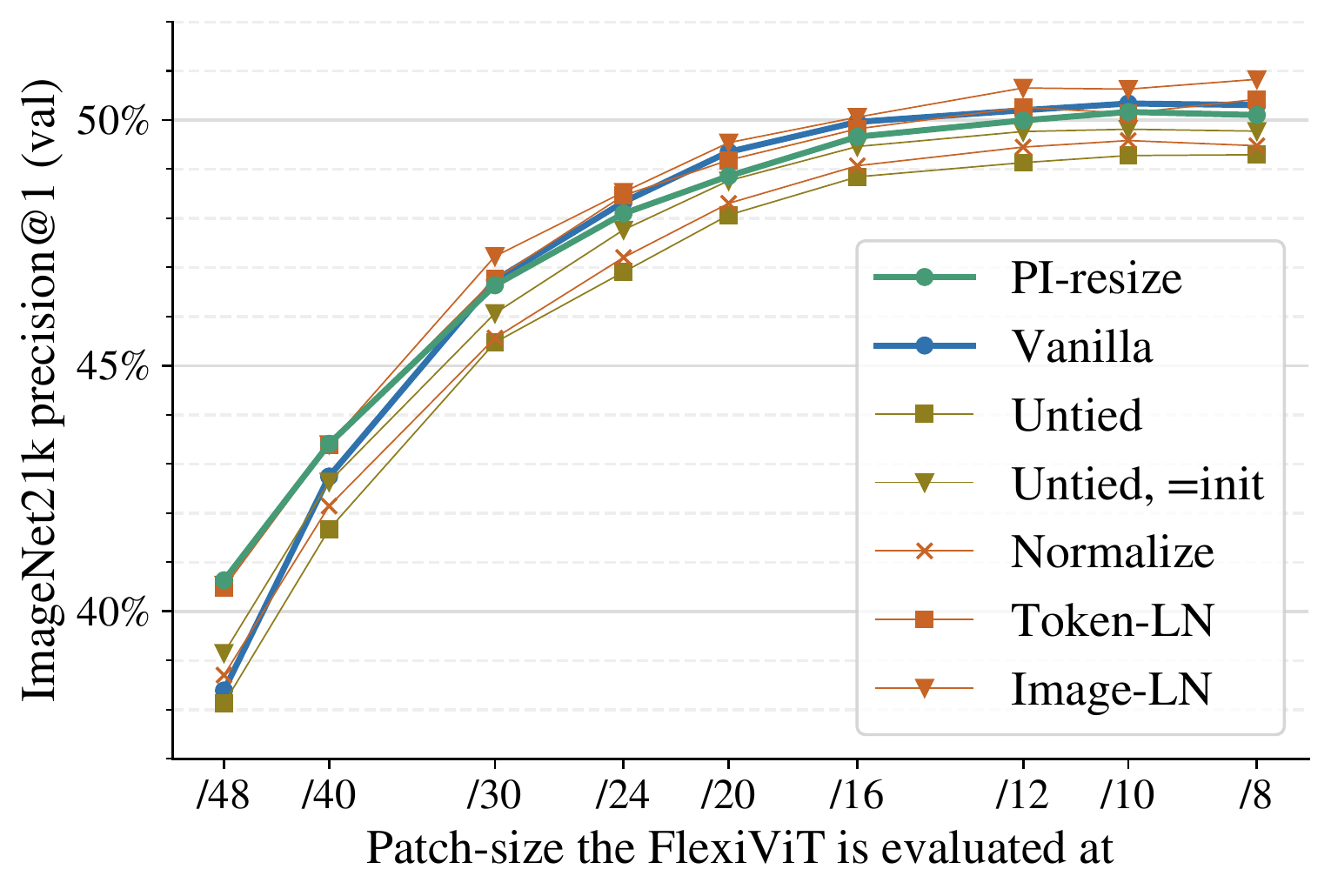}
  \caption{\textbf{Options for dealing with patch-embeddings}. Supervised training from-scratch on ImageNet-21k. Vanilla (bilinear) is the simplest method and works well. PI-resize further improves the large-patch case and provides other advantages (see text). Untied means learning separate patch-embedding kernels for each size, and does not work too well. Normalize, Token-LN, and Image-LN all but require modifications to the model making it incompatible with standard ViT.}
  \label{fig:app:flexisup}
\end{figure}

Besides bilinear resizing (called \textbf{Vanilla}) or \textbf{PI-resizing} the patch-embedding weights to deal with variable patch-sizes, there are a few other alternatives which we discuss and compare here.

\PAR{Untied} weights for each size, i.e.\ having separate trainable parameter buffers for each patch-size.

\PAR{Untied, =init} is the same as above, but initializing all patch embedding weights to the same (PI-resized) values as a reference initialization.
In this setting, the model is still compatible with standard ViT models at initialization time, however, the parameters can, and do, diverge during training, resulting in a non-standard ViT architecture.

\PAR{Normalize} simply l2-normalizes the tokens computed by the patch-embedding individually to unit-norm. This solves any norm-related issues in a simple, parameter-free way, but is incompatible with pre-trained standard ViT models.

\PAR{Token-LN and Image-LN} add a LayerNorm~\cite{layernorm} right after the patch-embedding and differ only in which axis they perform the normalization.
Again, this solves the norm-related issues, but does add learnable parameters and is incompatible with pre-trained standard ViT models.

A comparison of all these variants is performed in the label-supervised training setup on ImageNet-21k following \cite{vit_augreg}, but training for 90 epochs.
The result, presented in Figure~\ref{fig:app:flexisup}, indicates that plain resizing and PI-resizing are among the best solutions, but have the added benefit of resulting in standard ViT models.
Furthermore, not visible in this figure, both \emph{Untied} variants displayed slightly unstable training curves in the first half of training, while all other variants train smoothly.

\subsection{PI-resize derivation}\label{sec:app:flexi:pi_resize}

We can rewrite the objective function in Eq. \eqref{eq:pi_resize_optim} as follows: 
\begin{equation}
    \begin{split}
        &\mathbb{E}_{x \sim \mathcal{X}} \left[(\langle x, \omega\rangle - \langle Bx, \hat\omega\rangle)^2 \right] = \\
        &\mathbb{E}_{x \sim \mathcal{X}} \left[(x^T (\omega - B^T \hat\omega))^2 \right] = \\
        &\mathbb{E}_{x \sim \mathcal{X}} \left[((\omega - B^T \hat\omega)^T x)(x^T (\omega - B^T \hat\omega)) \right] = \\
        & (\omega - B^T \hat\omega)^T \mathbb{E}_{x \sim \mathcal{X}} \left[xx^T \right]  (\omega - B^T \hat\omega) = \\
        &\|\omega - B^T \hat\omega\|_{\Sigma}^2, 
    \end{split}
\end{equation}
where $\|v\|_\Sigma^2 = v^T \Sigma v$ and $\Sigma = \mathbb{E}_{x \sim \mathcal{X}} xx^T$ is the (uncentered) covariance matrix of $\mathcal{X}$.
In case when $\mathcal{X} = \mathcal{N}(0, I)$, we recover the standard euclidean norm $\|\omega - B^T \hat\omega\|^2$.

Finally, we note that the pseudoinverse matrix recovers a least squares solution to a linear system of equations:
\begin{equation}
    (B^T)^{+} \omega \in \arg\min_{\hat \omega} \|\omega - B^T \hat\omega\|^2.
\end{equation}

We can also derive an analytic solution for an arbitrary $\Sigma = \mathbb{E}_{x \sim \mathcal{X}} xx^T$.
Note that $\|v\|_\Sigma^2 = v^T \Sigma v = (\sqrt \Sigma v)^T \sqrt \Sigma v = \|\sqrt\Sigma v\|^2$.
Then, we have
\begin{equation}
    \|\omega - B^T \hat\omega\|_\Sigma^2 = \|\sqrt \Sigma \omega - \sqrt \Sigma B^T \hat\omega\|^2.
\end{equation}
The optimal solution is then given by
\begin{equation}
    (\sqrt \Sigma B^T)^{+} \sqrt \Sigma \omega \in \arg\min_{\hat \omega} \|\omega - B^T \hat\omega\|^2.
\end{equation}

\subsection{Visualization of some PI-resize matrices}\label{sec:app:flexi:pi_matrix}

\begin{figure}[t]
  \centering
  \includegraphics[width=1.0\linewidth]{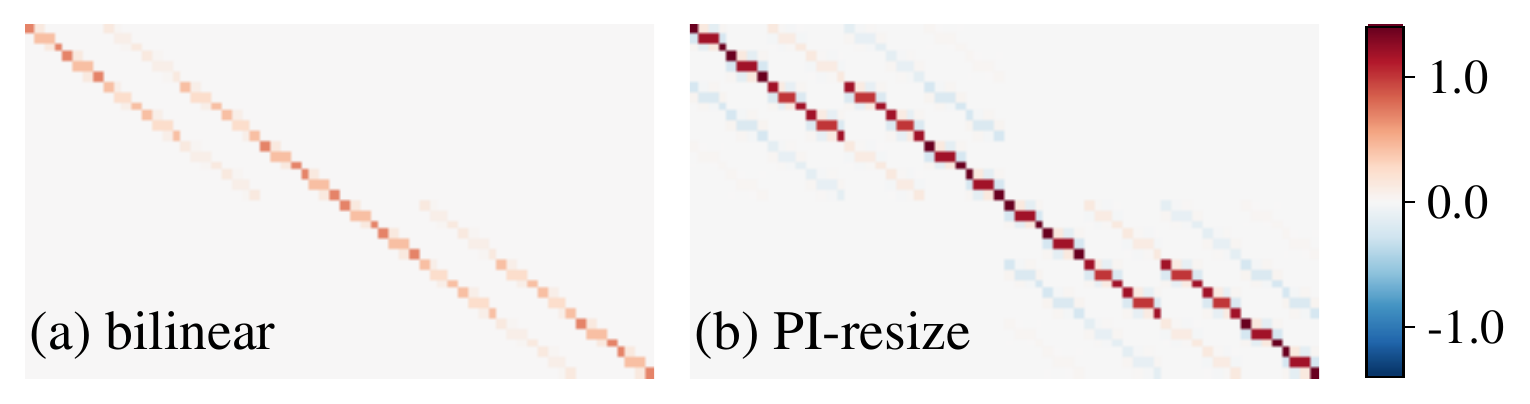}
  \includegraphics[width=1.0\linewidth]{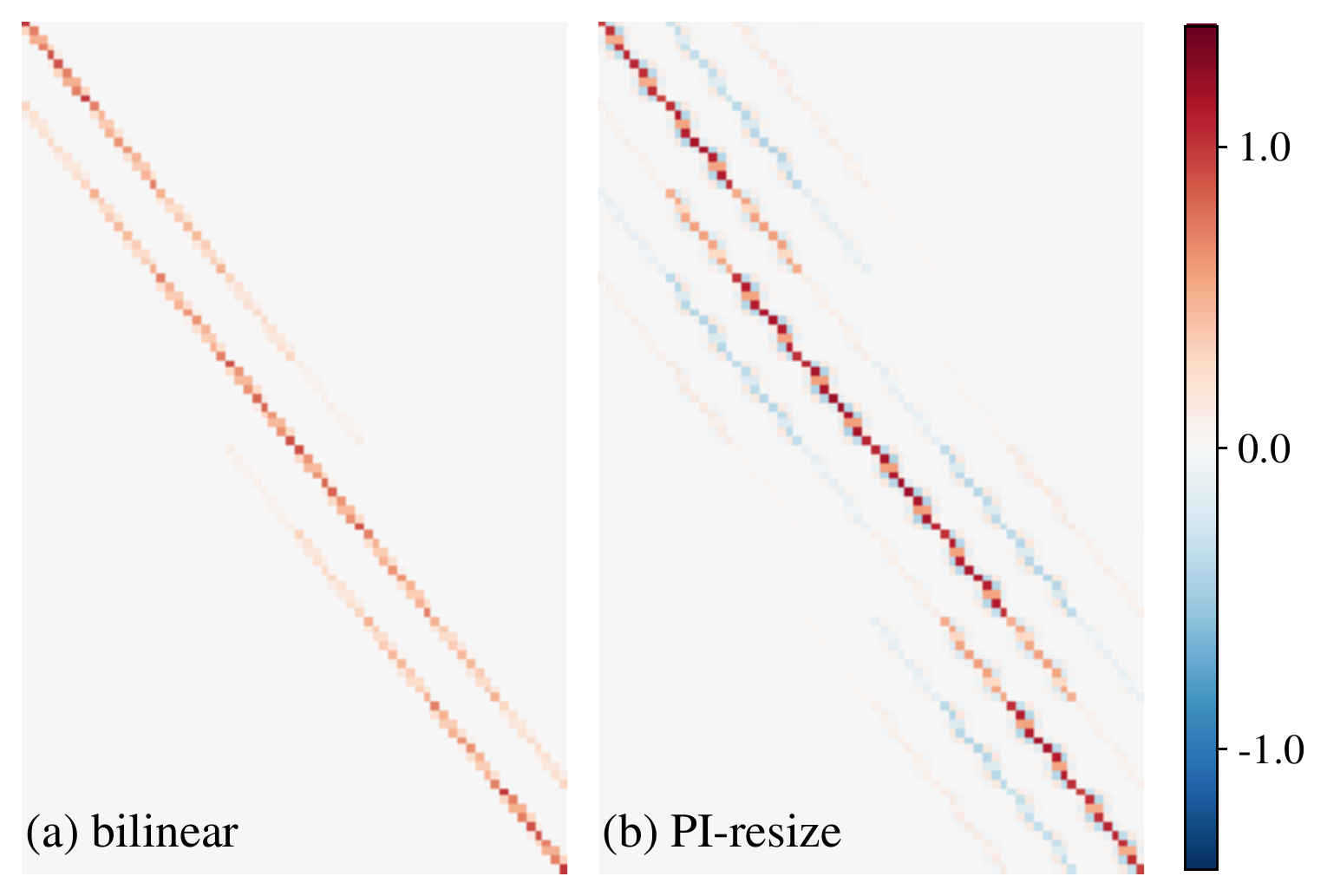}
  \caption{\textbf{Visualization of resize matrices.} Each row corresponds to the weights that are used in the computation of one output pixel.
  Off-diagonals correspond to pixels below/above the current one.
  We can see that PI-resize does include negative weights, has a larger \emph{receptive field}, and uses overall larger weights than bilinear resizing.}
  \label{fig:app:flexi:pi_matrix}
\end{figure}

We visualize an upscaling and a downscaling matrix for both bilinear and PI-resize operations for a visual comparison in Figure~\ref{fig:app:flexi:pi_matrix}.

\subsection{Visualization of patch-embedding weights}\label{sec:app:flexi:patchembs}

\begin{figure}
    \centering
    \includegraphics[width=0.95\linewidth]{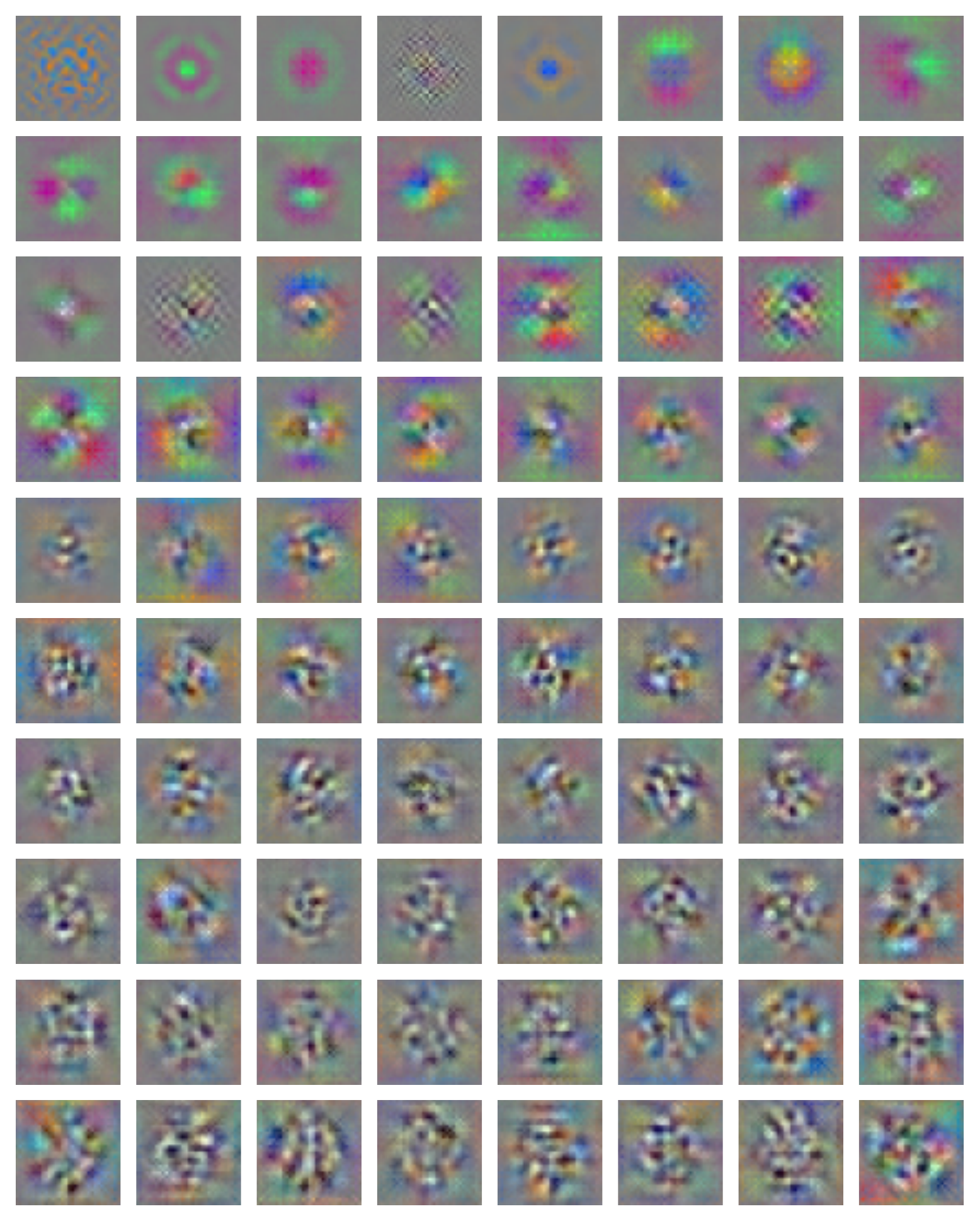}
    \caption{First 80 PCA components of the raw underlying 32x32 patch embedding weights of FlexiViT.}\label{fig:app:patchemb:32x32}
\end{figure}

\begin{figure}
\centering
\begin{minipage}[t]{.49\columnwidth}
    \includegraphics[width=1.0\columnwidth]{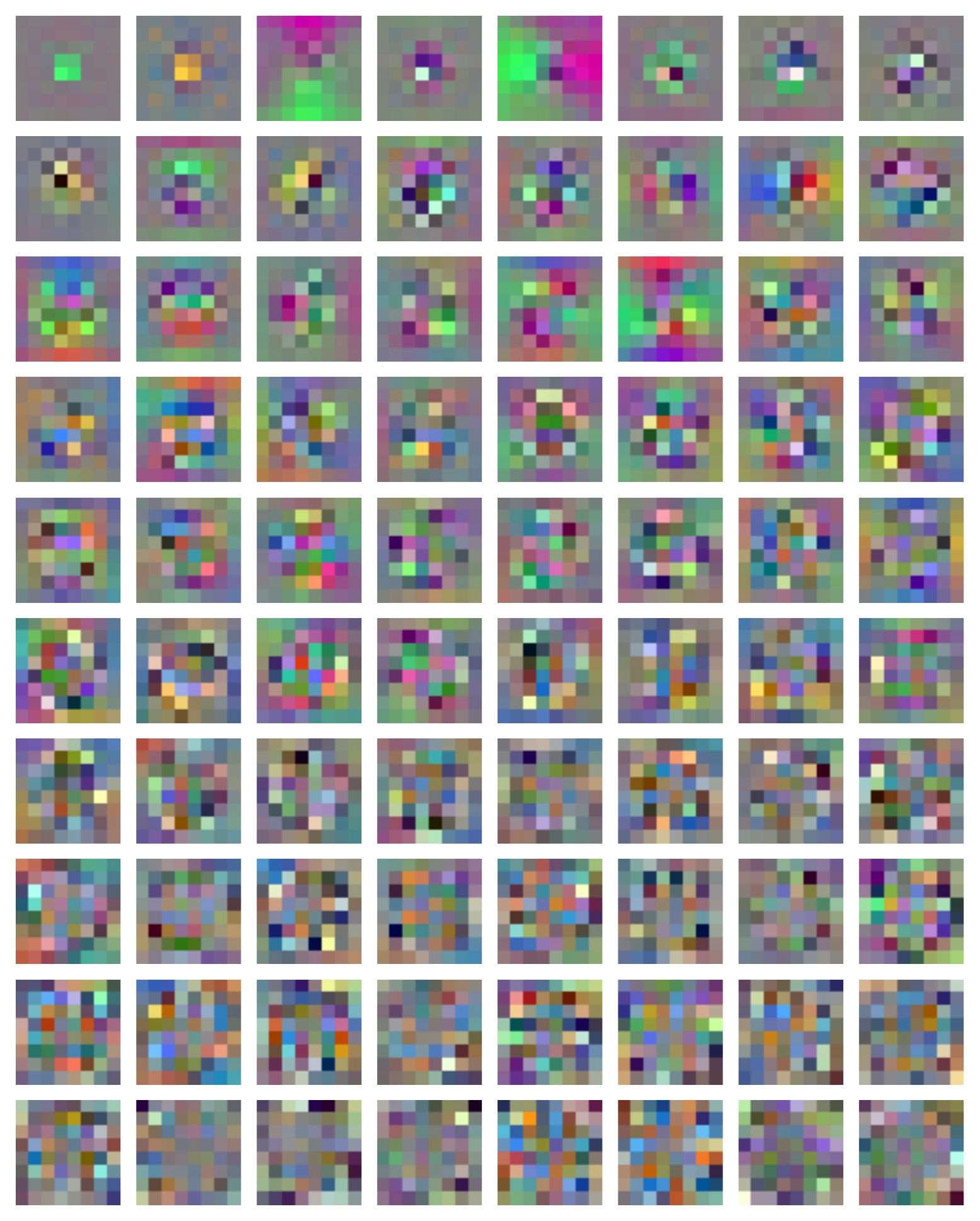}
    \caption{First 80 PCA components of the patch embedding weights from Figure~\ref{fig:app:patchemb:32x32} bilinearly resized to 8x8.}\label{fig:app:patchemb:bl8x8}
\end{minipage}\hfill
\begin{minipage}[t]{.49\columnwidth}
    \includegraphics[width=1.0\columnwidth]{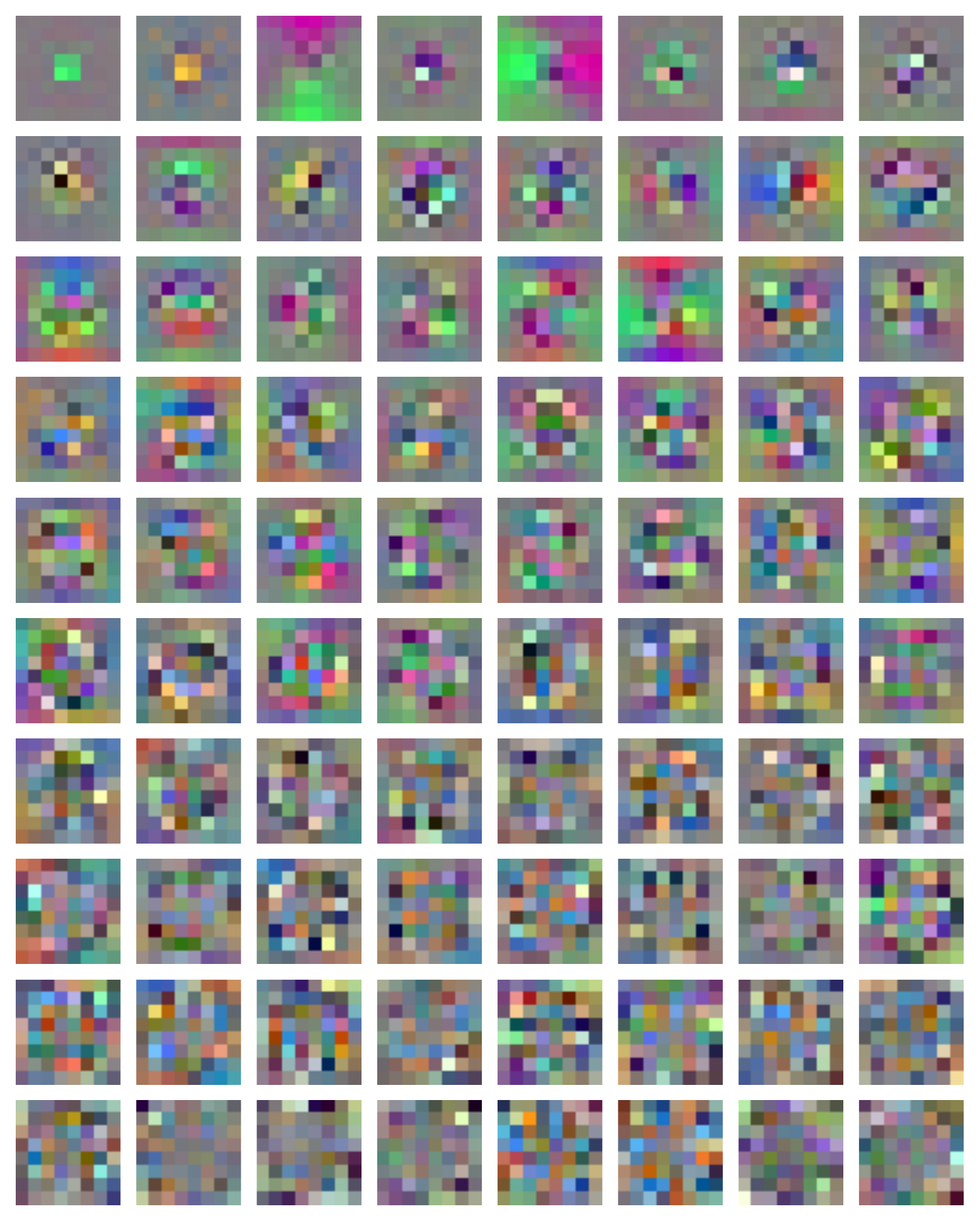}
    \caption{First 80 PCA components of the patch embedding weights from Figure~\ref{fig:app:patchemb:32x32} PI-resized to 8x8.}\label{fig:app:patchemb:pi8x8}
\end{minipage}
\end{figure}

\begin{figure}
    \centering
    \includegraphics[width=0.95\linewidth]{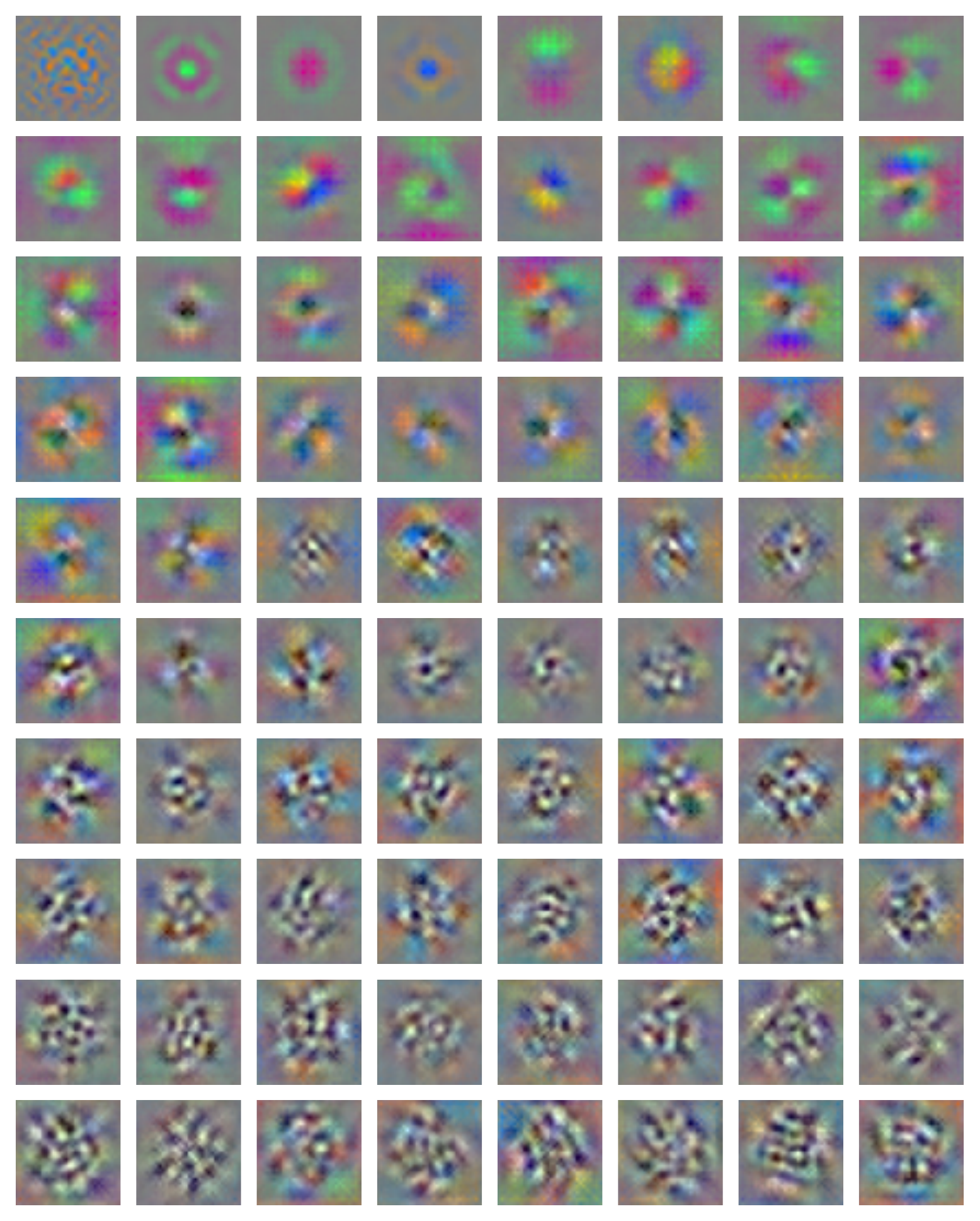}
    \caption{First 80 PCA components of the patch embedding weights from Figure~\ref{fig:app:patchemb:32x32} bilinearly resized to 48x48.}\label{fig:app:patchemb:bl48x48}
\end{figure}

\begin{figure}
    \centering
    \includegraphics[width=0.95\linewidth]{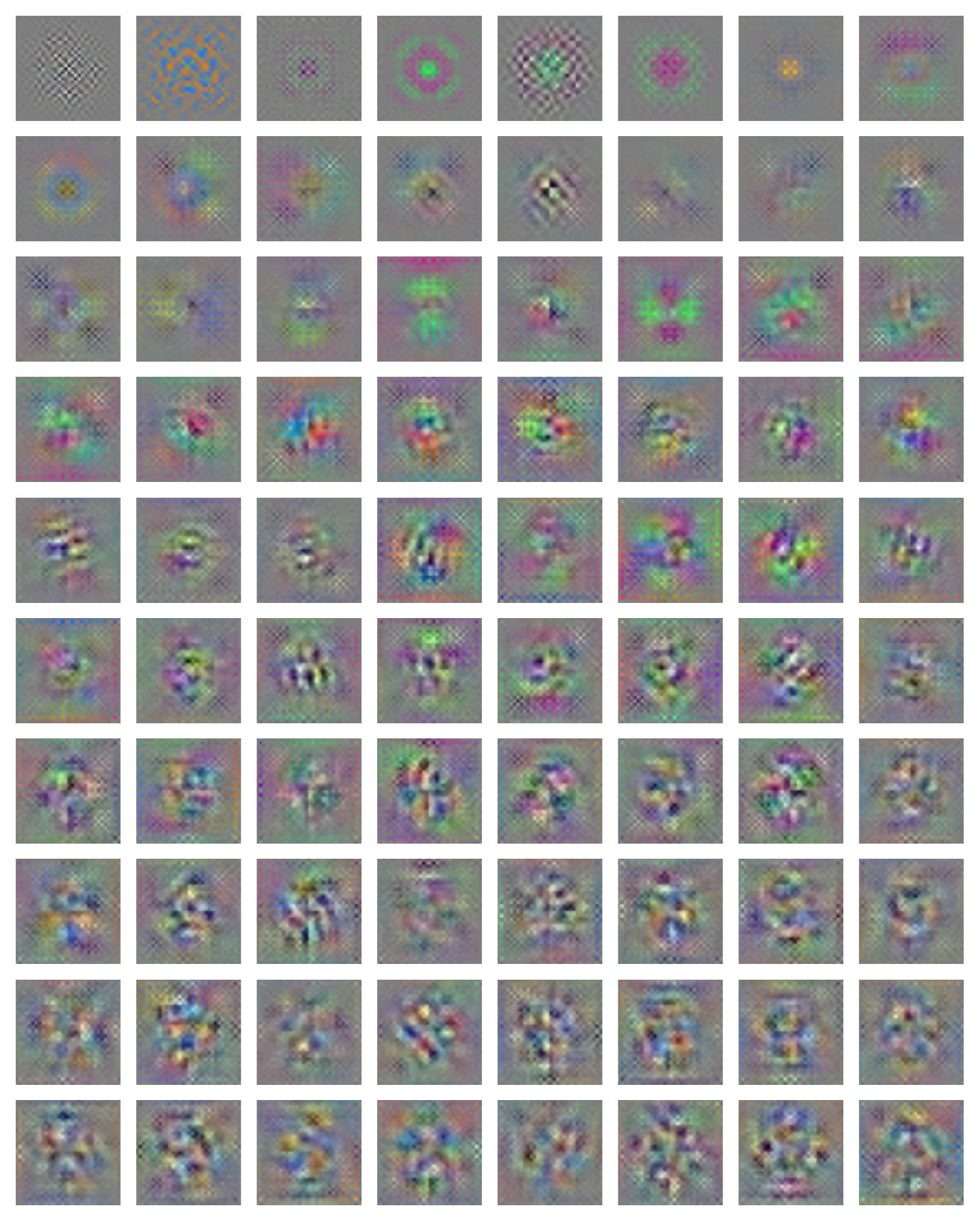}
    \caption{First 80 PCA components of the patch embedding weights from Figure~\ref{fig:app:patchemb:32x32} PI-resized to 48x48.}\label{fig:app:patchemb:pi48x48}
\end{figure}

PCA of patch embeddings (see~\cite{dosovitskiy2021vit}) in \cref{fig:app:patchemb:32x32,fig:app:patchemb:bl8x8,fig:app:patchemb:pi8x8,fig:app:patchemb:bl48x48,fig:app:patchemb:pi48x48}.

\section{The ``underlying'' parameter shapes}\label{sec:app:underlying}

\begin{figure}[b]\setlength{\hfuzz}{1.1\columnwidth}
\begin{minipage}{\textwidth}
  \centering
  \includegraphics[width=1.0\linewidth]{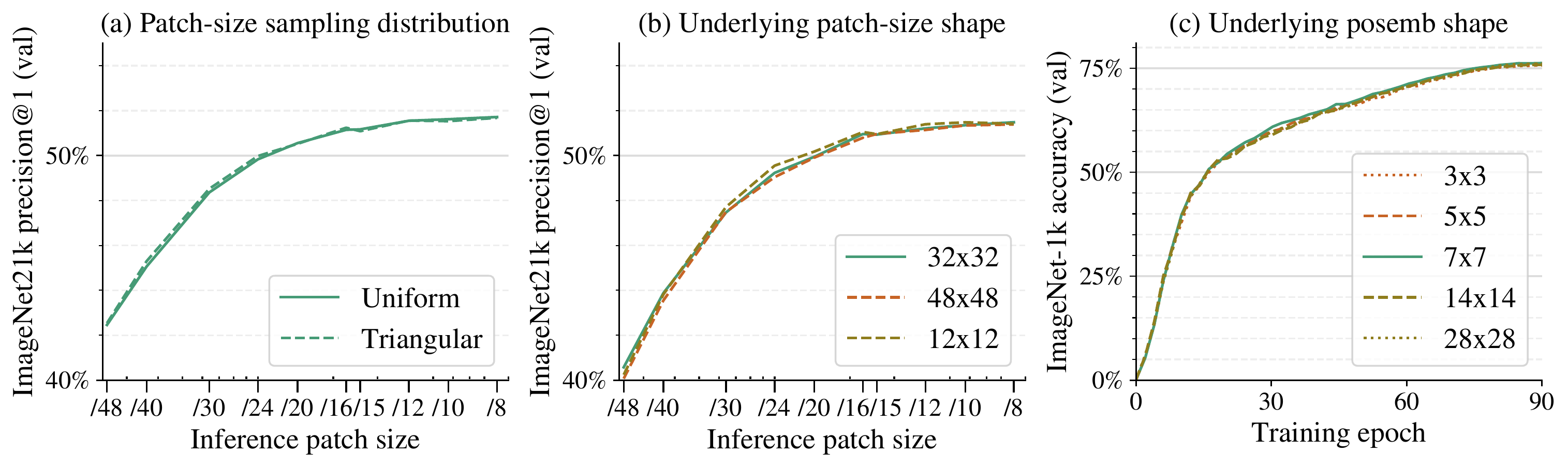}
  \caption{
  \textbf{Ablation of a few unimportant hyperparameters,} since changing their value shows no noteworthy difference in the result.
  \textbf{(a)} Sampling distribution of patch sizes.
  Uniform is preferable for its simplicity, but we use ``Triangular'' (more weight on mid-sized patches, less weight on extremely large or small ones) sometimes for legacy reasons.
  \textbf{(b)} Varying the shape of the underlying patch embedding parameter in the full FlexiViT training setup.
  \textbf{(c)} Varying the shape of the underlying position embeddings in a smaller supervised training of a ViT-S/16 baseline following~\cite{betterviti1k}.
  }\label{fig:app:underlying}
\end{minipage}
\end{figure}

FlexiViT does introduce two new hyper-parameters which were not present in the original ViT architecture: the size of the \emph{underlying} patch-embedding weights and position-embeddings (i.e.\ learned params, before resize).

However, both of these parameters have (maybe surprisingly) little influence on the final performance of the model, as long as they are in a ``reasonable'' range.
We verified these in two different settings.

\PAR{For the patch-size parameter,} since it is affected by the resize method used, we perform the ablation in the full FlexiViT setup.
Figure~\ref{fig:app:underlying}~(b) shows that there is no notable difference across all evaluation sizes, and hence we stick to the (initially arbitrary) default of 32 across all experiments.

\PAR{For the position embedding parameter,} we ran an early experiment with ViT-S/16 trained from scratch on ImageNet-1k following~\cite{betterviti1k}.
Figure~\ref{fig:app:underlying}~(c) shows that in general, even for plain ViT training, this approach could be taken and training curves are mostly unaffected by in-graph bilinear resizing of position embeddings.

\enlargethispage{-19.5\baselineskip}

\section{Distribution of patch-size sampling}\label{sec:app:distr}


During roughly the first half of this project, we sampled patch-size from a distribution which is not uniform, but samples patch-sizes between 16 and 30 up to three times more than patch-sizes outside this range.
This ``triangular'' distribution was based purely on gut-feeling, and once we verified that it is no better than uniform sampling (the experiment is shown in Figure~\ref{fig:app:underlying}~(a)), we decided to use a uniform distribution for simplicity.
We further decided to avoid the costly re-running of all experiments we did so far, and thus some experiments in the main paper were done with the ``triangular'' distribution.
However, in all direct comparisons presented throughout the paper, the curves being directly compared always were trained in the exact same way.

\begin{figure*}[t]
  \centering
  \includegraphics[width=1.0\linewidth]{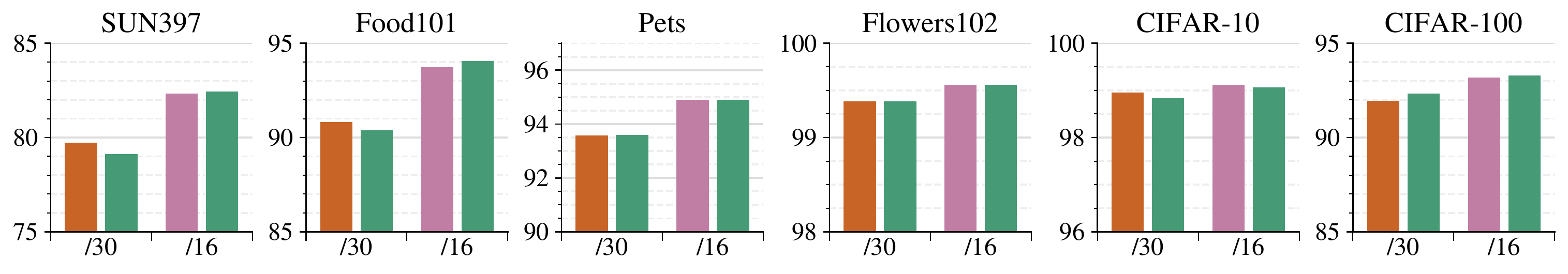}\vspace{-1em}%
  \caption{\textbf{More transfer results.}
  The legend is the same as in Figure~\ref{fig:using_flexivit}.
  The main take-away is that the results are qualitatively the same across a wide range of image classification tasks and it is safe to use a pre-trained FlexiViT model in lieu of a ViT model even when one only uses it at a fixed patch size after transfer.
  }\label{fig:app:using_flexivit_transfer}\vspace{-1em}
\end{figure*}

\section{Detais on resource-efficient transfer}\label{sec:app:fast-i1k-transfer}

For finetuning FlexiViT models on the Imagenet-1k dataset we generally follow the transfer learning setup from~\cite{dosovitskiy2021vit}. We use SGD momentum optimizer, with the initial learning rate of $0.03$ and cosine learning rate decay. We also reduce the learning rate for the pretrained parameters by a factor of 10. We optimize for $20000$ steps with inception crop and flip left-right augmentation, using batch size 512 and input image size of $480 \times 480$.

\section{Using pre-trained FlexiViT models: more details and results}\label{sec:app:transfer}

In this section, we provide more details and full results for the scenario described in Section~\ref{sec:using}: using pre-trained FlexiViT models.

For more details on flexified training procedures discussed in Section~\ref{sec:flexifying}, we redirect to
Appendix~\ref{sec:app:flexify_transfer} for flexified transfer learning,
Appendix~\ref{sec:app:flexify_lit} for flexified contrastive image-text learning (LiT and CLIP), and
Appendix~\ref{sec:app:flexify_owl} for flexified open-vocabulary detection (OWL-ViT), including results on ELEVATER.

\subsection{Using FlexiViT models for transfer}\label{sec:app:transfer:clf}

For transfer, we follow the simple BiT-HyperRule~\cite{kolesnikov2019large}.
In short, we transfer for a relatively brief number of steps (500 for flowers and pets, 2500 for food101 and sun, and 10\,000 for CIFAR), using the SGD optimizer with a momentum of $0.9$, no weight decay, no dropout, and no other augmentations besides flips and random crops.
We initialize the new classification layer to all-zeros and use a short learning-rate warmup, both of these having as effect to better preserve the pre-trained weights.
The only setting which differs from \cite{kolesnikov2019large} is that we do sweep the learning-rate across $\{0.03, 0.01, 0.003, 0.001\}$ for each task individually.
We show results for all 6 datasets we used in Figure~\ref{fig:app:using_flexivit_transfer}.

\subsection{Using FlexiViT models\\ in LiT for image-text tasks}\label{lit_flexivit}

\begin{figure*}[t]
  \centering
  \includegraphics[width=1.0\linewidth]{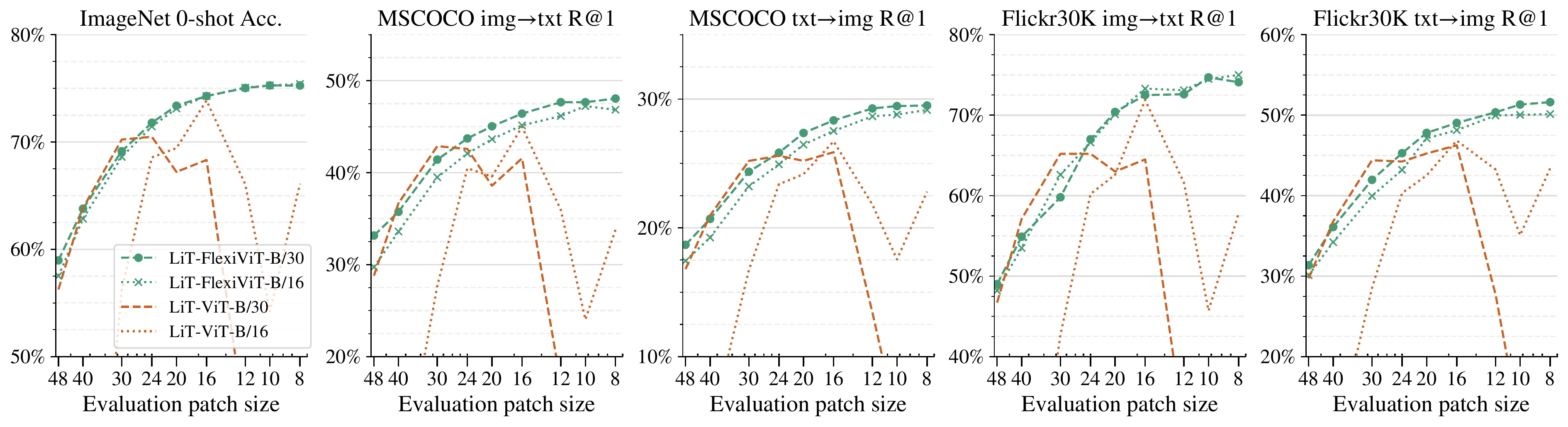}
  \caption{\textbf{Transfer of FlexiViT at a fixed patch size.}
  The LiT-ViT baselines match LiT-FlexiViT on the sequence length it has been trained for ($16^2$ or $30^2$), but performance drops quickly when using a different inference sequence length.
  We observe that the LiT-FlexiViT models work well across different inference sequence lengths, even though only a fixed sequence length is used during LiT transfer.
  This effect is similar to that described in Section~\ref{sec:using:fasttrans} and further explored in Figure~\ref{fig:app:flexilit}.
  }\label{fig:app:lit_flexivit}

  \includegraphics[width=1.0\linewidth]{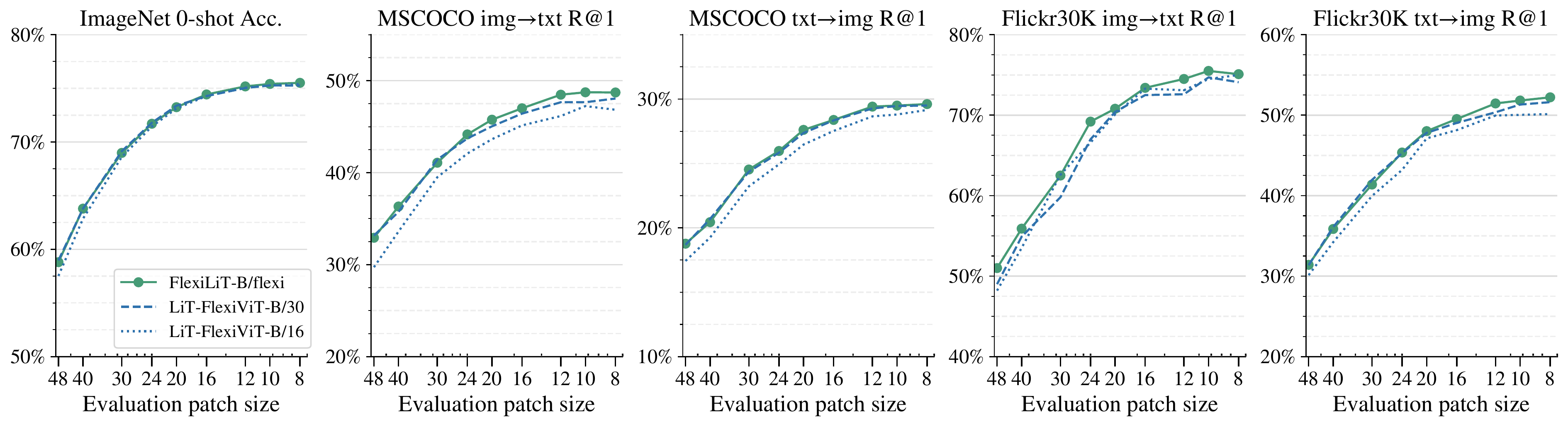}
  \caption{\textbf{FlexiLiT.} We observe consistent but marginal boost when flexifying the LiT training by randomizing patch size during LiT-tuning. This again shows that the LiT-FlexiViT baseline performs strongly and it allows fast transfer: transferred cheaply using a large patch size and served at smaller patch sizes for free.}
  \label{fig:app:flexilit}

  \includegraphics[width=1.0\linewidth]{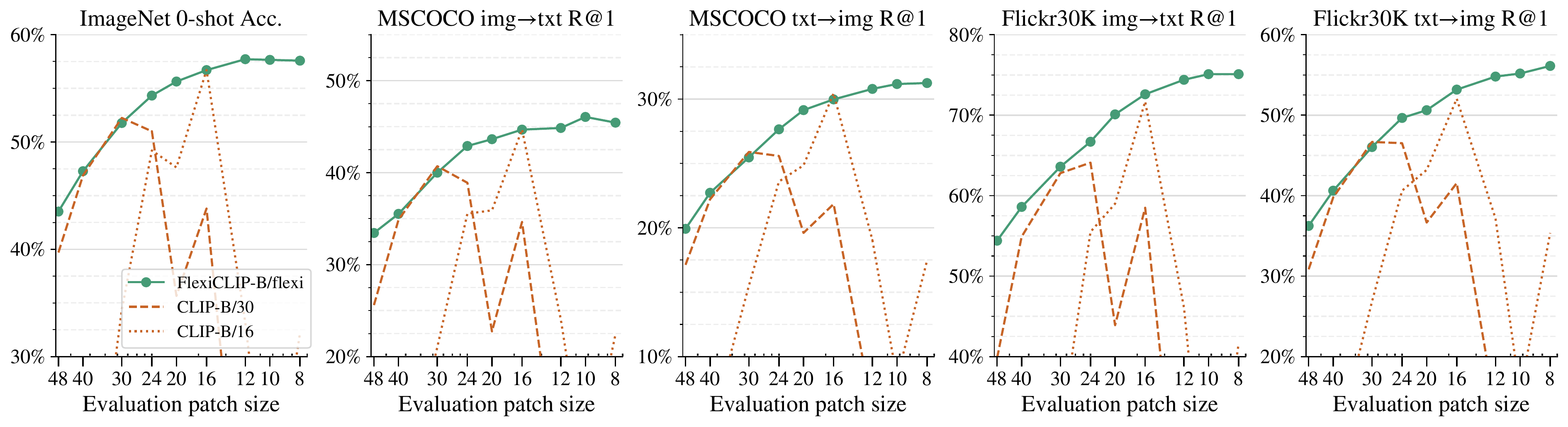}
  \caption{\textbf{FlexiCLIP.}
  We observe very similar conclusions as in FlexiViT, that the FlexiCLIP model works well across a large range of evaluation patch sizes during inference.
  It is interesting that a single fixed sequence length text tower, is able to produce embeddings aligning well with a FlexiViT image tower that allows multiple sequence lengths.
  }\label{fig:app:flexiclip}
  
\end{figure*}

We use the same 4B image-text pairs dataset as in~\cite{lit} to train the LiT models, and use identical hyper parameters as LiT models.
The only difference is to use the FlexiViT model here, instead of a standard ViT model.
FlexiViT models are transferred at a fixed sequence length, i.e. $30^2$ or $16^2$, with $240\times240$ image resolution.
We report zero-shot classification results on ImageNet~\cite{russakovsky2015imagenet}, zero-shot image-to-text / text-to-image retrieval results on MS-COCO~\cite{mscoco} and Flickr30K~\cite{flickr30k}, in Figure~\ref{fig:app:lit_flexivit}.

\subsection{Using FlexiViT models\\ in OWL-ViT for zero-shot detection}

To evaluate FlexiViT backbones for open-vocabulary detection (Figure~\ref{fig:using_flexivit}), we compare OWL-ViTs initialized with either fixed or flexible LiT-B models. Specifically, the backbones start with either a fixed or flexibly pre-trained ViT image model, which is then then frozen and LiT-tuned~\cite{lit} with a text model at a fixed patch size (the same as final evaluation, i.e. $30^2$ and $16^2$ respectively). OWL-ViTs using these backbones are then trained at a fixed patch size ($30^2$ or $16^2$) at a resolution of $720\times720$ on Objects365~\cite{shao2019vg} and Visual Genome~\cite{krishnavisualgenome} as in the original paper~\cite{minderer2022simple}. We report mean average precision (AP) on LVIS~\cite{gupta2019lvis}.

\subsection{Using FlexiViT models\\ in UViM for panoptic segmentation}

Apart from using FlexiViT model weights for the initialization, we follow the setup of the original UViM setup~\cite{kolesnikov2022uvim} as close as possible. In particular, we train the model on the COCO panoptic dataset~\cite{mscoco} and report the standard PQ metric~\cite{kirillov2019panoptic} on the official validation split. We train the model for 200 epochs, using the custom adafactor optimizer variant~\cite{zhai2022scaling} with the base learning rate of $0.001$. The learning rate for the pretrained part of the model is decreased by a factor of 10. The input image size is $512 \times 512$. More details on the training setting can be found in the UViM paper~\cite{kolesnikov2022uvim} and official repository of the UViM model~\footnote{\url{https://github.com/google-research/big_vision/tree/main/big_vision/configs/proj/uvim}}.

\subsection{Using FlexiViT models\\ in Segmenter for semantic segmentation}
We follow the experimental setup of Segmenter~\cite{strudel2021segmenter} for end-to-end finetuning of Vision Transformer with linear decoder.
For data augmentation during training, we apply random resizing of the image with a random ratio between 0.5 and 2.0, photometric augmentation and random horizontal flipping.
We randomly crop images to $480 \times 480$ resolution with padding, therefore preserving aspect ratio.
We use the $480 \times 480$ resolution for both Cityscapes and ADE20k.
We train for 127 epochs with minibatch size of 16 (resulting in 160k iterations on ADE20k).
We use the ``poly'' learning rate decay schedule and sweep the base learning rate in $\{1e-4, 3e-4, 8e-4\}$ for all of our runs.
Weight-decay is kept fixed at $0.01$.
At evaluation time, we use the sliding-window with a resolution $480 \times 480$ to handle varying image sizes during inference.
Table 3 row 6 in~\cite{strudel2021segmenter} reports $48.06$ mIoU.
Average of 6 runs in our codebase in the same setting gives $47.6 \pm 0.4$ mIoU.
The results on ADE20k are provided in Table~\ref{tbl:app:using_flexivit}.

\section{Full numerical ImageNet-1k-only results}\label{sec:app:i1k}

We provide full numerical results of the FlexiViTs trained purely on ImageNet-1k and presented in Figure~\ref{fig:i1k}, including on additional robustness and OOD test-sets in \cref{tbl:app:i1k_flexi1200,tbl:app:i1k_flexi600,tbl:app:i1k_flexi300,tbl:app:i1k_flexi90}.

\section{FlexiLiT and FlexiCLIP results}\label{sec:app:flexify_lit}

FlexiLiT follows exactly the same setup as described in Section~\ref{lit_flexivit}, but randomizes patch sizes during LiT training.
We show more FlexiLiT results in Figure~\ref{fig:app:flexilit}.

For FlexiCLIP, we simply replace the pre-trained and frozen backbone in FlexiLiT with a random initialized and unfrozen backbone, which corresponds to the \textit{uu} setting in~\cite{lit} and is equivalent to CLIP~\cite{clip}.
Figure~\ref{fig:app:flexiclip} shows the same conclusions in this setting.
It is reassuring that the patch size randomization does not hinder learning both image and text representations from scratch simultaneously.

\section{Accelerate pre-training}\label{sec:app:accelerate_pretraining}

\begin{figure}[t]
    \centering
    \includegraphics[width=\columnwidth]{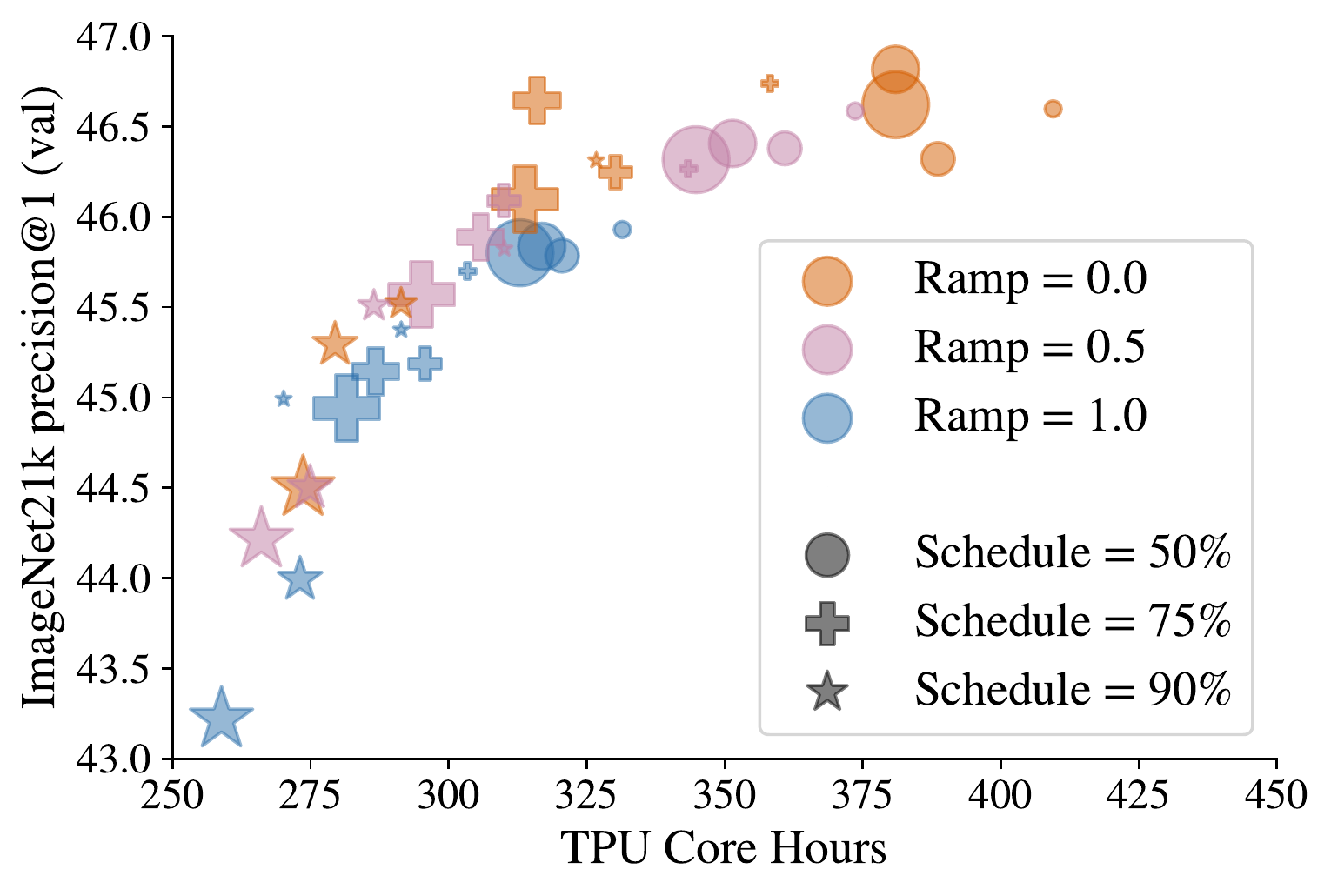}
    \caption{A detailed look into the impact of the larger patch size (size of the markers), schedule, and ramp periods on both compute and accuracy. See Appendix~\ref{sec:app:accelerate_pretraining} for details.}
    \label{fig:app:acc_pretrain_analysis}
\end{figure}

As discussed in Section~\ref{sec:accelerate_pretrain}, one can use FlexiViT's method of varying the patch size and resizing the embedding weights to pretrain ViTs faster. To do that, we specify a \emph{curriculum}: a sequence $(p_k)_{k=1}^K$ of probability distributions over the patch sizes along with a mapping $c:\mathbb{N}\to[K]$ that identifies which distribution $p_k$ to use at training step $t$. In this section, we use the sequence of patch sizes: $(48, 40, 30, 24, 16)$ to demonstrate the potential savings in compute.

We set $16\times 16$ to be the desired/target patch size during evaluation. Hence, we compare a curriculum-based approach of pretraining ViTs (denoted FasterViT) with the standard ViT/B/16  architecture in which the patch size is fixed to $16\times 16$ throughout training. Except for the variable patch sizes and the embedding layers, both architecture are otherwise identical. 

\begin{figure}[t]
    \centering
    \includegraphics[width=\columnwidth]{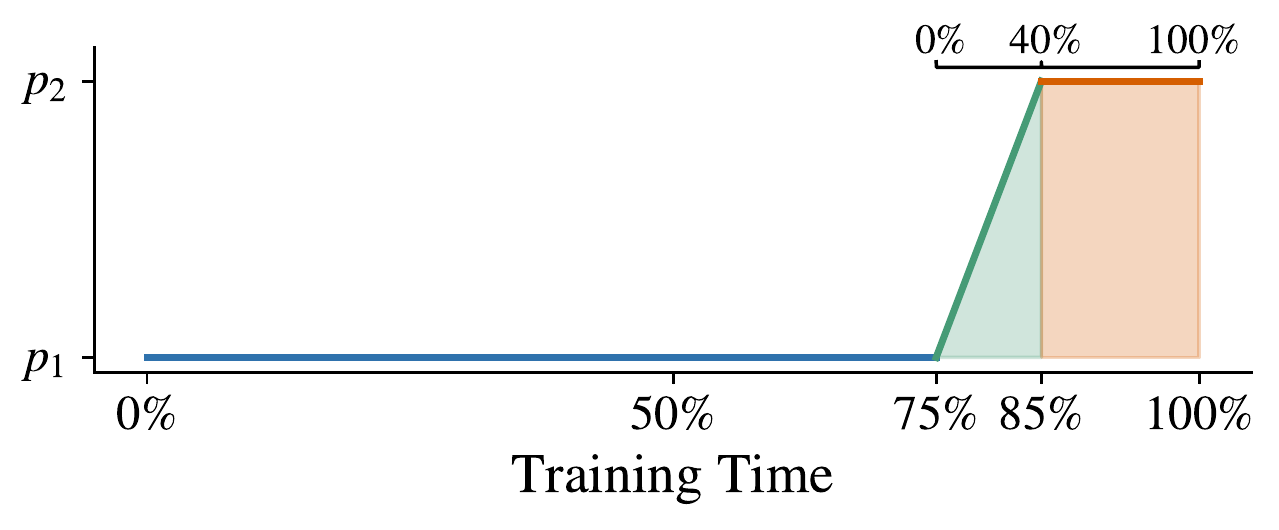}
    \caption{This figure illustrates how the distribution of patch sizes changes when accelerating pretraining (see Section~\ref{sec:accelerate_pretrain}) and Appendix~\ref{sec:app:accelerate_pretraining}. Here, the first 75\% of training steps use the larger patch size $p_1$ while the remaining 25\% are used for the desired/target patch size. The ramp period in this figure is 40\%.}
    \label{fig:app:pretrain_schedule}
\end{figure}

We use a simple curriculum in this evaluation. Specifically, we have one large patch size (e.g. $48\times 48$), which we denote by $p_0$ and the desired patch size $16\times 16$, which we denote by $p_1$. We initially use the larger patch size before swtiching to the smaller patch size using FlexiViT's PI-resize. We experiment with three different schedules: $(50\%, 75\%, 90\%)$, where $\mathrm{schedule} = 75\%$ means that 75\% of the training time uses the larger patch size alone. Instead of switching immediately between patch sizes, we also include an optional ramp period as illustrated in Figure \ref{fig:app:pretrain_schedule}. A ramp period of, say, 40\% means that 40\% of the  time allocated to the smaller patch size is used to transition gradually between the two distributions. We experiment with three ramp periods: $(0\%, 50\%, 100\%)$. We run each experiment independently and plot the resulting compute and accuracy in Figure \ref{fig:accelerate_pretrain}(x). As shown in the figure, FasterViT achieves the same level of accuracy as standard ViTs but with less compute, although the improvement in not quite significant.

In Figure \ref{fig:app:acc_pretrain_analysis}, we provide a detailed view into the impact of the three hyperparameters: patch size, schedule, and ramp period. Not surprisingly, increasing the fraction of time allocated to the larger patch size (e.g. by setting $\mathrm{schedule}=90\%$), improves compute at the expense of accuracy.

\section{Further analysis of cosine similarities between token representations across scales}\label{sec:app:cosine_scales}

\begin{figure}
  \centering
  \includegraphics[width=1.0\linewidth]{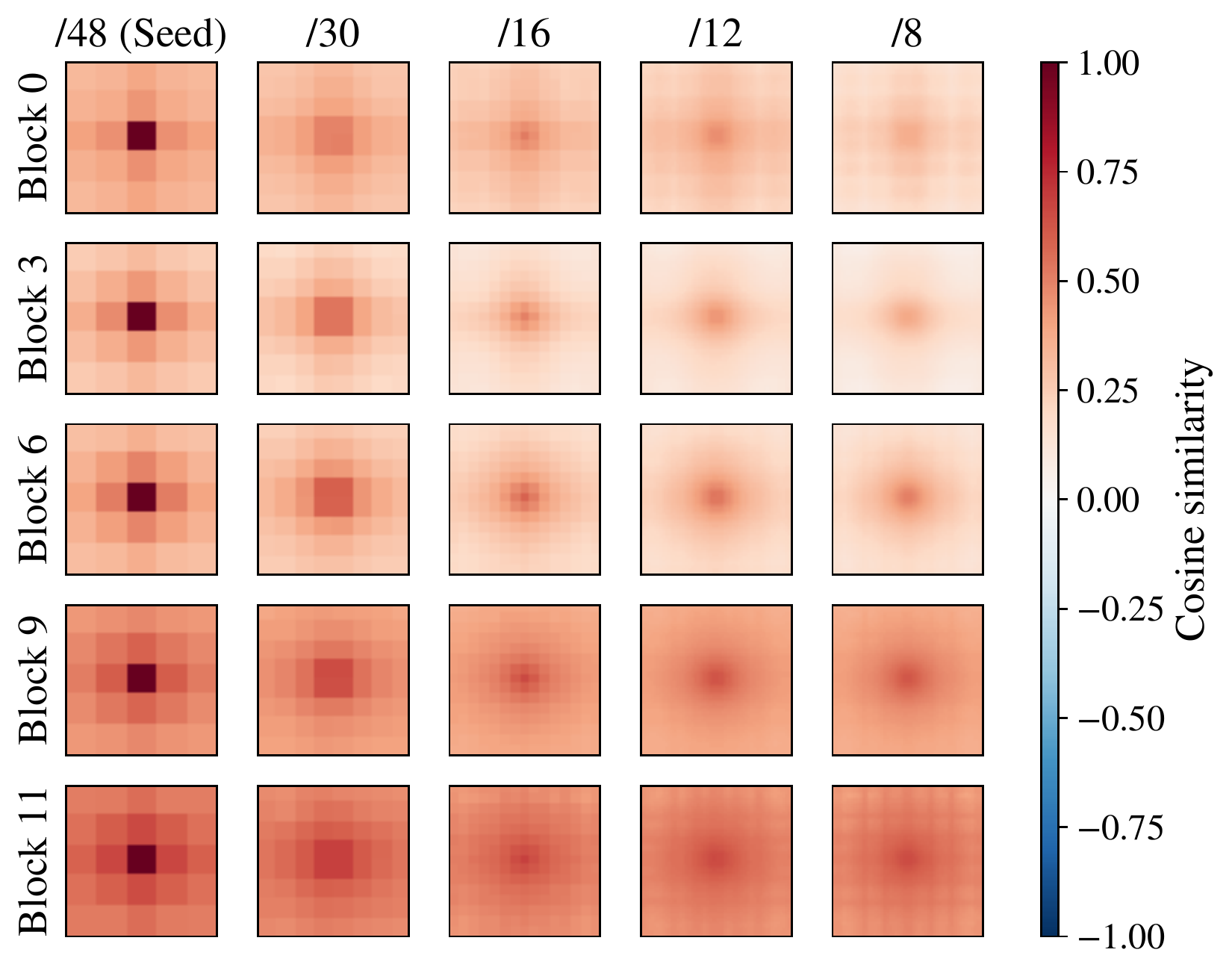}
  \includegraphics[width=1.0\linewidth]{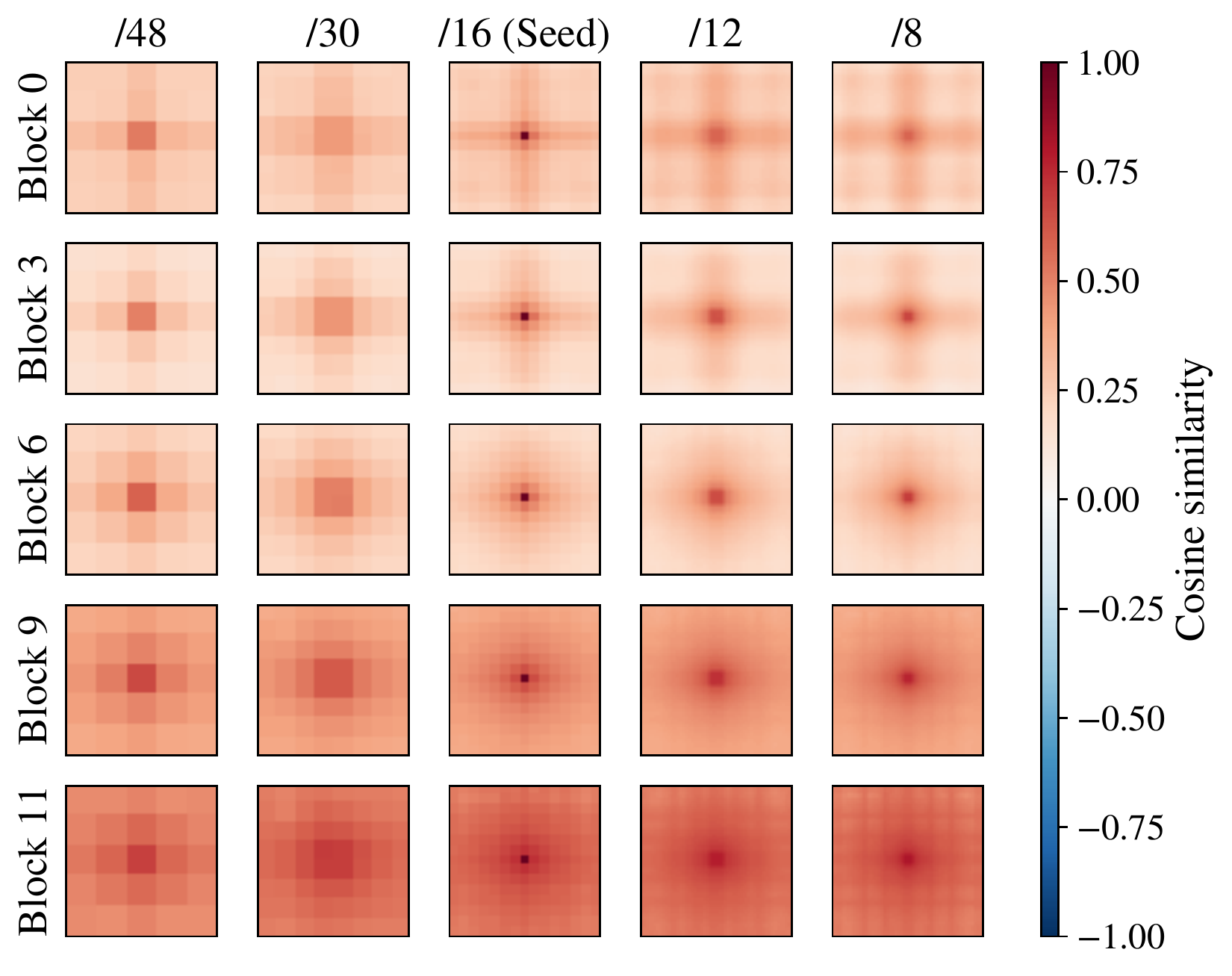}
  \includegraphics[width=1.0\linewidth]{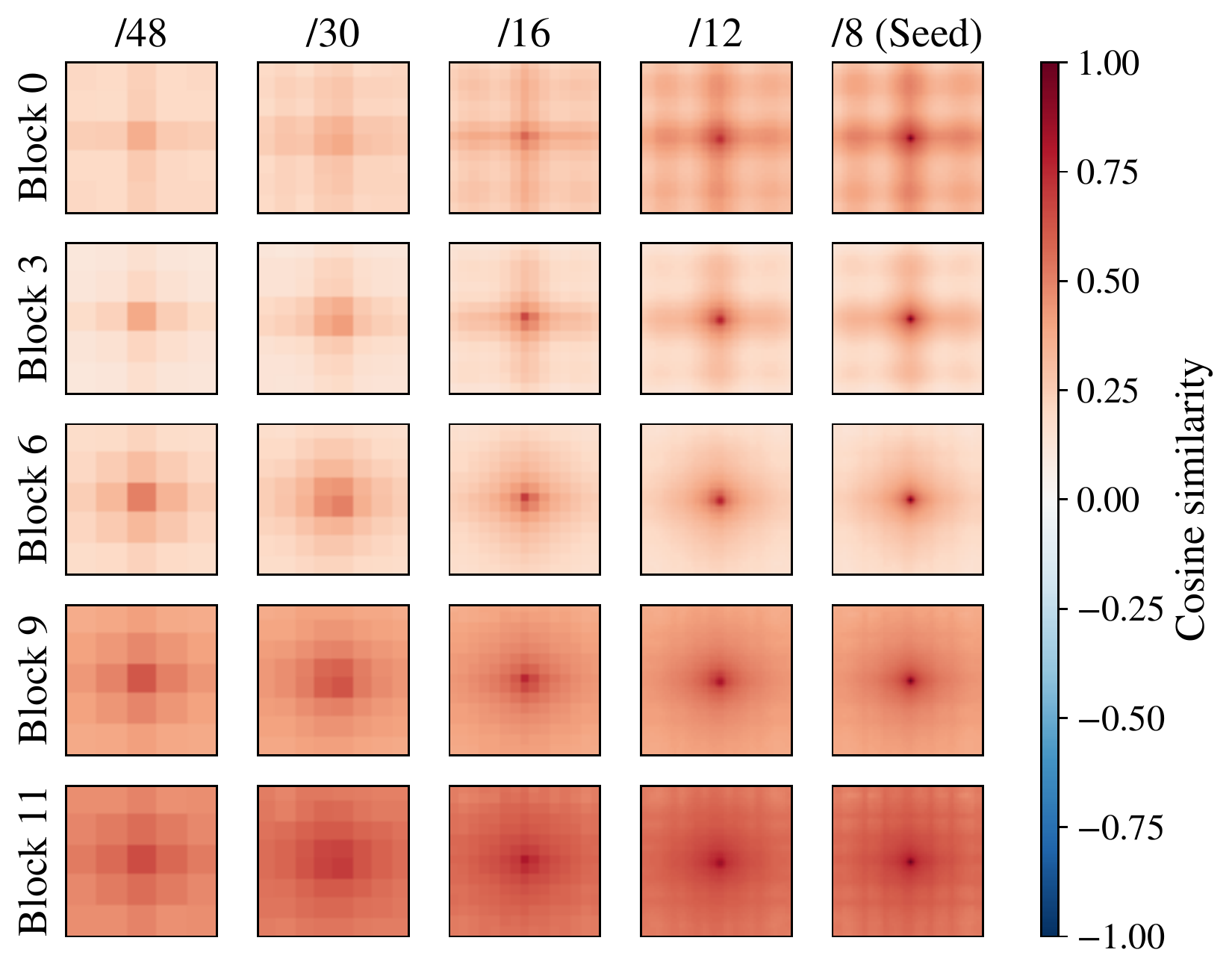}
  \vspace{-1.7em}
  \caption{\textbf{Supplemental analysis of similarities of token representations across scales.} Each row shows the cosine similarity between a seed token at the center of the feature map at one scale and tokens at different scales, for a single block. In the top group of plots, the seeds are taken from FlexiViT-B/48; in the middle, from FlexiViT-B/16; and at the bottom, from FlexiViT-B/8.}
  \label{fig:app:cosine_scales}
\end{figure}

In Section~\ref{sec:cosine_scales}, we measure cosine similarity between the representation of a seed token at one scale and representations of other tokens at other scales, demonstrating that the most similar tokens at other scales are those that represent the same spatial location. in Figure~\ref{fig:app:cosine_scales}, we provide additional results for seed tokens at other grid sizes and from additional blocks. Results are consistent with those in the main text.

\begin{figure}[b]
  \centering
  \includegraphics[width=1.0\linewidth]{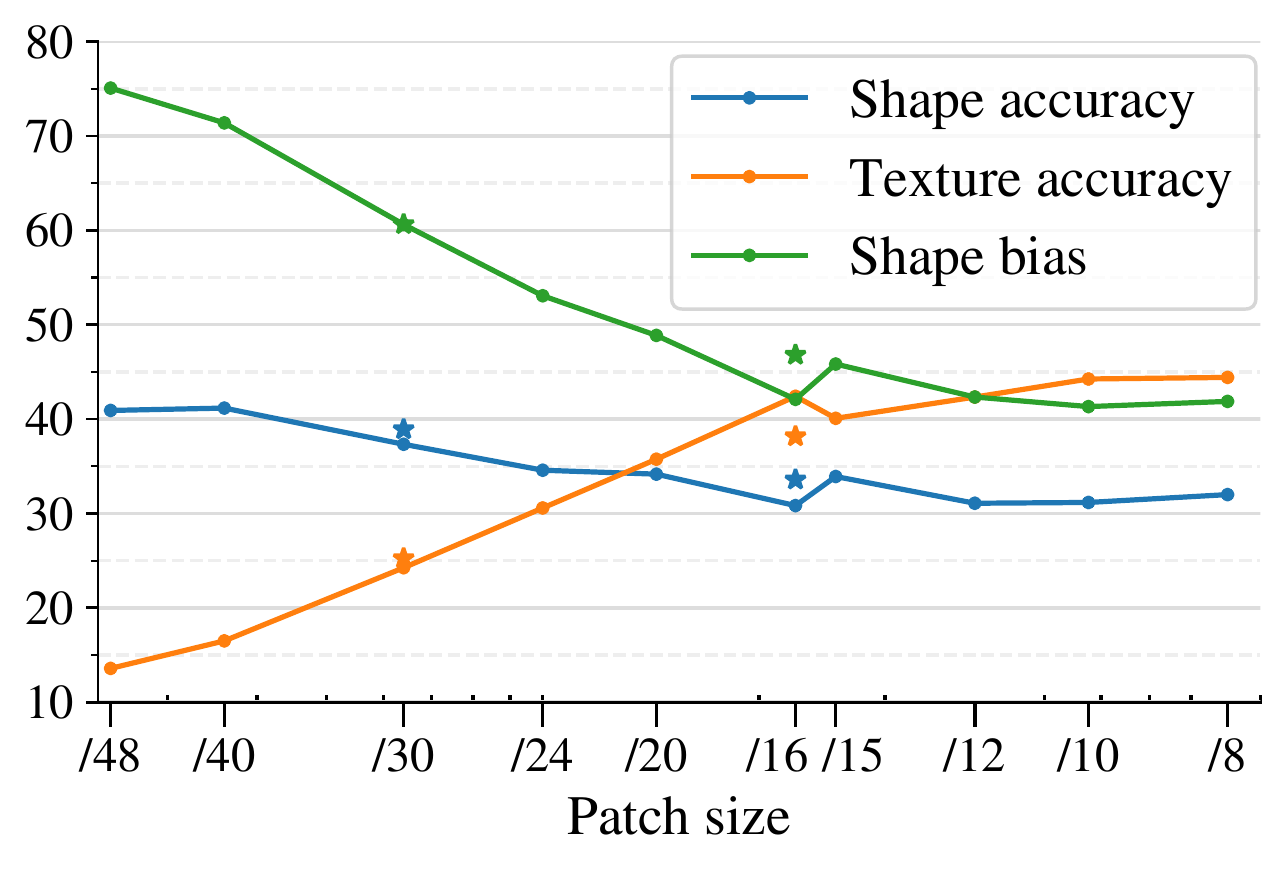}
  \caption{\textbf{Shape and texture bias of FlexiViT.} Lines reflect values for FlexiViT evaluated at different scales; stars reflect values for baseline ViT models trained at a single scale. Shape and texture accuracy are the top-1 accuracy of the model's prediction with respect to the shape and texture labels. Shape bias is the percentage of images that a classifier classifies by shape, provided that it correctly classifies them by either shape or texture.}
  \label{fig:app:shape_texture}
\end{figure}

\section{Ensembling FlexiViT predictions across scales}\label{sec:app:ensemble}
\begin{figure*}[h]
  \centering
  \includegraphics[width=1.0\linewidth]{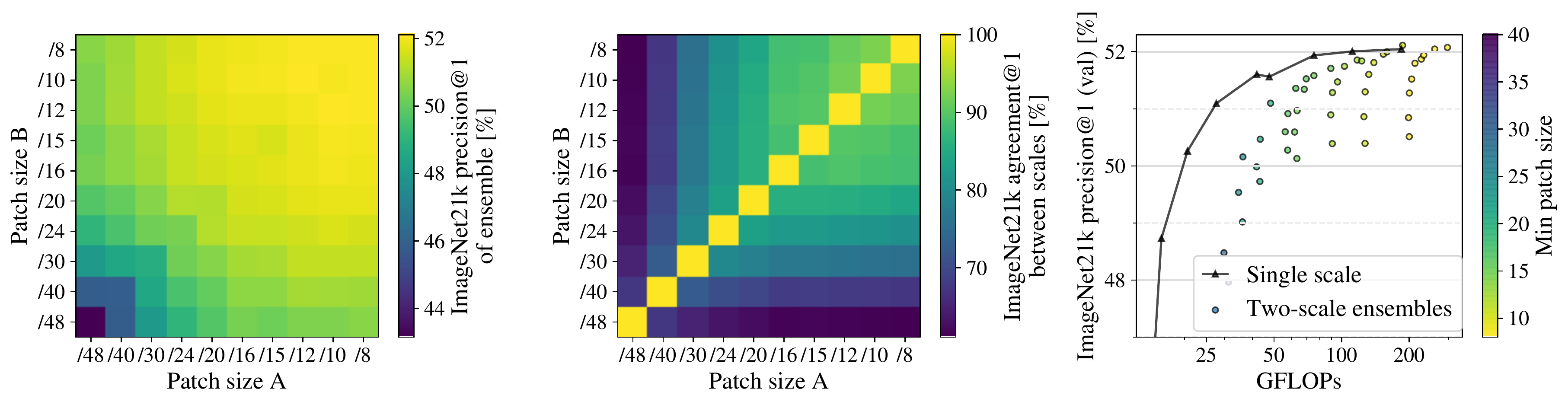}
  \caption{\textbf{Ensemble accuracy and agreement of FlexiViT-B evaluated at pairs of scales.}
  \textit{Left}: Accuracy of ensembles between scales. \textit{Middle}: Agreement of the top-1 predicted classes across scales. \textit{Right}: Accuracy of single-scale configurations and two-scale ensembles versus computational cost. Although agreement is relatively low at small patch sizes, there is little benefit to ensembling.}
  \label{fig:app:ensembling_pairs}
\end{figure*}

We ensemble FlexiViT models by averaging models' logits.\footnote{We have also explored ensembling based on averaging output probabilities. We find that results are nearly identical, but on average slightly worse.} When ensembling all models, we attain 51.7\% precision@1 on our ImageNet-21K validation set, which is slightly worse than the accuracy achieved at the largest grid size/smallest patch size (52.0\%).

We further explore ensembles of pairs of models in Figure~\ref{fig:app:ensembling_pairs}. Agreement between models evaluated at large patch sizes is relatively low, with models evaluated at the largest two patch sizes (/48 and /40) agreeing on only 67.4\% of examples (Figure~\ref{fig:app:ensembling_pairs} middle). Nonetheless, ensembling these models provides no accuracy improvement over simply using FlexiViT-B/40; both strategies achieve 45.8\% ImageNet-21K precision@1 (Figure~\ref{fig:app:ensembling_pairs} left).

When comparing the computational cost of ensembles of FlexiViT predictions across scales to applying FlexiViT at a single scale, a single scale is nearly always better (Figure~\ref{fig:app:ensembling_pairs} right). The only configuration where the accuracy of a two-scale ensemble exceeds the accuracy of a single scale with the same computational footprint is the ensemble of FlexiViT-B/10 and /12, which together have a similar computational cost to FlexiViT-B/8 (/10 + /12: 186.8 GFLOPs; /8: 184.5 GFLOPs). The ensemble attains marginally higher accuracy (52.1\% vs. 52.0\%), but this improvement is unlikely to be statistically significant nor practically meaningful.

\section{Shape and texture bias of FlexiViT}\label{sec:app:shapebias}

When confronted with images with conflicting shape and texture, ImageNet-trained models tend to produce labels that match their textures, whereas humans instead tend to assign labels that match their shapes~\cite{geirhos2018imagenet}. We evaluated FlexiViT using the same dataset as \cite{geirhos2018imagenet}, which was generated using the style transfer. In the dataset constructed by \cite{geirhos2018imagenet}, images have both shape and texture labels. We define the shape accuracy as the percentage of images for which the top-1 prediction matches the shape label, and texture accuracy as the percentage of images for which the top-1 prediction matches the texture label. As in \cite{geirhos2018imagenet}, we define shape bias by taking the ratio of the number of images classified according to their shape label to the number of images classified correctly according to either the shape or texture label and converting this ratio to a percentage. In other words, shape bias the ratio of shape accuracy to the sum of shape and texture accuracy $\times 100\%$. To evaluate ImageNet-21K models on the dataset of \cite{geirhos2018imagenet}, we use the mapping from WordNet IDs to the dataset classes provided by \cite{geirhos2018imagenet}. We take the top-1 class among the ImageNet-21K classes for which a mapping exists.

In Figure~\ref{fig:app:shape_texture}, we show that larger patch sizes lead to greater shape bias compared to smaller patch sizes. However, larger patch sizes have greater shape bias primarily because their texture accuracy is lower, rather than because their shape accuracy is higher. The shape biases of FlexiViT-B/16 and FlexiViT-B/30 are similar to the shape biases of ViT-B/16 and ViT-B/30 models trained at a single scale (stars).

\begin{figure*}
  \centering
  \includegraphics[width=1.0\linewidth]{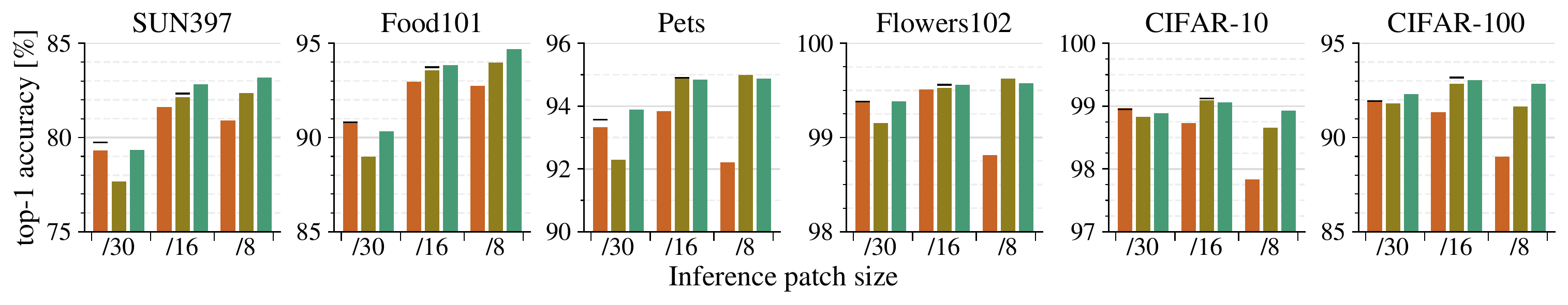}\vspace{-1em}%
  \caption{\textbf{More transfer results.}
  The legend is the same as in Figure~\ref{fig:flexify:transfer}.
  The main take-away is that the results are qualitatively the same across a wide range of image classification tasks.
  }\label{fig:app:flexify_transfer}\vspace{-1em}
\end{figure*}

\section{Attention relevances}\label{sec:app:attention_relevances}

\begin{figure}
  \centering
  \includegraphics[width=1.0\linewidth]{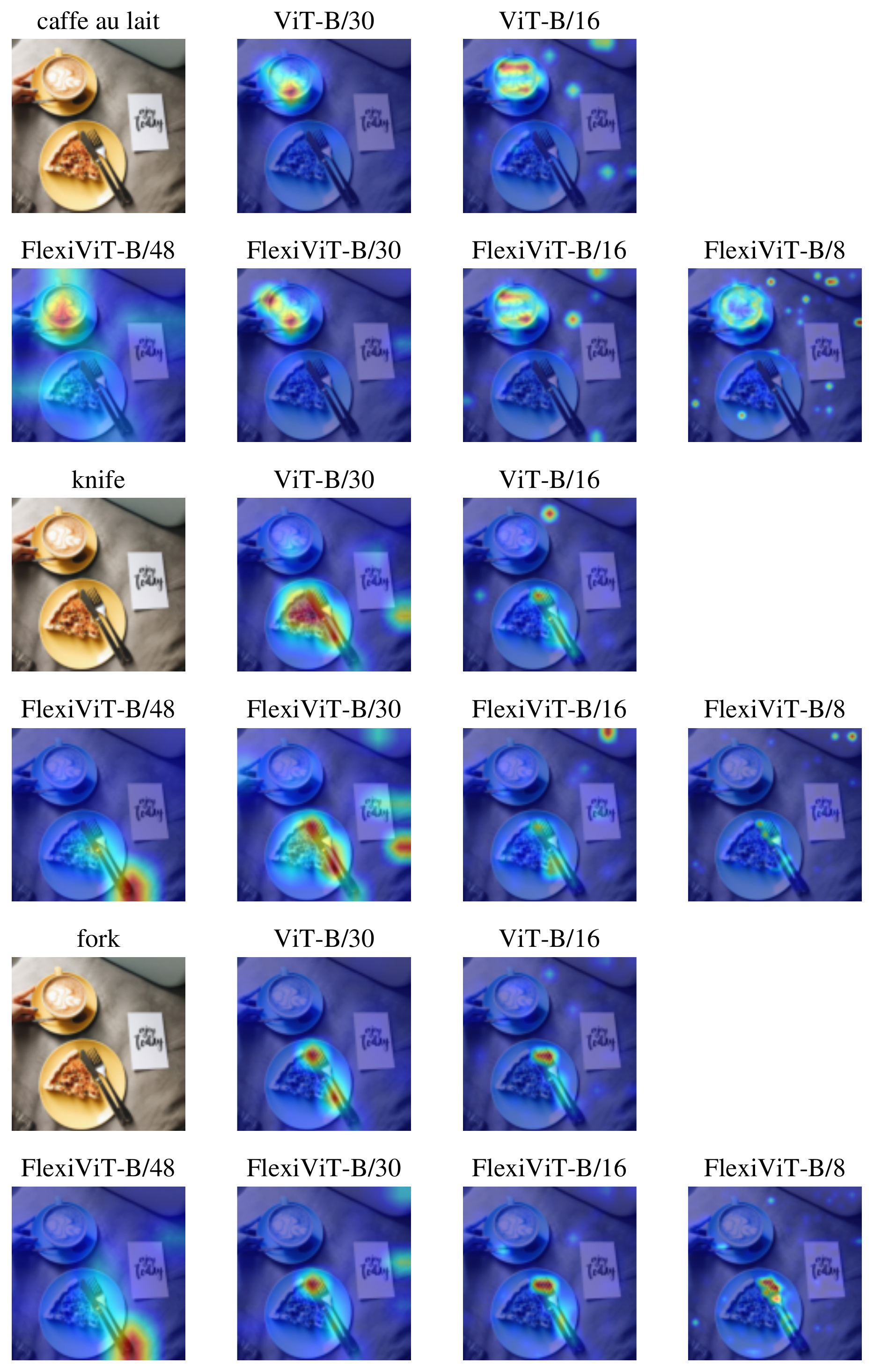}
  \caption{\textbf{Attention relevance maps} of same example as in Figure~\ref{fig:att_token}, with respect to three different objects present in the image (``caffe au lait'', ``knife'' and ``fork'').}
  \label{fig:app:att_3classes}
\end{figure}

We provide the attention relevance maps~\cite{Chefer_2021_CVPR} for the same image as shown in Figure~\ref{fig:att_token} in the main paper, but for all three classes present in the image, in Figure~\ref{fig:app:att_3classes}.

We further provide the attention relevance maps for a random selection of 10 royalty-free images obtained from \url{unsplash.com} for the class that was predicted by all models in Figure~\ref{fig:app:att_unsplash16} at the end of the Appendix.

\section{Flexifying transfer-learning}\label{sec:app:flexify_transfer}

When flexifying transfer-learning, we run the exact same setup as when transferring pre-trained FlexiViT models, described in Section~\ref{sec:app:transfer:clf}, except that we now randomize the patch size during transfer too.

This minor change shows good synergy when combined with using a pre-trained FlexiViT model (green bars), and even enables flexifying plain ViT models during transfer to some degree (orange and olive bars).

\section{Flexifying open-vocabulary detection (OWL-ViT)}\label{sec:app:flexify_owl}
For flexifying OWL-ViT (Section~\ref{sec:flexifying_owl}), we use LiT-\texttt{uu} backbones~\cite{lit}, i.e. CLIP-style models in which the image and text encoders are contrastively pre-trained together (as in the OWL-ViT paper~\cite{minderer2022simple}). We flexify detection training as described in Algorithm~\ref{alg:code} and use resolution $720\times720$. For flexible detection training, we use patch sizes from $48^2$ to $12^2$, i.e. omitting $10^2$ and $8^2$, which would use excessive memory at the higher resolution. Other training settings are as in the OWL-ViT paper~\cite{minderer2022simple}. We find that flexifying both the image-text pre-training and the detection training (Figure~\ref{fig:flexiowl}, pale green line) works slightly better than flexifying just the detection training.

\begin{figure*}[t]
  \centering
  \includegraphics[width=1.0\linewidth]{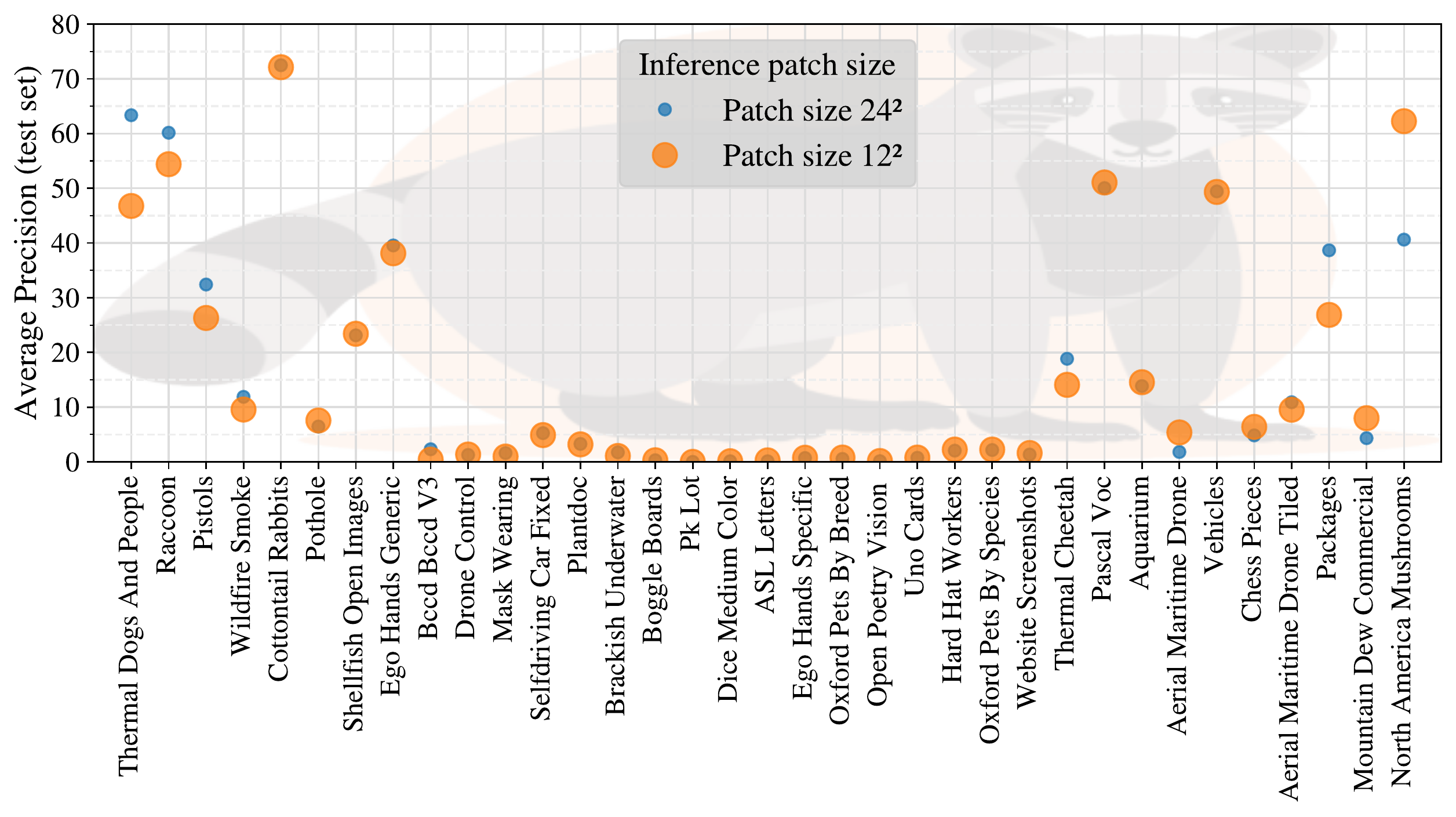}
  \caption{\textbf{Inference-time tuning of patch size can improve performance.} A single OWL-FlexiViT-B model is evaluated at two different patch sizes ($30^2$ and $16^2$) on the 35 different detection tasks of the ELEVATER benchmark.~\cite{li2022elevater}. Tasks are are ordered by the performance difference between patch size $30^2$ and $16^2$ on the \emph{validation} set; the $y$-axis shows performance on the \emph{test} set. The optimal patch size varies by task, such that tuning the patch size on the validation set improves mean test performance from 15.63~AP (at patch size $16^2$) or 16.19~AP (at patch size $30^2$) to 16.62~AP.}
  \label{fig:app:owl_elevater}
\end{figure*}

For evaluating on the ELEVATER~\cite{li2022elevater} set of datasets, we use the model for which both image-text pre-training and detection training were flexified (Figure~\ref{fig:app:owl_elevater}).

\section{Details on flexible depth, stride alternatives}\label{sec:app:alternatives}


\PAR{Common setup} The setup largely follows that introduced in Section~\ref{sec:distill}: distilling the ViT-B/8 model from \cite{vit_augreg} while simultaneously using it for initialization of the student.
Distillation is performed on ImageNet-21k, the labels are ignored and the KL-divergence between student and teacher is the only loss, following \cite{beyer2022knowledge}.
Training is performed for 90 epochs with uniform sampling of patch sizes.

\PAR{Flexible stride}
We train a FlexiViT model variant, where we flexibly change window stride when extracting image patches, but keep the  patch size fixed at $32 \times 32$. In order to perfectly match default grid sizes, and perfectly cover the whole image and avoid padding, we perform minimally required image resize. For example, to get grid size $8 \times 8$ we resize the image to size $242 \times 242$ (from $240 \times 240$) and apply stride 30, or to get grid size $24 \times 24$ we resize the image to size $239 \times 239$ and apply stride 9.

\PAR{Flexible depth}
For every batch, we sample a depth $d$ uniformly from $\{3, 5, 9, 12\}$ and perform a forward pass up to layer $d$. We then apply the classification head, which is shared across all depths, to the class token of layer $d$ and compute the loss. We also explored sampling a depth per-example, but this led to unstable training except when sampling $d$ from all layers $\{3, 4, \ldots, 12\}$. Finally, we also tried using a head per depth as well as a class token per depth, which did not lead to any significant improvement.

\section{Figure~\ref{fig:i1k} with GFLOPs}\label{sec:app:i1k_gflops}

\begin{figure}[t]
  \centering
  \includegraphics[width=1.0\linewidth]{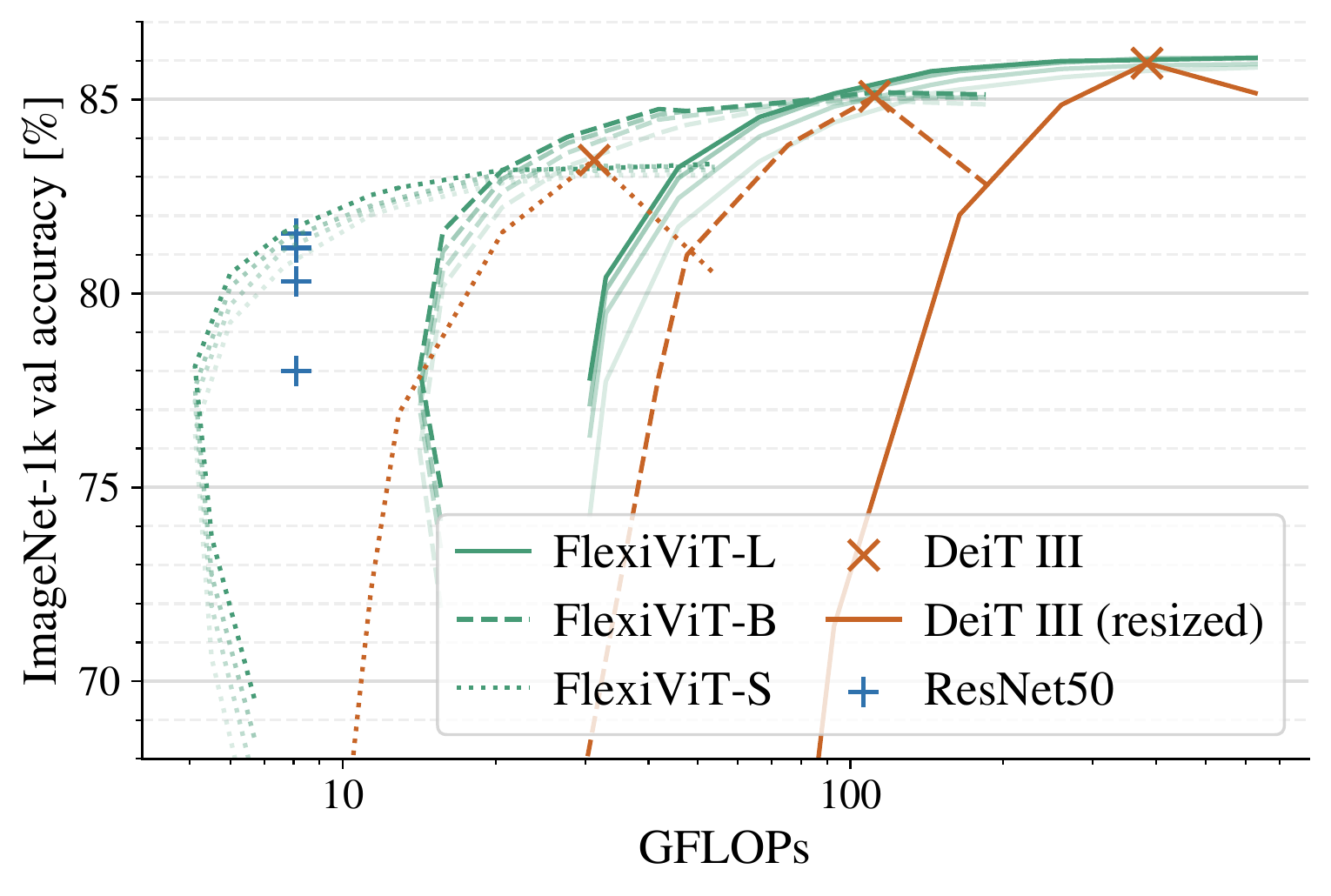}
  \caption{\textbf{Fig~\ref{fig:i1k} using GFLOPs.}}
  \label{fig:app:i1k_gflops}
\end{figure}

Following the Efficiency Misnomer~\cite{misnomer}, we also provide a copy of Figure~\ref{fig:i1k} using GFLOPs as the x-axis instead of inference time as Figure~\ref{fig:app:i1k_gflops}.
This confirms that FLOPs do not always directly translate to wall-clock time in all situations.

\section{Extrapolating patch-size}\label{sec:app:patch_extra}

\begin{figure}[t]
  \centering
  \includegraphics[width=1.0\linewidth]{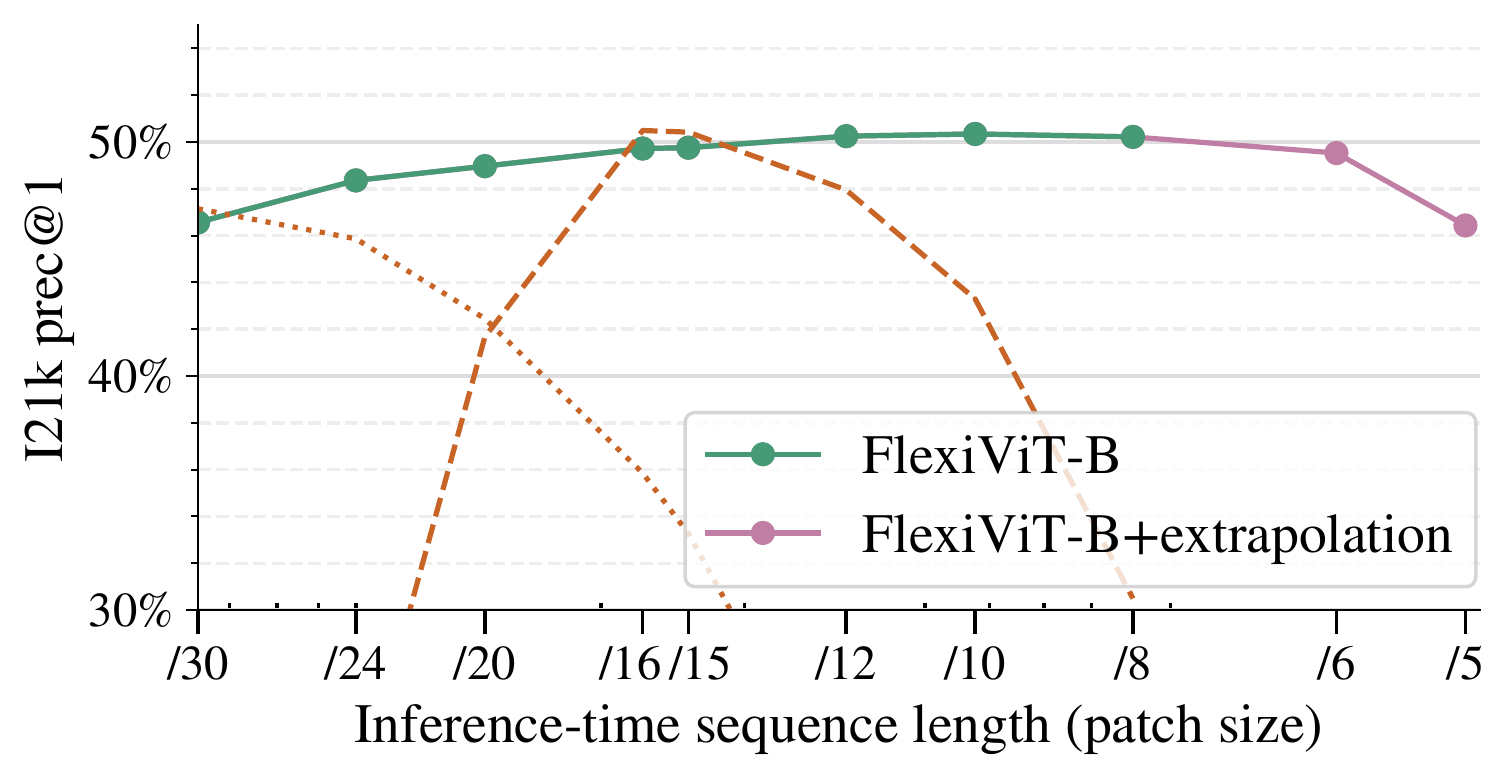}
  \caption{\textbf{Patch-size extrapolation.} We evaluate the model trained on the green patch-sizes at patch-sizes outside the range that was seen during training, and plot these in pink.
  The performance slowly deteriorating means that the model is not able to extrapolate beyond patchsizes or sequence lengths seen during training.}
  \label{fig:app:patch_extra}
\end{figure}

We take the model from the main paper's Section~3.3 and run inference at even smaller patchsizes, see Figure~\ref{fig:app:patch_extra}.
We observe that performance slowly starts to deteriorate.
This means that the model does not learn to generalize to patch-sizes or sequence lengths beyond those seen during the training.
Note that we do not claim extrapolation capabilities, but rather that training many patchsizes into a single model works without loss of quality.

\section{Full tabular results}\label{sec:app:tables}

We provide \cref{tbl:app:i1k_flexi1200,tbl:app:i1k_flexi600,tbl:app:i1k_flexi300,tbl:app:i1k_flexi90,tbl:app:inflexible,tbl:app:distill,tbl:app:using_flexivit,tbl:app:fast_transfer,tbl:app:flexify:transfer,tbl:app:flexilit,tbl:app:flexiowl,tbl:app:flexi_stride_depth,tbl:app:flexiload} which contain the numerical results from all plots from the main paper.

\section{Configuration file for Fig~\ref{fig:i1k}}\label{app:configs}

Algorithm~\ref{alg:config} shows the \texttt{big\_vision}\footnote{\url{https://github.com/google-research/big_vision}} config for training the FlexiViT models from Figure~\ref{fig:i1k}.


\clearpage

\begin{figure*}
  \centering
  \includegraphics[width=1.0\linewidth]{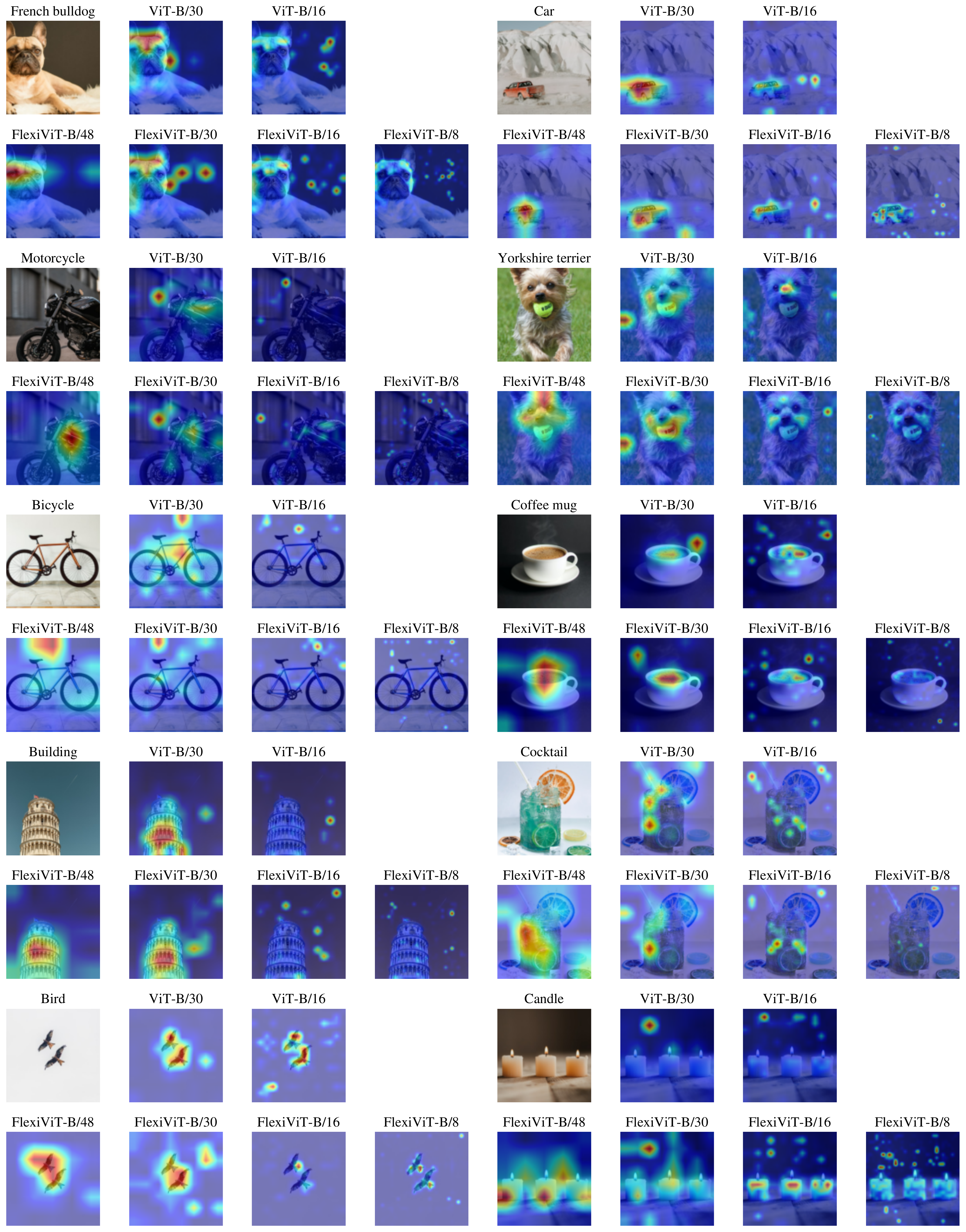}\vspace{-10pt}
  \caption{\textbf{Attention relevance} (as in \cite{Chefer_2021_CVPR}) can significantly change at different patch sizes for both ViT and FlexiViT.}
  \label{fig:app:att_unsplash16}\vspace{-10pt}
\end{figure*}

\clearpage

\clearpage

\lstset{style=overleaf}
\begin{lstlisting}[language=python,label=alg:config]
def get_config(arg=None):                                                                                       
  """Config for training FlexiViT on ImageNet1k."""                                                                
  c = bvcc.parse_arg(arg, variant='B')                                                 
  c.total_epochs = 90                                   
  c.num_classes = 1000        
  c.loss = 'softmax_xent'

  c.input = {}                                                                                                  
  c.input.data = dict(                                  
      name='imagenet2012',                                                                                      
      split='train[:99%]',
  )                                                                                                             
  c.input.batch_size = 1024
  c.input.shuffle_buffer_size = 250_000

  c.log_training_steps = 50
  c.ckpt_steps = 1000

  # Model section
  c.student_name = 'proj.flexi.vit'
  c.student_init = f'deit_3_{c.variant}_384_1k'
  c.student = dict(variant=c.variant,
                   pool_type='tok',
                   patch_size=(16, 16))

  c.teachers = ['prof']
  c.prof_name = 'vit'
  c.prof_init = f'deit_3_{c.variant}_384_1k'
  c.prof = dict(variant=c.variant,
                pool_type='tok',
                patch_size=(16, 16))
  
  pp_label = (
      '|onehot(1000, key="{lbl}", key_result="labels")'
      '|keep("image", "prof", "labels")')
  c.input.pp = (
      'decode|inception_crop|flip_lr'
      '|copy("image", "prof")'
      '|resize(240)'
      '|vgg_value_range'
      '|resize(384, key="prof")'
      '|vgg_value_range(key="prof")'
      + pp_label.format(lbl='label'))
  pp_eval_both = (                                                                                              
      'decode|copy("image", "prof")|'
      f'|resize({240//7*8})'
      '|central_crop(240)'
      '|vgg_value_range'
      f'|resize({384//7*8}, key="prof")'
      '|central_crop(384, key="prof")'
      '|vgg_value_range(key="prof")|')
  pp_eval_student = (
      'decode'
      f'|resize({240//7*8})|central_crop({240})'
      '|value_range(-1, 1)')
  pp_eval_prof = (
      'decode'
      f'|resize({384//7*8})|central_crop(384)'
      '|vgg_value_range(outkey="prof")')

  # Distillation settings
  c.mixup = dict(p=1.0, n=2)
  c.distance = 'kl'
  c.distance_kw = dict(t=1.0)

  # Optimizer section
  c.grad_clip_norm = 1.0
  c.optax_name = 'scale_by_adam'
  c.optax = dict(mu_dtype='bfloat16')

  c.lr = 1e-4
  c.wd = 1e-5
  c.schedule = dict(
      warmup_steps=5000,
      decay_type='cosine')

  # Define the flexible model params:
  c.flexi = dict()
  c.flexi.seqhw = dict(
      # The settings to sample from.
      # Corresponding patch-sizes at 240px:
      # 48, 40, 30, 24, 20, 16, 15, 12, 10, 8
      v=(5, 6, 8, 10, 12, 15, 16, 20, 24, 30),
      # The probs/weights of them (uniform):
      p=(1, 1, 1, 1, 1, 1, 1, 1, 1, 1),
  )

  ####
  # All the rest is just evaluations.
  minitrain = 'train[:2%]'
  minival = 'train[99%:]'

  def get_eval(s, split, dataset='imagenet2012'):       
    return dict(
        type='classification',                          
        pred=f'student_seqhw={s}',                      
        data=dict(name=dataset, split=split),
        pp_fn=(pp_eval_student +
               pp_label.format(lbl='label')),
        loss_name='sigmoid_xent',
        log_percent=0.05,
        cache_final=False,
    )

  c.evals = {}
  for s in c.flexi.seqhw.v:
    c.evals[f'student_minitrain_{s:02d}'] = \
        get_eval(s, minitrain)
    c.evals[f'student_minival_{s:02d}'] = \
        get_eval(s, minival)
    c.evals[f'student_val_{s:02d}'] = \
        get_eval(s, 'validation')
    c.evals[f'student_v2_{s:02d}'] = \
        get_eval(s, 'test', 'imagenet_v2')
    c.evals[f'student_a_{s:02d}'] = \
        get_eval(s, 'test', 'imagenet_a')
    c.evals[f'student_r_{s:02d}'] = \
        get_eval(s, 'test', 'imagenet_r')
    c.evals[f'student_real_{s:02d}'] = \
        get_eval(s, 'validation',
                 'imagenet2012_real')
    c.evals[f'student_real_{s:02d}'].pp_fn = (
        pp_eval_student +
        pp_label.format(lbl='real_label'))

  # A bunch more evals here ...

  return c
\end{lstlisting}


\clearpage

\begin{table}
  \setlength{\tabcolsep}{0pt}
  \setlength{\extrarowheight}{5pt}
  \renewcommand{\arraystretch}{0.75}
  \newcolumntype{C}{>{\centering\arraybackslash}X}
  \newcolumntype{R}{>{\raggedleft\arraybackslash}X}
  \centering
  \caption{Scores for 1200ep ImageNet-1k-only runs from Figure~\ref{fig:i1k}.}\label{tbl:app:i1k_flexi1200}
  \vspace{-0.9em}
  \begin{tabularx}{\linewidth}{p{1.8cm}p{0.1cm}p{1cm}p{0.01cm}p{0.7cm}p{0.01cm}Cp{0.01cm}Cp{0.01cm}Cp{0.01cm}Cp{0.01cm}C}
    \toprule[1pt]
    \bf{Model} && \bf{Eps} && \bf{PS} && \bf{Val} && \bf{ReaL} && \bf{v2} && \bf{-A} && \bf{-R} \\
    \midrule
FlexiViT-S && 1200 && 48² && 69.6 && 76.1 && 55.5 &&  3.3 && 24.1 \\
FlexiViT-S && 1200 && 40² && 73.7 && 80.2 && 60.3 &&  4.7 && 26.4 \\
FlexiViT-S && 1200 && 30² && 78.1 && 84.3 && 65.1 &&  7.4 && 29.2 \\
FlexiViT-S && 1200 && 24² && 80.5 && 86.3 && 68.4 &&  9.4 && 30.5 \\
FlexiViT-S && 1200 && 20² && 81.6 && 87.2 && 70.3 && 12.2 && 31.8 \\
FlexiViT-S && 1200 && 16² && 82.5 && 87.9 && 71.7 && 15.0 && 31.4 \\
FlexiViT-S && 1200 && 15² && 82.7 && 88.1 && 71.8 && 15.7 && 32.9 \\
FlexiViT-S && 1200 && 12² && 83.2 && 88.4 && 72.7 && 17.8 && 32.9 \\
FlexiViT-S && 1200 && 10² && 83.2 && 88.4 && 72.9 && 19.4 && 33.0 \\
FlexiViT-S && 1200 &&  8² && 83.3 && 88.5 && 72.9 && 19.3 && 32.6 \\
\arrayrulecolor{lightgray}\midrule[0.25pt]\arrayrulecolor{black}
FlexiViT-B && 1200 && 48² && 75.0 && 80.5 && 61.1 &&  5.7 && 28.1 \\
FlexiViT-B && 1200 && 40² && 78.0 && 83.3 && 64.7 &&  7.5 && 30.0 \\
FlexiViT-B && 1200 && 30² && 81.6 && 86.5 && 69.7 && 11.4 && 33.0 \\
FlexiViT-B && 1200 && 24² && 83.2 && 87.7 && 71.7 && 15.5 && 34.5 \\
FlexiViT-B && 1200 && 20² && 84.0 && 88.4 && 73.0 && 18.4 && 35.6 \\
FlexiViT-B && 1200 && 16² && 84.7 && 88.8 && 74.0 && 21.7 && 35.8 \\
FlexiViT-B && 1200 && 15² && 84.7 && 88.8 && 74.3 && 22.7 && 36.7 \\
FlexiViT-B && 1200 && 12² && 84.9 && 89.1 && 74.8 && 25.3 && 37.1 \\
FlexiViT-B && 1200 && 10² && 85.2 && 89.2 && 75.0 && 26.7 && 37.2 \\
FlexiViT-B && 1200 &&  8² && 85.1 && 89.2 && 74.9 && 27.1 && 37.2 \\
\arrayrulecolor{lightgray}\midrule[0.25pt]\arrayrulecolor{black}
FlexiViT-L && 1200 && 48² && 77.8 && 83.3 && 63.9 &&  7.1 && 30.0 \\
FlexiViT-L && 1200 && 40² && 80.4 && 85.6 && 67.2 &&  9.9 && 32.7 \\
FlexiViT-L && 1200 && 30² && 83.2 && 87.9 && 70.8 && 14.8 && 35.9 \\
FlexiViT-L && 1200 && 24² && 84.5 && 88.8 && 73.6 && 19.9 && 38.1 \\
FlexiViT-L && 1200 && 20² && 85.1 && 89.4 && 74.9 && 23.5 && 39.4 \\
FlexiViT-L && 1200 && 16² && 85.7 && 89.7 && 76.0 && 28.6 && 39.6 \\
FlexiViT-L && 1200 && 15² && 85.8 && 89.9 && 76.0 && 29.1 && 40.6 \\
FlexiViT-L && 1200 && 12² && 86.0 && 90.0 && 76.5 && 32.0 && 40.8 \\
FlexiViT-L && 1200 && 10² && 86.0 && 90.0 && 76.8 && 33.6 && 40.9 \\
FlexiViT-L && 1200 &&  8² && 86.1 && 90.0 && 76.7 && 34.1 && 41.2 \\
   \bottomrule
  \end{tabularx}
\end{table}

\begin{table}
  \setlength{\tabcolsep}{0pt}
  \setlength{\extrarowheight}{5pt}
  \renewcommand{\arraystretch}{0.75}
  \newcolumntype{C}{>{\centering\arraybackslash}X}
  \newcolumntype{R}{>{\raggedleft\arraybackslash}X}
  \centering
  \caption{Scores for 600ep ImageNet-1k-only runs from Figure~\ref{fig:i1k}.}\label{tbl:app:i1k_flexi600}
  \vspace{-0.9em}
  \begin{tabularx}{\linewidth}{p{1.8cm}p{0.1cm}p{1cm}p{0.01cm}p{0.7cm}p{0.01cm}Cp{0.01cm}Cp{0.01cm}Cp{0.01cm}Cp{0.01cm}C}
    \toprule[1pt]
    \bf{Model} && \bf{Eps} && \bf{PS} && \bf{Val} && \bf{ReaL} && \bf{v2} && \bf{-A} && \bf{-R} \\
    \midrule
FlexiViT-S && 600 && 48² && 68.6 && 75.1 && 54.2 &&  3.2 && 23.9 \\
FlexiViT-S && 600 && 40² && 72.7 && 79.4 && 59.3 &&  4.4 && 26.3 \\
FlexiViT-S && 600 && 30² && 77.6 && 83.8 && 64.6 &&  6.9 && 29.0 \\
FlexiViT-S && 600 && 24² && 80.2 && 86.0 && 67.8 &&  9.2 && 30.4 \\
FlexiViT-S && 600 && 20² && 81.4 && 87.0 && 69.9 && 11.5 && 31.5 \\
FlexiViT-S && 600 && 16² && 82.3 && 87.7 && 71.2 && 14.2 && 31.2 \\
FlexiViT-S && 600 && 15² && 82.5 && 87.9 && 71.5 && 15.1 && 32.7 \\
FlexiViT-S && 600 && 12² && 83.1 && 88.3 && 72.5 && 17.5 && 32.7 \\
FlexiViT-S && 600 && 10² && 83.3 && 88.5 && 72.7 && 19.5 && 32.7 \\
FlexiViT-S && 600 &&  8² && 83.3 && 88.5 && 72.8 && 19.4 && 32.4 \\
\arrayrulecolor{lightgray}\midrule[0.25pt]\arrayrulecolor{black}
FlexiViT-B && 600 && 48² && 74.1 && 80.0 && 60.2 &&  5.3 && 27.7 \\
FlexiViT-B && 600 && 40² && 77.5 && 83.0 && 64.4 &&  7.5 && 30.0 \\
FlexiViT-B && 600 && 30² && 81.1 && 86.1 && 68.9 && 11.2 && 32.7 \\
FlexiViT-B && 600 && 24² && 82.9 && 87.5 && 71.6 && 15.0 && 34.2 \\
FlexiViT-B && 600 && 20² && 83.9 && 88.2 && 72.6 && 17.5 && 35.4 \\
FlexiViT-B && 600 && 16² && 84.6 && 88.7 && 73.9 && 22.1 && 35.7 \\
FlexiViT-B && 600 && 15² && 84.7 && 88.8 && 73.9 && 22.6 && 36.6 \\
FlexiViT-B && 600 && 12² && 84.9 && 89.0 && 74.7 && 25.4 && 36.9 \\
FlexiViT-B && 600 && 10² && 85.1 && 89.2 && 74.8 && 26.9 && 37.2 \\
FlexiViT-B && 600 &&  8² && 85.0 && 89.2 && 74.8 && 27.0 && 36.9 \\
\arrayrulecolor{lightgray}\midrule[0.25pt]\arrayrulecolor{black}
FlexiViT-L && 600 && 48² && 77.1 && 82.6 && 62.7 &&  7.2 && 30.1 \\
FlexiViT-L && 600 && 40² && 80.1 && 85.2 && 66.6 &&  9.4 && 32.6 \\
FlexiViT-L && 600 && 30² && 83.0 && 87.7 && 71.0 && 14.6 && 36.0 \\
FlexiViT-L && 600 && 24² && 84.4 && 88.8 && 73.4 && 19.3 && 38.1 \\
FlexiViT-L && 600 && 20² && 85.1 && 89.4 && 74.7 && 22.5 && 39.3 \\
FlexiViT-L && 600 && 16² && 85.6 && 89.7 && 76.0 && 27.7 && 39.7 \\
FlexiViT-L && 600 && 15² && 85.7 && 89.8 && 75.9 && 28.3 && 40.8 \\
FlexiViT-L && 600 && 12² && 85.9 && 89.9 && 76.4 && 31.0 && 41.0 \\
FlexiViT-L && 600 && 10² && 86.1 && 90.0 && 76.6 && 32.8 && 41.1 \\
FlexiViT-L && 600 &&  8² && 86.1 && 90.0 && 76.6 && 33.2 && 41.3 \\

   \bottomrule
  \end{tabularx}
\end{table}

\begin{table}
  \setlength{\tabcolsep}{0pt}
  \setlength{\extrarowheight}{5pt}
  \renewcommand{\arraystretch}{0.75}
  \newcolumntype{C}{>{\centering\arraybackslash}X}
  \newcolumntype{R}{>{\raggedleft\arraybackslash}X}
  \centering
  \caption{Scores for 300ep ImageNet-1k-only runs from Figure~\ref{fig:i1k}.}\label{tbl:app:i1k_flexi300}
  \vspace{-0.9em}
  \begin{tabularx}{\linewidth}{p{1.8cm}p{0.1cm}p{1cm}p{0.01cm}p{0.7cm}p{0.01cm}Cp{0.01cm}Cp{0.01cm}Cp{0.01cm}Cp{0.01cm}C}
    \toprule[1pt]
    \bf{Model} && \bf{Eps} && \bf{PS} && \bf{Val} && \bf{ReaL} && \bf{v2} && \bf{-A} && \bf{-R} \\
    \midrule
FlexiViT-S && 300 && 48² && 67.4 && 74.0 && 53.2 &&  2.8 && 23.5 \\
FlexiViT-S && 300 && 40² && 71.9 && 78.6 && 58.2 &&  3.9 && 26.0 \\
FlexiViT-S && 300 && 30² && 77.2 && 83.5 && 64.2 &&  6.5 && 29.0 \\
FlexiViT-S && 300 && 24² && 79.7 && 85.6 && 67.6 &&  8.8 && 30.2 \\
FlexiViT-S && 300 && 20² && 81.1 && 86.7 && 69.6 && 11.1 && 31.4 \\
FlexiViT-S && 300 && 16² && 82.1 && 87.6 && 71.1 && 13.9 && 31.1 \\
FlexiViT-S && 300 && 15² && 82.4 && 87.9 && 71.3 && 14.9 && 32.7 \\
FlexiViT-S && 300 && 12² && 83.0 && 88.2 && 72.3 && 17.6 && 32.5 \\
FlexiViT-S && 300 && 10² && 83.2 && 88.3 && 72.9 && 19.3 && 32.4 \\
FlexiViT-S && 300 &&  8² && 83.2 && 88.3 && 73.0 && 19.4 && 32.2 \\
\arrayrulecolor{lightgray}\midrule[0.25pt]\arrayrulecolor{black}
FlexiViT-B && 300 && 48² && 73.4 && 79.3 && 59.7 &&  4.9 && 27.4 \\
FlexiViT-B && 300 && 40² && 77.0 && 82.6 && 63.7 &&  7.0 && 29.6 \\
FlexiViT-B && 300 && 30² && 80.6 && 85.8 && 68.4 && 10.4 && 32.7 \\
FlexiViT-B && 300 && 24² && 82.6 && 87.3 && 71.2 && 14.6 && 34.1 \\
FlexiViT-B && 300 && 20² && 83.6 && 88.1 && 72.5 && 17.3 && 35.1 \\
FlexiViT-B && 300 && 16² && 84.5 && 88.6 && 73.8 && 21.8 && 35.4 \\
FlexiViT-B && 300 && 15² && 84.6 && 88.7 && 73.9 && 22.5 && 36.5 \\
FlexiViT-B && 300 && 12² && 84.9 && 89.0 && 74.6 && 25.5 && 36.9 \\
FlexiViT-B && 300 && 10² && 85.0 && 89.1 && 74.8 && 27.3 && 37.0 \\
FlexiViT-B && 300 &&  8² && 85.1 && 89.1 && 75.0 && 27.1 && 36.9 \\
\arrayrulecolor{lightgray}\midrule[0.25pt]\arrayrulecolor{black}
FlexiViT-L && 300 && 48² && 76.3 && 81.8 && 62.4 &&  6.5 && 30.2 \\
FlexiViT-L && 300 && 40² && 79.5 && 84.7 && 66.4 &&  9.0 && 32.6 \\
FlexiViT-L && 300 && 30² && 82.4 && 87.3 && 70.5 && 13.7 && 35.9 \\
FlexiViT-L && 300 && 24² && 84.0 && 88.5 && 72.8 && 18.0 && 37.7 \\
FlexiViT-L && 300 && 20² && 84.8 && 89.1 && 74.3 && 21.7 && 39.2 \\
FlexiViT-L && 300 && 16² && 85.4 && 89.6 && 75.7 && 26.7 && 39.4 \\
FlexiViT-L && 300 && 15² && 85.5 && 89.7 && 75.8 && 28.1 && 40.6 \\
FlexiViT-L && 300 && 12² && 85.8 && 89.9 && 76.4 && 30.9 && 40.8 \\
FlexiViT-L && 300 && 10² && 85.9 && 89.9 && 76.8 && 32.7 && 41.0 \\
FlexiViT-L && 300 &&  8² && 85.9 && 90.0 && 76.7 && 33.4 && 41.2 \\

   \bottomrule
  \end{tabularx}
\end{table}

\begin{table}
  \setlength{\tabcolsep}{0pt}
  \setlength{\extrarowheight}{5pt}
  \renewcommand{\arraystretch}{0.75}
  \newcolumntype{C}{>{\centering\arraybackslash}X}
  \newcolumntype{R}{>{\raggedleft\arraybackslash}X}
  \centering
  \caption{Scores for 90ep ImageNet-1k-only runs from Figure~\ref{fig:i1k}.}\label{tbl:app:i1k_flexi90}
  \vspace{-0.9em}
  \begin{tabularx}{\linewidth}{p{1.8cm}p{0.1cm}p{1cm}p{0.01cm}p{0.7cm}p{0.01cm}Cp{0.01cm}Cp{0.01cm}Cp{0.01cm}Cp{0.01cm}C}
    \toprule[1pt]
    \bf{Model} && \bf{Eps} && \bf{PS} && \bf{Val} && \bf{ReaL} && \bf{v2} && \bf{-A} && \bf{-R} \\
    \midrule
FlexiViT-S && 90 && 48² && 65.9 && 72.5 && 51.9 &&  2.9 && 23.2 \\
FlexiViT-S && 90 && 40² && 70.6 && 77.3 && 56.9 &&  3.5 && 26.0 \\
FlexiViT-S && 90 && 30² && 76.4 && 82.9 && 62.9 &&  5.9 && 29.3 \\
FlexiViT-S && 90 && 24² && 79.2 && 85.3 && 66.8 &&  7.8 && 30.6 \\
FlexiViT-S && 90 && 20² && 80.7 && 86.5 && 69.0 && 10.6 && 31.5 \\
FlexiViT-S && 90 && 16² && 82.0 && 87.5 && 70.9 && 13.9 && 31.7 \\
FlexiViT-S && 90 && 15² && 82.2 && 87.7 && 71.1 && 14.4 && 32.7 \\
FlexiViT-S && 90 && 12² && 82.8 && 88.1 && 72.0 && 17.3 && 32.5 \\
FlexiViT-S && 90 && 10² && 83.0 && 88.2 && 72.7 && 19.0 && 32.4 \\
FlexiViT-S && 90 &&  8² && 83.0 && 88.3 && 72.7 && 19.4 && 32.2 \\
\arrayrulecolor{lightgray}\midrule[0.25pt]\arrayrulecolor{black}
FlexiViT-B && 90 && 48² && 71.9 && 77.8 && 58.0 &&  4.6 && 26.9 \\
FlexiViT-B && 90 && 40² && 75.9 && 81.6 && 62.2 &&  6.1 && 29.3 \\
FlexiViT-B && 90 && 30² && 80.2 && 85.3 && 67.3 &&  9.5 && 32.3 \\
FlexiViT-B && 90 && 24² && 82.2 && 87.0 && 70.2 && 13.1 && 34.1 \\
FlexiViT-B && 90 && 20² && 83.3 && 87.8 && 71.9 && 16.5 && 35.1 \\
FlexiViT-B && 90 && 16² && 84.1 && 88.4 && 73.2 && 21.3 && 35.2 \\
FlexiViT-B && 90 && 15² && 84.3 && 88.5 && 73.9 && 21.8 && 36.4 \\
FlexiViT-B && 90 && 12² && 84.8 && 88.8 && 74.4 && 25.2 && 36.8 \\
FlexiViT-B && 90 && 10² && 85.0 && 89.0 && 74.5 && 27.3 && 36.9 \\
FlexiViT-B && 90 &&  8² && 84.9 && 89.0 && 74.7 && 27.7 && 36.6 \\
\arrayrulecolor{lightgray}\midrule[0.25pt]\arrayrulecolor{black}
FlexiViT-L && 90 && 48² && 74.3 && 80.0 && 60.2 &&  5.6 && 29.4 \\
FlexiViT-L && 90 && 40² && 77.7 && 83.2 && 64.5 &&  7.7 && 32.2 \\
FlexiViT-L && 90 && 30² && 81.7 && 86.7 && 69.4 && 12.1 && 35.3 \\
FlexiViT-L && 90 && 24² && 83.4 && 88.0 && 72.4 && 17.1 && 37.3 \\
FlexiViT-L && 90 && 20² && 84.4 && 88.8 && 73.9 && 20.9 && 38.9 \\
FlexiViT-L && 90 && 16² && 85.1 && 89.3 && 75.4 && 26.5 && 39.4 \\
FlexiViT-L && 90 && 15² && 85.3 && 89.5 && 75.6 && 27.2 && 40.3 \\
FlexiViT-L && 90 && 12² && 85.6 && 89.7 && 76.3 && 31.2 && 40.5 \\
FlexiViT-L && 90 && 10² && 85.7 && 89.8 && 76.7 && 33.1 && 40.7 \\
FlexiViT-L && 90 &&  8² && 85.8 && 89.9 && 76.6 && 33.7 && 40.6 \\

   \bottomrule
  \end{tabularx}
\end{table}

\clearpage


\begin{table}
  \setlength{\tabcolsep}{0pt}
  \setlength{\extrarowheight}{5pt}
  \renewcommand{\arraystretch}{0.75}
  \newcolumntype{C}{>{\centering\arraybackslash}X}
  \newcolumntype{R}{>{\raggedleft\arraybackslash}X}
  \centering
  \caption{Numerical data for Figure~\ref{fig:inflexible}.}\label{tbl:app:inflexible}
  \vspace{-0.9em}
  \begin{tabularx}{\linewidth}{p{1.0cm}p{0.01cm}Cp{0.01cm}Cp{0.01cm}Cp{0.01cm}Cp{0.01cm}Cp{0.01cm}Cp{0.01cm}Cp{0.01cm}Cp{0.01cm}Cp{0.01cm}C}
    \toprule[1pt]
\bf{Model} && \bf{/48} && \bf{/40} && \bf{/30} && \bf{/24} && \bf{/20} && \bf{/16} && \bf{/15} && \bf{/12} && \bf{/10} && \bf{/8} \\ 
    \midrule
Flexi && 39.5 && 43.2 && 46.6 && 48.4 && 49.0 && 49.7 && 49.8 && 50.2 && 50.3 && 50.2 \\ 
B/16   && \phantom{0}0.0 && \phantom{0}0.1 && \phantom{0}2.4 && 21.6 && 41.7 && 50.5 && 50.4 && 47.9 && 43.3 && 30.5 \\ 
B/30   && 14.0 && 30.2 && 47.1 && 45.9 && 42.5 && 35.9 && 33.3 && 21.0 && 11.9 && \phantom{0}2.9 \\ 
   \bottomrule
  \end{tabularx}
\end{table}


\begin{table}
  \setlength{\tabcolsep}{0pt}
  \setlength{\extrarowheight}{5pt}
  \renewcommand{\arraystretch}{0.75}
  \newcolumntype{C}{>{\centering\arraybackslash}X}
  \newcolumntype{R}{>{\raggedleft\arraybackslash}X}
  \centering
  \caption{Numerical data for Figure~\ref{fig:distill}.}\label{tbl:app:distill}
  \vspace{-0.9em}
  \begin{tabularx}{\linewidth}{p{0.4cm}p{0.8cm}p{0.01cm}Cp{0.01cm}Cp{0.01cm}Cp{0.01cm}Cp{0.01cm}Cp{0.01cm}Cp{0.01cm}Cp{0.01cm}Cp{0.01cm}Cp{0.01cm}C}
    \toprule[1pt]
&&& \bf{/5} && \bf{/6} && \bf{/8} && \bf{/10} && \bf{/12} && \bf{/15} && \bf{/16} && \bf{/20} && \bf{/24} && \bf{/30} \\ 
    \midrule
\multicolumn{22}{c}{\bf{Top-1 accuracy}} \\
    \arrayrulecolor{lightgray}\midrule[0.25pt]\arrayrulecolor{black}
\parbox[t]{2mm}{\multirow{3}{*}{\rotatebox[origin=c]{90}{T-init}}}
& 90 && 40.8 && 43.8 && 47.9 && 49.4 && 50.5 && 51.3 && 51.4 && 51.9 && 51.9 && 52.0 \\ 
& 300 && 43.1 && 45.9 && 48.7 && 50.3 && 51.1 && 51.6 && 51.6 && 51.9 && 52.0 && 52.0 \\ 
& 1000 && 44.1 && 46.6 && 49.2 && 50.6 && 51.2 && 51.9 && 51.8 && 52.1 && 52.3 && 52.2 \\ 
\multicolumn{2}{l}{R-init} && 41.5 && 44.2 && 47.0 && 48.5 && 48.9 && 49.6 && 49.8 && 50.0 && 50.1 && 50.0 \\ 
\multicolumn{2}{l}{None} && 40.6 && 43.4 && 46.6 && 48.1 && 48.9 && 49.7 && 49.7 && 50.0 && 50.2 && 50.1 \\ 
\multicolumn{2}{l}{Teacher} && \multicolumn{19}{c}{52.2} \\ 
    \arrayrulecolor{lightgray}\midrule[0.25pt]\arrayrulecolor{black}
\multicolumn{22}{c}{\bf{Top-1 agreement}} \\
    \arrayrulecolor{lightgray}\midrule[0.25pt]\arrayrulecolor{black}
\parbox[t]{2mm}{\multirow{3}{*}{\rotatebox[origin=c]{90}{T-init}}}
& 90 && 56.0 && 61.9 && 69.5 && 74.7 && 78.1 && 81.1 && 81.8 && 83.8 && 84.5 && 84.4 \\ 
& 300 && 59.6 && 65.7 && 73.0 && 77.3 && 80.1 && 82.8 && 82.9 && 84.2 && 84.5 && 84.5 \\ 
& 1000 && 62.0 && 67.5 && 74.4 && 78.6 && 81.3 && 83.4 && 83.7 && 84.6 && 85.2 && 85.0 \\ 
\multicolumn{2}{l}{R-init} && 56.4 && 60.7 && 66.2 && 69.0 && 70.7 && 72.1 && 72.4 && 72.8 && 73.1 && 73.1 \\ 
    \arrayrulecolor{lightgray}\midrule[0.25pt]\arrayrulecolor{black}
\multicolumn{22}{c}{\bf{CKA similarity}} \\
    \arrayrulecolor{lightgray}\midrule[0.25pt]\arrayrulecolor{black}
\parbox[t]{2mm}{\multirow{3}{*}{\rotatebox[origin=c]{90}{T-init}}}
& 90 && .65 && .69 && .76 && .80 && .84 && .86 && .87 && .88 && .89 && .89 \\ 
& 300 && .68 && .72 && .78 && .82 && .85 && .86 && .87 && .87 && .88 && .87 \\ 
& 1000 && .68 && .72 && .78 && .81 && .83 && .85 && .85 && .86 && .86 && .86 \\ 
\multicolumn{2}{l}{R-init} && .41 && .43 && .45 && .47 && .48 && .49 && .50 && .50 && .50 && .50 \\ 
   \bottomrule
  \end{tabularx}
\end{table}


\begin{table}
  \setlength{\tabcolsep}{0pt}
  \setlength{\extrarowheight}{5pt}
  \renewcommand{\arraystretch}{0.75}
  \newcolumntype{C}{>{\centering\arraybackslash}X}
  \newcolumntype{R}{>{\raggedleft\arraybackslash}X}
  \centering
  \caption{Numerical data for Figure~\ref{fig:using_flexivit}.}\label{tbl:app:using_flexivit}
  \vspace{-0.9em}
  \begin{tabularx}{\linewidth}{p{3.0cm}p{0.01cm}Cp{0.01cm}Cp{0.01cm}Cp{0.01cm}C}
    \toprule[1pt]
\bf{Method} && \bf{/30\,\shorttextrightarrow{}\,/30} && \bf{F\,\shorttextrightarrow{}\,/30} && \bf{/16\,\shorttextrightarrow{}\,/16} && \bf{F\,\shorttextrightarrow{}\,/16} \\ 
    \midrule
Clf SUN397 && 79.1 && 79.7 && 82.3 && 82.5 \\ 
Clf Food101 && 90.4 && 90.8 && 93.7 && 94.1 \\ 
Clf Pets && 93.6 && 93.6 && 94.9 && 94.9 \\ 
Clf Flowers102 && 99.4 && 99.4 && 99.6 && 99.6 \\ 
Clf CIFAR-10 && 98.8 && 99.0 && 99.1 && 99.1 \\ 
Clf CIFAR-100 && 92.3 && 91.9 && 93.2 && 93.3 \\ 
UViM coco PQ && 24.8 && 24.1 && 30.5 && 34.3 \\ 
OWL-ViT lvis AP && 18.8 && 18.4 && 22.7 && 23.4 \\ 
LiT i2t coco && 42.2 && 41.6 && 44.4 && 45.8 \\ 
Seg (lin) City mIoU && 61.0 && 61.1 && 69.3 && 70.0 \\ 
Seg (lin) ADE mIoU && 43.1 && 43.5 && 46.1 && 47.5 \\ 
   \bottomrule
  \end{tabularx}
\end{table}


\begin{table}
  \setlength{\tabcolsep}{0pt}
  \setlength{\extrarowheight}{5pt}
  \renewcommand{\arraystretch}{0.75}
  \newcolumntype{C}{>{\centering\arraybackslash}X}
  \newcolumntype{R}{>{\raggedleft\arraybackslash}X}
  \centering
  \caption{Numerical data for Figure~\ref{fig:fast_transfer}.}\label{tbl:app:fast_transfer}
  \vspace{-0.9em}
  \begin{tabularx}{\linewidth}{p{1.0cm}p{0.01cm}Cp{0.01cm}Cp{0.01cm}Cp{0.01cm}Cp{0.01cm}Cp{0.01cm}Cp{0.01cm}C}
    \toprule[1pt]
\bf{/40} && \bf{/30} && \bf{/24} && \bf{/20} && \bf{/15} && \bf{/12} && \bf{/10} && \bf{/8} \\ 
    \midrule
    \multicolumn{15}{c}{\bf{ViT-B/30 transferred at /30}} \\
78.3 && 82.4 && 83.0 && 82.7 && 80.4 && 75.1 && 66.7 && 58.6 \\ 
    \arrayrulecolor{lightgray}\midrule[0.25pt]\arrayrulecolor{black}
    \multicolumn{15}{c}{\bf{FlexiViT-B transferred at /30}} \\
78.0 && 81.8 && 83.4 && 84.3 && 85.0 && 85.2 && 85.3 && 84.9 \\ 
    \arrayrulecolor{lightgray}\midrule[0.25pt]\arrayrulecolor{black}
    \multicolumn{15}{c}{\bf{FlexiViT-B transferred at /16}} \\
77.0 && 81.2 && 83.2 && 84.5 && 85.5 && 85.9 && 86.0 && 85.7 \\ 
    \arrayrulecolor{lightgray}\midrule[0.25pt]\arrayrulecolor{black}
    \multicolumn{15}{c}{\bf{FlexiViT-B transferred at /8}} \\
76.2 && 80.6 && 82.8 && 84.1 && 85.3 && 85.8 && 86.1 && 86.4 \\ 
    \bottomrule
  \end{tabularx}
\end{table}


\begin{table}
  \setlength{\tabcolsep}{0pt}
  \setlength{\extrarowheight}{5pt}
  \renewcommand{\arraystretch}{0.75}
  \newcolumntype{C}{>{\centering\arraybackslash}X}
  \newcolumntype{R}{>{\raggedleft\arraybackslash}X}
  \centering
  \caption{Numerical data for Figure~\ref{fig:flexify:transfer}.}\label{tbl:app:flexify:transfer}
  \vspace{-0.9em}
  \begin{tabularx}{\linewidth}{p{1.8cm}p{0.01cm}p{1.4cm}p{0.01cm}Cp{0.01cm}Cp{0.01cm}Cp{0.01cm}Cp{0.01cm}Cp{0.01cm}C}
    \toprule[1pt]
\bf{Model} && \bf{Transfer} && \bf{SUN} && \bf{Food} && \bf{Pet} && \bf{Flow} && \bf{C10} && \bf{C100} \\ 
    \midrule
    \multicolumn{15}{c}{\bf{Evaluated at /30}} \\
    \arrayrulecolor{lightgray}\midrule[0.25pt]\arrayrulecolor{black}
ViT-B/30 && /30 && 79.7 && 90.8 && 93.6 && 99.4 && 99.0 && 91.9 \\ 
ViT-B/30 && Flexi && 79.3 && 90.8 && 93.3 && 99.4 && 99.0 && 91.9 \\ 
ViT-B/16 && Flexi && 77.7 && 89.0 && 92.3 && 99.2 && 98.8 && 91.8 \\ 
FlexiViT-B && Flexi && 79.3 && 90.3 && 93.9 && 99.4 && 98.9 && 92.3 \\ 
    \arrayrulecolor{lightgray}\midrule[0.25pt]\arrayrulecolor{black}
    \multicolumn{15}{c}{\bf{Evaluated at /16}} \\
    \arrayrulecolor{lightgray}\midrule[0.25pt]\arrayrulecolor{black}
ViT-B/16 && /16 && 82.3 && 93.7 && 94.9 && 99.6 && 99.1 && 93.2 \\ 
ViT-B/30 && Flexi && 81.6 && 93.0 && 93.8 && 99.5 && 98.7 && 91.3 \\ 
ViT-B/16 && Flexi && 82.1 && 93.6 && 94.9 && 99.5 && 99.1 && 92.8 \\ 
FlexiViT-B && Flexi && 82.8 && 93.8 && 94.8 && 99.6 && 99.1 && 93.0 \\ 
    \arrayrulecolor{lightgray}\midrule[0.25pt]\arrayrulecolor{black}
    \multicolumn{15}{c}{\bf{Evaluated at /8}} \\
    \arrayrulecolor{lightgray}\midrule[0.25pt]\arrayrulecolor{black}
ViT-B/30 && Flexi && 80.9 && 92.7 && 92.2 && 98.8 && 97.8 && 89.0 \\ 
ViT-B/16 && Flexi && 82.4 && 94.0 && 95.0 && 99.6 && 98.7 && 91.6 \\ 
FlexiViT-B && Flexi && 83.2 && 94.7 && 94.9 && 99.6 && 98.9 && 92.8 \\ 
    \bottomrule
  \end{tabularx}
\end{table}


\begin{table}
  \setlength{\tabcolsep}{0pt}
  \setlength{\extrarowheight}{5pt}
  \renewcommand{\arraystretch}{0.75}
  \newcolumntype{C}{>{\centering\arraybackslash}X}
  \newcolumntype{R}{>{\raggedleft\arraybackslash}X}
  \centering
  \caption{Numerical data for Figure~\ref{fig:flexilit}.}\label{tbl:app:flexilit}
  \vspace{-0.9em}
  \begin{tabularx}{\linewidth}{p{2.0cm}p{0.01cm}p{1.0cm}p{0.01cm}Cp{0.01cm}Cp{0.01cm}Cp{0.01cm}Cp{0.01cm}Cp{0.01cm}Cp{0.01cm}C}
    \toprule[1pt]
\bf{Base} && \bf{LiT} && \bf{/48} && \bf{/30} && \bf{/24} && \bf{/16} && \bf{/12} && \bf{/10} && \bf{/8} \\ 
    \midrule
ViT-B/30 && /30 && 46.7 && 65.2 && 65.2 && 64.5 && 30.9 && 10.3 && \phantom{0}3.2 \\ 
ViT-B/16 && /16 && \phantom{0}3.2 && 42.7 && 60.2 && 71.9 && 61.6 && 45.7 && 57.8 \\ 
FlexiViT-B && /30 && 49.0 && 59.8 && 67.0 && 72.5 && 72.6 && 74.7 && 74.1 \\ 
FlexiViT-B && /16 && 48.2 && 62.6 && 66.6 && 73.3 && 73.1 && 74.5 && 75.0 \\ 
FlexiViT-B && Flexi && 51.0 && 62.5 && 69.2 && 73.4 && 74.5 && 75.5 && 75.1 \\ 
    \bottomrule
  \end{tabularx}
\end{table}


\begin{table}
  \setlength{\tabcolsep}{0pt}
  \setlength{\extrarowheight}{5pt}
  \renewcommand{\arraystretch}{0.75}
  \newcolumntype{C}{>{\centering\arraybackslash}X}
  \newcolumntype{R}{>{\raggedleft\arraybackslash}X}
  \centering
  \caption{Numerical data for Figure~\ref{fig:flexiowl}.}\label{tbl:app:flexiowl}
  \vspace{-0.9em}
   \begin{tabularx}{\linewidth}{p{2.0cm}p{0.01cm}p{1.0cm}p{0.01cm}Cp{0.01cm}Cp{0.01cm}Cp{0.01cm}Cp{0.01cm}Cp{0.01cm}Cp{0.01cm}C}
    \toprule[1pt]
\bf{Base} && \bf{OWL} && \bf{/48} && \bf{/40} && \bf{/30} && \bf{/24} && \bf{/20} && \bf{/16} && \bf{/12} \\ 
    \midrule
LiT-B/30 && /30 && \phantom{0}6.2 && 10.8 && 20.6 && 15.6 && \phantom{0}7.8 && \phantom{0}1.7 && \phantom{0}0.3 \\ 
LiT-B/30 && /16 && \phantom{0}0.1 && \phantom{0}0.5 && \phantom{0}4.1 && \phantom{0}9.5 && 16.7 && 26.8 && 15.0 \\ 
LiT-B/30 && Flexi && 15.7 && 17.8 && 21.2 && 23.0 && 24.2 && 25.6 && 26.0 \\ 
FlexiLiT-B && Flexi && 16.0 && 18.5 && 21.5 && 23.5 && 24.9 && 26.7 && 27.1 \\ 
    \bottomrule
  \end{tabularx}
\end{table}


\begin{table*}[t]
  \setlength{\tabcolsep}{0pt}
  \setlength{\extrarowheight}{5pt}
  \renewcommand{\arraystretch}{0.75}
  \newcolumntype{C}{>{\centering\arraybackslash}X}
  \newcolumntype{R}{>{\raggedleft\arraybackslash}X}
  \centering
  \caption{Numerical data for Figure~\ref{fig:flexiload}.}\label{tbl:app:flexiload}
  \vspace{-0.9em}
  \begin{tabularx}{\linewidth}{p{1.0cm}p{0.01cm}Cp{0.01cm}Cp{0.01cm}Cp{0.01cm}Cp{0.01cm}Cp{0.01cm}Cp{0.01cm}Cp{0.01cm}Cp{0.01cm}Cp{0.01cm}Cp{0.01cm}Cp{0.01cm}Cp{0.01cm}Cp{0.01cm}Cp{0.01cm}Cp{0.01cm}C}
    \toprule[1pt]
\bf{Resize} && \bf{/2} && \bf{/4} && \bf{/6} && \bf{/8} && \bf{/10} && \bf{/12} && \bf{/14} && \bf{/16} && \bf{/18} && \bf{/20} && \bf{/22} && \bf{/24} && \bf{/26} && \bf{/28} && \bf{/30} && \bf{/32} \\ 
    \midrule
PI && 10.3 && 44.5 && 48.9 && 52.4 && 52.4 && 52.4 && 52.4 && 52.3 && 52.3 && 52.4 && 52.3 && 52.4 && 52.3 && 52.3 && 52.4 && 52.4 \\ 
Area && 24.5 && 42.6 && 46.1 && 52.4 && 47.7 && 48.3 && 48.6 && 48.1 && 48.7 && 48.4 && 48.5 && 48.7 && 48.5 && 48.5 && 48.6 && 48.4 \\ 
Norm && 10.8 && 40.7 && 46.4 && 52.4 && 45.6 && 44.1 && 42.0 && 39.4 && 37.1 && 33.4 && 29.8 && 26.1 && 22.0 && 18.1 && 14.8 && 11.9 \\ 
Vanilla && 25.5 && 41.6 && 45.6 && 52.4 && 42.5 && 37.2 && 25.5 && 12.8 && \phantom{0}5.4 && \phantom{0}1.9 && \phantom{0}0.6 && \phantom{0}0.2 && \phantom{0}0.1 && \phantom{0}0.0 && \phantom{0}0.0 && \phantom{0}0.0 \\ 
   \bottomrule
  \end{tabularx}
\end{table*}


\begin{table}
  \setlength{\tabcolsep}{0pt}
  \setlength{\extrarowheight}{5pt}
  \renewcommand{\arraystretch}{0.75}
  \newcolumntype{C}{>{\centering\arraybackslash}X}
  \newcolumntype{R}{>{\raggedleft\arraybackslash}X}
  \centering
  \caption{Numerical data for Figure~\ref{fig:flexi_stride_depth}.}\label{tbl:app:flexi_stride_depth}
  \vspace{-0.9em}
  \begin{tabularx}{\linewidth}{p{2.5cm}p{0.01cm}Cp{0.01cm}Cp{0.01cm}Cp{0.01cm}C}
    \toprule[1pt]
\bf{Setting} && \bf{Params} && \bf{GFLOPs} && \bf{Speed} && \bf{Prec} \\ 
    \midrule
    \multicolumn{9}{c}{\bf{Patch FlexiViT-B}} \\
    \arrayrulecolor{lightgray}\midrule[0.25pt]\arrayrulecolor{black}
Eval at patch /30 && 87.5M && 15.7 && 2745 && 47.9 \\ 
Eval at patch /24 && 87.5M && 20.6 && 2022 && 49.4 \\ 
Eval at patch /20 && 87.5M && 27.7 && 1135 && 50.5 \\ 
Eval at patch /16 && 87.5M && 41.9 && 806 && 51.3 \\ 
Eval at patch /15 && 87.5M && 47.6 && 595 && 51.4 \\ 
Eval at patch /12 && 87.5M && 75.3 && 362 && 51.9 \\ 
Eval at patch /10 && 87.5M && 111.5 && 246 && 51.9 \\ 
Eval at patch /8 && 87.5M && 184.5 && 128 && 52.0 \\ 
    \arrayrulecolor{lightgray}\midrule[0.25pt]\arrayrulecolor{black}
    \multicolumn{9}{c}{\bf{Stride FlexiViT-B/32}} \\
    \arrayrulecolor{lightgray}\midrule[0.25pt]\arrayrulecolor{black}
Eval at stride 30 && 87.8M && 11.7 && 2317 && 47.5 \\ 
Eval at stride 23 && 87.8M && 18.2 && 1768 && 49.3 \\ 
Eval at stride 19 && 87.8M && 26.3 && 1041 && 50.3 \\ 
Eval at stride 15 && 87.8M && 41.7 && 750 && 51.0 \\ 
Eval at stride 14 && 87.8M && 47.6 && 563 && 51.2 \\ 
Eval at stride 11 && 87.8M && 76.4 && 347 && 51.5 \\ 
Eval at stride 9 && 87.8M && 113.7 && 235 && 51.6 \\ 
Eval at stride 7 && 87.8M && 188.4 && 124 && 51.7 \\ 
    \arrayrulecolor{lightgray}\midrule[0.25pt]\arrayrulecolor{black}
    \multicolumn{9}{c}{\bf{Depth FlexiViT-B/8}} \\
    \arrayrulecolor{lightgray}\midrule[0.25pt]\arrayrulecolor{black}
Eval at depth 3 && 22.1M && 46.3 && 403 && 35.2 \\ 
Eval at depth 6 && 43.4M && 92.2 && 216 && 44.8 \\ 
Eval at depth 9 && 64.6M && 138.2 && 147 && 48.7 \\ 
Eval at depth 12 && 85.9M && 184.2 && 112 && 51.0 \\ 
    \bottomrule
  \end{tabularx}
\end{table}

\end{document}